\documentclass[twoside]{article}

\usepackage[accepted]{aistats2024}
\usepackage[utf8]{inputenc} % allow utf-8 input
\usepackage[T1]{fontenc}    % use 8-bit T1 fonts
\usepackage{hyperref}       % hyperlinks
\usepackage{url}            % simple URL typesetting
\usepackage{booktabs}       % professional-quality tables
\usepackage{amsfonts}       % blackboard math symbols
\usepackage{nicefrac}       % compact symbols for 1/2, etc.
\usepackage{microtype}      % microtypography
\usepackage{xcolor}         % colors
\usepackage{bold-extra}
\usepackage[scaled]{beramono}

\usepackage[citestyle=authoryear,
natbib=true,backend=biber,doi=false,isbn=false,url=false,maxcitenames=2,uniquelist=false]{biblatex}
\AtEveryBibitem{\clearlist{language}}
\DeclareSourcemap{
  \maps[datatype=bibtex, overwrite]{
    \map{
      \step[fieldset=editor, null]
    }
  }
}

\usepackage{hyperref} % For hyperlinks
\usepackage{url} % For \url
\usepackage{enumitem} % For removing the space before and between items
\usepackage{outlines} % For simple itemize lists
\usepackage{xr} % For referencing external files

\usepackage{amsmath,amsthm,amsfonts,amssymb} % Math
\usepackage{bbm} % For the indicator function
\usepackage{bm} % Bold mathematics
\usepackage{dsfont} % \mathds
\usepackage{mathtools} % For KL divergence
\usepackage{braket} % For \Set
\usepackage{thmtools} % Restate theorems
\usepackage{thm-restate} % Restate theorems

\usepackage[ruled,vlined,noend]{algorithm2e} % Algorithms
\usepackage{listings} % Typeset programs within the document

\usepackage{csquotes} % For quotes in the bibliography
\usepackage[capitalize,noabbrev]{cleveref} % \cref
\usepackage{nameref} % References by names

\usepackage{float} % H option
\usepackage{graphicx} % For \includegraphics
\usepackage[export]{adjustbox} % Extends \includegraphics
\usepackage{svg} % To include svg images
\usepackage{tikz} % Draw graphics in LaTeX
\usepackage{caption} % Caption options
\usepackage{subcaption} % (incompatible with subfigure)
\usepackage{stfloats} % To improve the placement of floats
\usepackage{placeins} % For \FloatBarrier

\usepackage{array,tabularx} % Extends tabular
\usepackage{booktabs} % For /toprule (used by pandas dataframes)
\usepackage{longtable} % For multipage tables
\usepackage{multirow}
\usepackage{multicol} % Multicolumn bibliography
\usepackage{makecell}
\usepackage{afterpage} % For blank pages before tables

\usepackage{soul} % Highlighting with \hl
\usepackage{indentfirst} % Indent first paragraph after section title
\usepackage{hyphenat}

\usepackage[toc,page,header]{appendix}
\usepackage{minitoc}

\AtNextBibliography{\small}
\renewbibmacro{in:}{\ifentrytype{article}
	{}
	{\bibstring{in}\printunit{\intitlepunct}}
}

\SetEndCharOfAlgoLine{}
\SetKwComment{Comment}{$\triangleright$\ }{}
\let\oldnl\nl% Store \nl in \oldnl
\newcommand{\nonl}{\renewcommand{\nl}{\let\nl\oldnl}}% Remove line number for one line

\crefname{equation}{}{}

\theoremstyle{definition}

\theoremstyle{definition}

\theoremstyle{remark}

\makeatletter
\newcommand{\includesvgfullpath}[2][\textwidth]{%
	\filename@parse{#2}%
	\includesvg[inkscapepath=svg-inkscape/\filename@area,width=#1]{#2}%
}
\makeatother

\newcommand{\R}{\mathbb{R}}

\newcommand{\D}{\mathcal{D}}

\DeclarePairedDelimiterX{\infdivx}[2]{(}{)}{%
	#1\;\delimsize\|\;#2%
}

\newcommand{\KL}{D_{\mathrm{KL}}\infdivx*}

\DeclareMathAlphabet{\mathmybb}{U}{bbold}{m}{n}
\newcommand{\indicator}{\mathmybb{1}}

\let\mc\mathcal                                             % FANCY FONT (Big-Oh)
                                                 % ...
                                                % ...
\let\tt\texttt                                              % MONOSPACED TEXT
\def\ceil#1{\lceil #1 \rceil}

\DeclareMathAlphabet{\mathsfit}{\encodingdefault}{\sfdefault}{m}{sl}
\SetMathAlphabet{\mathsfit}{bold}{\encodingdefault}{\sfdefault}{bx}{n}

\begin{document}
	
	% If your paper is accepted and the title of your paper is very long,
	% the style will print as headings an error message. Use the following
	% command to supply a shorter title of your paper so that it can be
	% used as headings.
	%
	\runningtitle{Probabilistic Calibration by Design for Neural Network Regression}
	
	% If your paper is accepted and the number of authors is large, the
	% style will print as headings an error message. Use the following
	% command to supply a shorter version of the authors names so that
	% they can be used as headings (for example, use only the surnames)
	%
	%\runningauthor{Surname 1, Surname 2, Surname 3, ...., Surname n}
	
	\twocolumn[
	
	\aistatstitle{Probabilistic Calibration by Design\\for Neural Network Regression}
	
	\aistatsauthor{ Victor Dheur \And Souhaib Ben Taieb }
	
	\aistatsaddress{ Department of Computer Science,\\ University of Mons, Belgium } ]
	
	%%%%%%%%%%%%%%%%%%%%%%%%%%%%%%%%%%%%%%%%%%%%%%%%%%%%%
	
	\begin{abstract}

Generating calibrated and sharp neural network predictive distributions for regression problems is essential for optimal decision-making in many real-world applications. To address the miscalibration issue of neural networks, various methods have been proposed to improve calibration, including post-hoc methods that adjust predictions after training and regularization methods that act during training. While post-hoc methods have shown better improvement in calibration compared to regularization methods, the post-hoc step is completely independent of model training. We introduce a novel end-to-end model training procedure called Quantile Recalibration Training, integrating post-hoc calibration directly into the training process without additional parameters. We also present a unified algorithm that includes our method and other post-hoc and regularization methods, as particular cases. We demonstrate the performance of our method in a large-scale experiment involving 57 tabular regression datasets, showcasing improved predictive accuracy while maintaining calibration. We also conduct an ablation study to evaluate the significance of different components within our proposed method, as well as an in-depth analysis of the impact of the base model and different hyperparameters on predictive accuracy.

	\end{abstract}
	
	\section{INTRODUCTION}
	Critical decisions depend on the predictions made by neural networks in many applications such as medical diagnostics and autonomous driving \citep{Begoli2019-go,Michelmore2018-uv}.
	To make decisions effectively, it is often crucial to quantify predictive uncertainty accurately \citep{Gawlikowski2021-ty,Abdar2021-zq}.
	Yet, neural networks might exhibit miscalibration \citep{Guo2017-ow}.
	
	We focus on regression models that output a predictive distribution.
	Central to our study, \textit{probabilistic calibration}\footnote{In this paper, we refer to probabilistic calibration as calibration to simplify terminology.} \citep{Gneiting2007-la} is an important property that states that all quantiles must be calibrated. 
	This implies that the predicted 90\% quantiles should exceed 90\% of the corresponding realizations.
	%Furthermore, \citet{Gneiting2007-la} introduced the paradigm of maximizing the sharpness of the predictive distributions subject to calibration, which means that predictive distributions should provide maximum information at the condition of representing uncertainty accurately.
	
	%The finding that neural networks are miscalibrated \citep{Guo2017-ow} has led to an increased interest in neural network calibration
	
	Several methods have been proposed to improve probabilistic calibration and they can be divided into two main categories.
	Post-hoc methods such as Quantile Recalibration \citep{Kuleshov2018-tb} act after training a base model and transform the predictions based on a separate calibration dataset.
	Regularization methods act during training and add a regularization term that penalizes calibration \citep{Chung2021-rh}. Empirical evidence suggests that post-hoc methods outperform regularization methods in terms of calibration within the context of regression \citep{Dheur2023-bo}. This superiority has been attributed to the finite-sample guarantee from which post-hoc methods benefit.
	%In regression, post-hoc methods have been shown empirically to outperform regularization methods in terms of calibration, which has been explained by the finite-sample guarantee that they provide \citep{Dheur2023-bo}.

	This paper introduces a novel method called \emph{Quantile Recalibration Training} that seamlessly integrates post-hoc calibration into the training process, resulting in an end-to-end approach. Our method leverages the concept of minimizing the sharpness of predictions while ensuring calibration \citep{Gneiting2007-la}. By minimizing the negative log-likelihood (NLL), our approach achieves the desired sharpness, while simultaneously ensuring calibration at each training step using a dedicated calibration dataset.
	Recalibration Training stands apart from other regularization methods by offering improvements in both the NLL and calibration of the final model.
	% This is in contrast to traditional regularization methods that often focus solely on optimizing the NLL, without explicit consideration for calibration. 
	Our approach aligns with the recommendation made by \citet{Wang2021-jy} to view model training and post-hoc calibration as an integrated framework rather than treating them as separate steps.
	The code base, available at \url{https://github.com/Vekteur/quantile-recalibration-training}, has been used for the implementation of all methods to ensure a fair comparison.

	We make the following main contributions:
	\begin{enumerate}
		\item We propose a novel training procedure to learn predictive distributions that are probabilistically calibrated at every training step, called \emph{Quantile Recalibration Training} (see Section \ref{sec:recalibration_training}). We also propose an algorithm which unifies our Quantile Recalibration Training with Quantile Recalibration, Quantile Regularization and standard NLL minimization.

		\item We demonstrate the effectiveness of our method in a large-scale experiment based on 57 tabular datasets. The results show improved NLL on the test set while at the same time ensuring calibration (see \cref{sec:experiments}).
		
		%Additionally, our method can be synergistically combined with post-hoc calibration techniques to achieve further improvements in calibration performance.
		
		\item We provide an in-depth analysis of the impact of the base model and different hyperparameters on predictive accuracy and calibration.
		We also conduct an ablation study to evaluate the significance of different components within our proposed method (see \cref{sec:other_experiments}).
		
		%Furthermore, we identify and discuss a potential issue associated with NLL minimization on datasets with discrete targets. By addressing these aspects, our study offers valuable insights and strengthens the overall validity and applicability of our proposed method.
		
	\end{enumerate}
	
	\section{BACKGROUND ON PROBABILISTIC CALIBRATION}
	\label{sec:background}
	
	We consider a univariate regression problem where a target variable $Y \in \mc{Y}$ depends on an input variable $X \in \mc{X}$, where $\mc{X}$ is the input space and $\mc{Y} \subseteq \R$ is the target space.
	Our goal is to approximate the conditional distribution $P_{Y \mid X}$ based on i.i.d. training data $\D = \Set{(X_i, Y_i)}_{i=1}^N$ where $(X_i, Y_i) \sim P_{X, Y} = P_{Y \mid X} P_X$.
	
	A probabilistic predictor, denoted as $F_\theta: \mc{X} \to \mc{F}$ is defined by its parameters $\theta$ from the parameter space $\Theta$. This function maps an input $x \in \mc{X}$ to a predictive cumulative distribution function (CDF) $F_\theta(\cdot \mid x)$ in the space $\mc{F}$ of distributions over $\R$. This CDF has an associated probability density function (PDF) given by $f_\theta(\cdot \mid x)$.
	%This model belongs to an hypothesis space $\mc{H} = \Set{ h_{\theta} : \theta \in \Theta }$.
	%Similarly, the CDF or PDF of a random variable $Y$ is denoted by $F_Y$ or $f_Y$, respectively.
	
	\paragraph{Probabilistic calibration}
	Given a possibly miscalibrated CDF $F_\theta$, let $Z = F_\theta(Y \mid X) \in [0, 1]$ denote the probability integral transform (PIT) of $Y$ conditional on $X$ and $F_Z(\alpha) = \text{Pr}(Z \leq \alpha)$ the corresponding CDF.
	The model $F_\theta$ is probabilistically calibrated (also known as PIT-calibrated or quantile calibrated) if $Z$ is distributed uniformly between 0 and 1:
	\begin{equation}
		\label{eq:probabilistic_calibration}
		F_Z(\alpha) = \alpha \quad \forall \alpha \in [0, 1].
	\end{equation}
	The CDF $F_Z$ is usually estimated from data using the empirical CDF, that we denote $\Phi_\theta^\text{EMP}(\alpha) = \frac{1}{N} \sum_{i=1}^N \indicator(Z_i \leq \alpha)$, where $Z_i = F_\theta(Y_i \mid X_i)$. We measure probabilitic calibration using the probabilistic calibration error (PCE), defined as
	\begin{equation}
		\label{eq:PCE}
		\text{PCE}(F_\theta) = \frac{1}{M} \sum_{j=1}^M \left|\alpha_j - \Phi_\theta^\text{EMP}(\alpha_j) \right|,
	\end{equation}
	where $0 < \alpha_1 < \dots < \alpha_M < 1$ are equidistant quantile levels. The number of levels $M$ is fixed at 100 in the paper. In essence, PCE computes the discrepancy between the r.h.s and l.h.s. of \eqref{eq:probabilistic_calibration} for multiple values of $\alpha$.

	%Based on \cref{eq:probabilistic_calibration}, we can derive the probabilistic calibration error (PCE), which is a common way to evaluate probabilistic calibration:
	
	\paragraph{Quantile recalibration}
	\textit{Quantile Recalibration} (QR, \cite{Kuleshov2018-tb}) computes a probabilistically calibrated CDF $F'_\theta = F_Z \circ F_\theta$, where $F_Z$ is estimated from data. In fact, for each quantile level $\alpha \in [0, 1]$, we have:
	\begin{align}
		\text{Pr}(F'_\theta(Y \mid X) \leq \alpha) &= \text{Pr}(F_\theta(Y \mid X) \leq F^{-1}_Z(\alpha)) \\
		 &= F_Z(F^{-1}_Z(\alpha)) \\
		 &= \alpha,
	\end{align}
	which shows that $F'_\theta$ is calibrated.
	
	\paragraph{Calibration map}
	The estimator of $F_Z$, called a calibration map, can be the empirical CDF $\Phi_\theta^\text{EMP}$ computed from the PITs $Z'_i = F_\theta(Y'_i \mid X'_i)$ of a separate i.i.d. calibration dataset $\D' = \Set{(X'_i, Y'_i)}_{i=1}^{N'}$.
	
	\sloppy Since $\Phi_\theta^\text{EMP}$ is not differentiable, the resulting calibrated CDF $F'_\theta$ is not differentiable either.
	\citet{Dheur2023-bo} proposed to compute a differentiable calibration map
	\begin{equation}
		\label{eq:Phi_theta^KDE}
		\Phi_\theta^\text{KDE}(\alpha) = \frac{1}{N'} \sum_{i=1}^{N'} F_\text{Log}(\alpha; Z'_i, b^2 N'^{\nicefrac{-2}{5}}),
	\end{equation}
	based on kernel density estimation (KDE). This corresponds to a mixture of logistic CDFs $F_\text{Log}$ with means $Z'_1, \dots, Z'_N$ and a variance $b^2 N'^{\nicefrac{-2}{5}}$ following Scott's rule \citep{Scott1979-qe}. The bandwidth $b > 0$ is a hyperparameter controlling the smoothness of the calibration map.
	Note that $\Phi_\theta^\text{KDE}$ converges to $\Phi_\theta^\text{EMP}$ as $b \to 0$.
	
	Furthermore, \citet{Dheur2023-bo} showed that QR provides a finite-sample guarantee with a specific calibration map, namely:
	\begin{equation}
		\label{eq:calibration_guarantee}
		\text{Pr}(\Phi_\theta^\text{DCP}(F_\theta(Y \mid X)) \leq \alpha) = \frac{\ceil{(N' + 1) \alpha}}{N' + 1} \approx \alpha,
	\end{equation}
	where $\Phi_\theta^\text{DCP}(\alpha) = \frac{1}{N'+1} \sum_{i=1}^{N'} \indicator(Z'_i \leq \alpha)$ is a calibration map derived from Distributional Conformal Prediction \citep{Chernozhukov2021-sg,Izbicki2020-ed}.
	This property is approximately obtained by other calibration maps such as $\Phi_\theta^\text{EMP}$ and $\Phi_\theta^\text{KDE}$.
	We note that the probability in \cref{eq:calibration_guarantee} is also taken over the calibration dataset $\D'$.
	
	%We emphasize that the guarantee \cref{eq:calibration_guarantee} is only valid in expectation over the calibration datasets, and not conditionally to a specific calibration dataset:
	%\begin{equation}
	%	\label{eq:calibration_guarantee_expectation}
	%	\text{Pr}(\Phi_\theta^\text{DCP}(F_\theta(Y \mid X)) \leq \alpha)
	%	=
	%	\EE{\D'}{\text{Pr}(\Phi_\theta^\text{DCP}(F_\theta(Y \mid X)) \leq \alpha \mid \D')}.
	%\end{equation}
	
	\paragraph{Quantile Regularization}
	
	Recently, there has been a surge of interest in regularization strategies for calibration based on differentiable objectives that are optimized during training (see \cref{sec:related_work}).
	%The basic idea is to make a regularization objective that encourages calibration with a gradient that is effective for learning.
	\textit{Quantile Regularization} (QREG, \cite{Utpala2020-nw}) minimizes a loss function of the form
	\begin{equation}
		-\frac{1}{N} \sum_{i=1}^N \log f_\theta(Y_i \mid X_i) + \lambda \mc{R}_\text{QREG}(\theta),
	\end{equation}
	where the first term is the NLL and $\lambda > 0$ is a regularization hyperparameter.
	%and $\mc{L}(\theta)$ should be a strictly proper scoring rule such as the negative log-likelihood (NLL) $\mc{L}(\theta) = $.
	The regularization function $\mc{R}_\text{QREG}(\theta)$ encourages calibration by minimizing the KL divergence between $Z$ and a uniformly distributed random variable $U$. This reduces to maximizing the differential entropy $H(Z)$ of $Z$ since $\KL{Z}{U} = -H(Z)$.
	
	\cite{Utpala2020-nw} propose to estimate $H(Z)$ using sample-spacing entropy estimation \citep{Vasicek1976-fa}:
	\begin{align}
		& \mc{R}_{\text{QREG}}(\theta) \\           %                                                      %   \\
		& = \frac{1}{N-k} \sum_{i=1}^{N-k} \log \left[ \frac{N+1}{k} (Z_{(i+k)} - Z_{(i)}) \right] \\
		& \approx -H(Z) \label{eq:quantile_regularization},
	\end{align}
	where $k$ is a hyperparameter such that $1 \leq k \leq N$ and $Z_{(i)}$ is the $i$th order statistics of $Z_1, \dots, Z_N$.
	To ensure differentiability during optimization, the authors employed a differentiable relaxation technique called NeuralSort \citep{Grover2019-iw}, as sorting is a non-differentiable operation.
	
	We note that this approach should be distinguished from regularizers in classification \citep{Pereyra2017-lk} that maximize the entropy of the target $Y$ and not the differential entropy of the PIT $Z$.
	
%	Due to the indicator function, PCE \cref{eq:PCE} is not differentiable and can not be used as a regularization objective.
%	However, \citet{Dheur2023-bo} showed that, based on the differentiable calibration map $\Phi_\theta^\text{KDE}$, it is possible to derive a regularization term for probabilistic calibration given by:
%	\begin{align}
%		\label{eq:PCE-KDE}
%		\mc{R}_\text{PCE-KDE}(\theta) = \frac{1}{M} \sum_{j=1}^M \left|\alpha_j - \Phi_\theta^\text{KDE}(\alpha_j) \right|.
%	\end{align}
%	
%	The resulting loss function is of the form $\mc{L}(\theta) + \lambda \mc{R}_\text{PCE-KDE}(\theta)$, where $\lambda > 0$ is a regularization hyperparameter and $\mc{L}(\theta)$ should be a strictly proper scoring rule such as the negative log-likelihood (NLL) $\mc{L}(\theta) = -\frac{1}{N} \sum_{i=1}^N \log f_\theta(Y_i \mid X_i)$.

	\section{QUANTILE RECALIBRATION TRAINING}
	\label{sec:recalibration_training}
	
	We introduce \textit{Quantile Recalibration Training} (QRT), a novel method for training neural network regression models.
	Predictive distributions are iteratively recalibrated during model training and are hence calibrated by design.
	
	\subsection{The QRT learning procedure}
	\label{sec:decomposition}
	
	Recall that QR involves training $F_\theta$ by minimizing the NLL and then adjusting it by producing a revised predictive distribution $F'_\theta = \Phi_\theta^\text{KDE} \circ F_\theta$.
	Given that the recalibration map $\Phi_\theta^\text{KDE}$ is differentiable, our QRT procedure integrates it end-to-end into the optimization procedure. Specifically, we directly minimize the NLL of $F'_\theta$ which involves iteratively recalibrating it during training.
	Using the chain rule, the NLL of $F'_\theta$ can be conveniently decomposed as follows:
	\begin{align}
		&{\textstyle \sum}_{i=1}^N {-\log f'_\theta(Y_i \mid X_i)} \\
		&= {\textstyle \sum}_{i=1}^N{-\log f_\theta(Y_i \mid X_i)} {-\log f_Z(F_\theta(Y_i \mid X_i))} \label{eq:f_Z} \\
		&= {\textstyle \sum}_{i=1}^N{-\log f_\theta(Y_i \mid X_i)} + \hat{H}(Z) \label{eq:RT_decomposition}.
	\end{align}
%	\begin{align}
%		&\E{-\log f'_\theta(Y \mid X)} \\
%		&= \E{-\log f_\theta(Y \mid X)} + \E{-\log f_Z(F_\theta(Y \mid X))} \label{eq:f_Z} \\
%		&= \E{-\log f_\theta(Y \mid X)} + H(Z) \label{eq:RT_decomposition},
%	\end{align}
	The first term in \cref{eq:RT_decomposition} is the NLL of the base model $F_\theta$ and $\hat{H}(Z)$ can be interpreted as the entropy of $Z$.
	
	Interestingly, the second term $\hat{H}(Z)$ corresponds to the opposite of the regularization term of QREG \cref{eq:quantile_regularization}.
	This observation could seem counter-intuitive since it implies that, when training QRT, the PCE of $F_\theta$ will be maximized by the second term $\hat{H}(Z)$ in the decomposition.
	%This implies that, during QRT, the PCE of $F_\theta$ can be arbitrarily high.
	However, QRT is valid since it produces $F'_\theta$ by minimizing the NLL of $F'_\theta$, which is a strictly proper scoring rule.
	
	To compute the second term in \cref{eq:f_Z}, we estimate $f_Z$ using the derivative of the calibration map $\Phi_\theta^\text{KDE}$, which has a closed-form expression given by
	\begin{align}
		\label{eq:derivative_calibration_map}
		\phi^\text{KDE}_\theta(\alpha) 
		&= \nicefrac{\partial \Phi_\theta^\text{KDE}(\alpha)}{\partial \alpha} \\
		&= \frac{1}{N} \sum_{i=1}^{N} f_\text{Log}(\alpha; Z_i, b^2 N^{\nicefrac{-2}{5}}),
	\end{align}
	where $f_\text{Log}$ is the PDF of a logistic distribution, as described in \cref{sec:background}.
	
	During training, $\phi^\text{KDE}_\theta$ is computed on the current batch and $F'_\theta$ is thus, by design, calibrated on the current batch.
	However, it does not satisfy the calibration guarantee \cref{eq:calibration_guarantee} since the current batch has been seen during training.
	Hence, as a final step, we perform QR on a separate calibration dataset to obtain the calibration guarantee.
	We give more details in \cref{sec:RT_framework}.
	
	Finally, to account for the finite domain $[0, 1]$ of the PIT $Z$, we perform a slight adjustment to the calibration map $\phi_\theta^\text{KDE}$.
	The standard approach is to truncate the distribution by redistributing the density that has been estimated outside of $[0, 1]$ uniformly in $[0, 1]$.
	Instead, \cite{Blasiok2023-oh} propose to redistribute the density slightly outside of $[0, 1]$ near $0$ and $1$, assuming that $\phi_\theta^\text{KDE}(x) = 0$ for $x \not \in [-1, 2]$.
    The resulting calibration map $\phi_\theta^\text{REFL}$ is defined as:
	\begin{align}
		&\phi_\theta^\text{REFL}(x) = \label{eq:refl} \\ 
		&\begin{cases}
			\phi_\theta^\text{KDE}(x) + \phi_\theta^\text{KDE}(-x) + \phi_\theta^\text{KDE}(2 - x) &\quad \text{if $x \in [0, 1]$} \\
			0 &\quad \text{if $x \not\in [0, 1]$}. \nonumber
		\end{cases} 
	\end{align}
        % \begin{align*}
        %     &\Phi_\theta^\text{REFL}(x) = \\
        %     &\begin{cases}
        %         \Phi_\theta^\text{KDE}(x) - \Phi_\theta^\text{KDE}(2 - x) + 1 - \Phi_\theta^\text{KDE}(2 - x) &\quad \text{if $x \in [0, 1]$} \\
        %         0 &\quad \text{if $x < 0$} \\
        %         1 &\quad \text{if $x > 1$} \\
        %     \end{cases}
        % \end{align*}
	This approach avoids an ill-defined calibration map and often leads to improved NLL.
	More motivation and details, including the definition of the corresponding CDF $\Phi_\theta^\text{REFL}$, are given in \cref{sec:kde_finite}.
	
	\begin{figure*}[t]
		\centering
		\includegraphics[width=\linewidth]{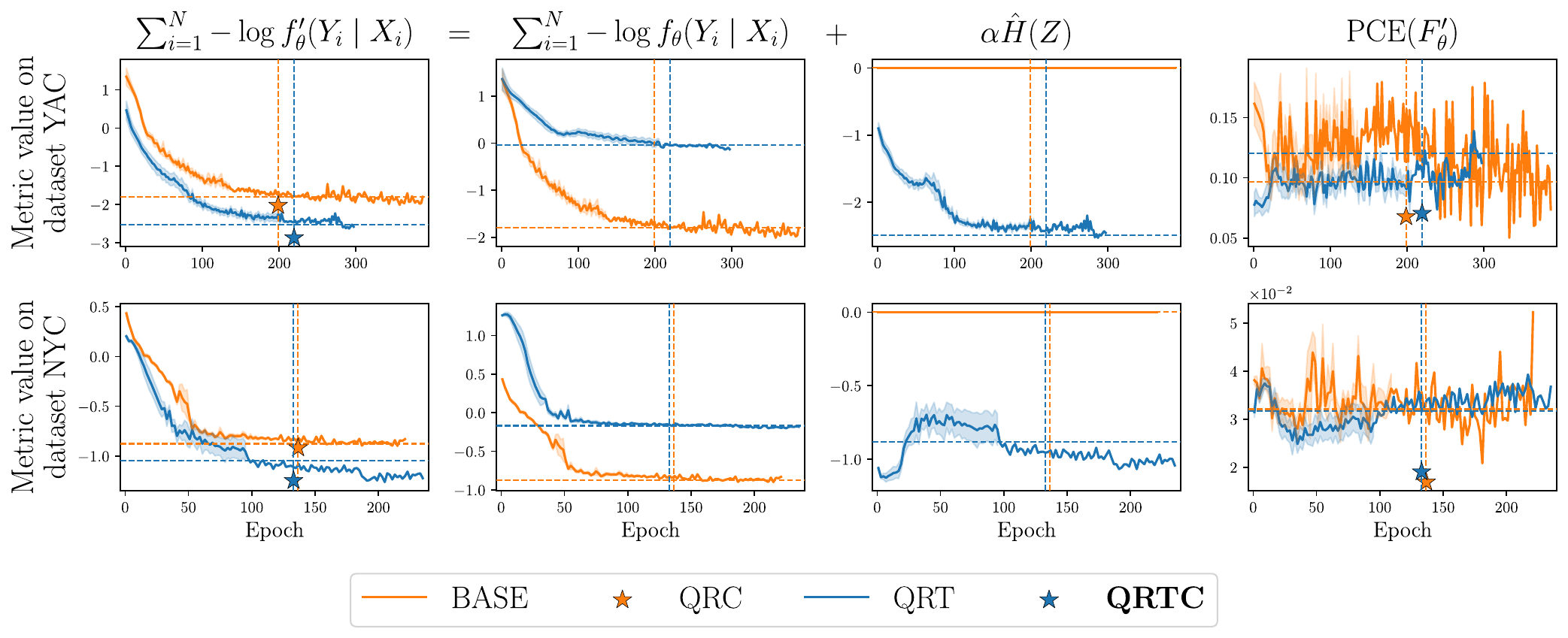}
		\caption{
			Comparison of QRT and BASE according to different metrics computed on the validation dataset.
			The three first columns show the decomposition of the NLL of QRT, where $\alpha = 1$ for QRT and $\alpha = 0$ for BASE.
			Each row represents one dataset and each column one metric.
			The training curves are averaged over 5 runs and the shaded area corresponds to one standard error.
			The vertical bars represent the epoch that was selected by early stopping (the one that minimizes the validation NLL), averaged over the 5 runs.
			The horizontal bars represent the value of the metric at the selected epoch, averaged over the 5 runs.
		}
		\label{fig:metric_per_dataset_per_epoch}
	\end{figure*}
	
	\subsection{Illustrative example}
	\label{sec:motivating_example}
	
	\cref{fig:metric_per_dataset_per_epoch} illustrates the decomposition \cref{eq:RT_decomposition}, where the NLL of $F'_\theta$ (first column) is equal to the sum of the NLL of the base model $F_\theta$ (second column) and the entropy of the PIT (third column).
	To allow a comparison between \tt{QRT} and \tt{BASE}, we alter the decomposition by introducing a coefficient $\alpha$ to the second term.
	When $\alpha = 1$, we obtain the exact decomposition of \tt{QRT}.
	When $\alpha = 0$, the first and second column are equal and correspond to the loss of \tt{BASE}.
	Metrics on this figure are computed on the validation dataset and metrics computed on the training dataset are available in \cref{sec:metrics_per_epoch}.
	The vertical bars correspond to the epoch selected by early stopping while the horizontal bars correspond to the value of the metric at the epoch selected by early stopping, on average over 5 runs.
	The stars indicate the models \tt{QRC} and \tt{QRTC}, corresponding to \tt{BASE} and \tt{QRT}, respectively, after QR on a separate calibration dataset.
	
	We can see that \texttt{QRT} achieves a lower validation NLL, suggesting improved probabilistic predictions, even though the NLL of $F_\theta$ is higher, which means that \tt{QRT} relies on the calibration map to achieve a lower NLL.
	We note that the calibration map does not introduce any additional parameters.
	In terms of calibration, we can see that the PCE of \texttt{Base} has a higher variance across the epochs compared to \tt{QRT}.
	The PCE of \texttt{QRT} is more stable during training and often lower.
	By constraining the model to be calibrated on a specific dataset at each training step, \tt{QRT} involves a form of regularization which is fundamentally different from QREG.
	
	The stars indicate that, after the post-hoc step, \tt{QRTC} still benefits from improved NLL compared to \tt{QRC}, and the PCE is improved in both cases due to the finite-sample guarantee provided by QR.
	These metrics reported on the 57 datasets that we consider in \cref{sec:experiments} are available in \cref{sec:metrics_per_epoch}, where we obtain similar observations on most datasets despite their heterogeneity.
	
	\subsection{A Unified Algorithm}
	\label{sec:RT_framework}
	
	\cref{algo:RT_framework} unifies QRT, QREG and BASE, with or without QR, where the methods only differ by the hyperparameters $\alpha$ and $C$, as indicated in \cref{table:RT_framework}.
	The hyperparameter $\alpha$, introduced in \cref{sec:motivating_example}, is a coefficient for the second term of the decomposition \cref{eq:RT_decomposition}.
	A value of $\alpha = 1$ corresponds to QRT and $\alpha = 0$ corresponds to NLL minimization without QRT.
	Tuning the hyperparameter $\alpha$ in order to minimize $\text{PCE}(F_\theta)$ corresponds to QREG with regularization strength $\lambda = -\alpha$.
	The hyperparameter $C$ controls whether the final model is recalibrated on a separate calibration dataset using QR.
	
	\SetAlgoNoLine
	\SetKwProg{Fn}{Function}{:}{}
	\SetKwFunction{CalibrationMap}{CalibrationMap}
	\SetKwInOut{Input}{Input}
	
	\begin{algorithm}
		\caption{QRT framework}
		\label{algo:RT_framework}
		\Input{Predictive CDF $F_\theta$, regularization strength $\alpha \in \R$, boolean $C$, training dataset $\D$, calibration dataset $\D'$}
		%	{\bfseries Input:} Predictive CDF $F_\theta$, training dataset $\D = \Set{(X_i, Y_i)}_{i=1}^{N}$, optional calibration dataset $\D' = \Set{(X'_i, Y'_i)}_{i=1}^{N'}$ \;
		\ForEach{minibatch $\Set{(X_i, Y_i)}_{i=1}^{B} \subseteq \D$, until early stopping}{
			Compute $Z_i \gets F_\theta(Y_i \mid X_i)$, for $i = 1, \dots, B$ \;
			Define $\phi^\text{REFL}_{\theta}$ from $Z_1, \dots, Z_B$ using \cref{eq:refl} \;
			$\mc{L}(\theta) = -\frac{1}{B} \sum_{i=1}^B \underbrace{\log f_\theta(Y_i \mid X_i) + \alpha \log \phi^\text{REFL}_\theta(Z_i)}_{\log f'_\theta(Y_i \mid X_i)}$ \;
			Update parameters $\theta$ using $\nabla_\theta \mc{L}(\theta)$ \;
		}
		\If{$C$ is True}{
			Compute $Z'_i \gets F_\theta(Y'_i \mid X'_i)$, for $i = 1, \dots, |\D'|$ \;
			Define $\Phi^\text{REFL}_{\theta}$ from $Z'_1, \dots, Z'_{|\D'|}$ \; %using \cref{eq:Phi_theta^KDE} \;
			\KwRet the predictive CDF $\Phi_\theta^\text{REFL} \circ F_\theta$ \;
		} \Else{
			\KwRet the predictive CDF $F_\theta$ \;
		}
	\end{algorithm}
	
	In \cref{algo:RT_framework}, the calibration map $\phi_\theta^\text{KDE}$ is computed at each step on the current batch, allowing QRT to simultaneously use the neural network outputs to compute the first term and the second term of the decomposition \cref{eq:RT_decomposition}.
	In \cref{sec:impact_of_cal_size}, we investigate the impact of computing the calibration map from data sampled randomly in the training dataset, which allows to compute the calibration map on a larger dataset.
	We observe that the approach in \cref{algo:RT_framework} provides similar NLL than the approach in \cref{sec:impact_of_cal_size}, while being more computationally efficient.
	In \cref{sec:inhoc_alpha}, we confirm that $\alpha = 1$ provides the best NLL compared to other values of $\alpha$.
	
	\begin{table}[h]
		\footnotesize
		\centering
		\caption{Summary of the compared methods, which differ only by the hyperparameters $\alpha$ and $C$ in \cref{algo:RT_framework}. We recommend using \tt{QRTC}.}
		\begin{tabular}{l|llllll}
			\toprule
			Method & \tt{BASE} & \tt{QRC} & \tt{QREG} & \tt{QREGC} & \tt{QRT} & \tt{QRTC} \\
			\midrule
			$\alpha$ & 0 & 0 & Tuned & Tuned & 1 & 1 \\
			$C$ & False & True & False & True & False & True \\
			\bottomrule
		\end{tabular}
		\label{table:RT_framework}
	\end{table}

	\subsection{Time complexity}
	\label{sec:time_complexity}
	
	The proposed method can introduce increased computational demand due to evaluating $\log \phi^\text{KDE}_\theta(Z_i)$, which results in $O(B^2)$ evaluations of $f_\text{Log}$ per minibatch, where $B$ is the batch size ($B = 512$ in our experiments).
	More precisely, $3B^2$ evaluations of $f_\text{Log}$ are performed due to using the estimator $\phi_\theta^\text{REFL}$ (see \cref{sec:kde_finite}).
	This additional computational demand does not depend on the size of the underlying neural network and hence becomes less significant when training highly computationally demanding models.
	In practice, we observe the time per minibatch to be nearly two-fold compared to a method without QR, as detailed in \cref{sec:time_measurements}.
	
	%\subsection{On the opposite objectives of minimizing the PCE of $F_\theta$ and minimizing the NLL of $F'_\theta$}
	
		\section{RELATED WORK}
	\label{sec:related_work}
	
	%\textbf{Strictly proper scoring rules} provide evaluation and estimation methods \citep{Gneiting2007-hb} that encourage to be honest.
	%The importance of using strictly proper scoring rules, both in classification \citep{Murphy1973-vi} and regression \citep{Matheson1976-lc}, has been noted in the literature \citep{Brocker2007-ps}.
	
	\paragraph{Post-hoc calibration methods}
	%Since the finding that neural networks are not calibrated 
	Many post-hoc calibration methods have been proposed for classification problems \citep{Kumar2019-lm,Gupta2020-oi}. The most popular one is called temperature scaling \citep{Guo2017-ow} and has the useful property of preserving accuracy. Conformal prediction, pioneered by \citet{Vovk2005-ib}, is an attractive approach due to the finite-sample coverage guarantee that it provides.
	In regression, multiple approaches based on conformal prediction have been proposed, including Conformal Quantile Regression \citep{Romano2019-kp} and Distributional Conformal Prediction \citep{Chernozhukov2021-sg}.
	Quantile Recalibration \citep{Kuleshov2018-tb} is another method which transforms predictive distributions using a recalibration map, and has been shown to be closely related to Distributional Conformal Prediction \citep{Dheur2023-bo}.
	Finally, methods have been proposed to target a stronger notion of calibration, called distribution calibration \citep{Song2019-bk,Kuleshov2022-pv}.
	
	\paragraph{Regularization methods} Regularization-based calibration methods aim to improve calibration during training, e.g. using ensembling \citep{Lakshminarayanan2017-zg}, mixup \citep{Zhang2018-cw}, label smoothing \citep{Muller2019-gr}, or penalizing high confidence predictions \citep{Pereyra2017-lk}.
	Multiple regularization objectives have been proposed in classification \citep{Kumar2018-cw,Karandikar2021-ci,Popordanoska2022-ip,Yoon2023-ds} and regression \citep{Pearce2018-lo,Utpala2020-nw,Chung2021-rh,Dheur2023-bo}.
	While these methods allow to improve calibration, they may negatively impact  other accuracy metrics. In fact, \citet{Karandikar2021-ci} and \citet{Yoon2023-ds} reported selecting hyperparameters that minimize the expected calibration error while decreasing the accuracy by about 1\%. Similarly, \citet{Dheur2023-bo} selected the regularization factor $\lambda$ that minimizes the PCE with a maximum CRPS increase of 10\%. In contrast to these methods, Recalibration Training did not impose such a trade-off in our large-scale experiment and resulted in both improved NLL and PCE.

	\paragraph{Towards unifying model training and post-hoc calibration}
	
	Despite the potential benefits of combining post-hoc and regularization strategies to improve calibration, empirical evidence from both classification \citep{Wang2021-jy} and regression \citep{Dheur2023-bo} contexts has indicated that frequently utilized regularization methods result in neural networks that are less calibratable. The method we propose is consistent with the recommendation made by \citet{Wang2021-jy} to regard model training and post-hoc calibration as a unified framework. Finally, recent works in classification \citep{Stutz2022-uh,Einbinder2022-pm} proposed integrating conformal prediction into neural network training. The outcome is precise coverage with smaller prediction sets.
	
	\section{A LARGE-SCALE EXPERIMENTAL STUDY}
	\label{sec:experiments}
	
	We compare the performance of \tt{QRTC} (\cref{sec:recalibration_training}) against \tt{BASE}, \tt{QRC} and \tt{QREG} on several metrics in a large-scale experiment. We also consider multiple ablated versions of \tt{QRTC}. 
	We build on the large-scale empirical study of \citet{Dheur2023-bo} \footnote{\url{https://github.com/Vekteur/probabilistic-calibration-study}} and consider the same underlying neural network architectures, datasets and metrics.
	For these experiments, 81926 models were trained during a total of 180 hours on 40 CPUs.
	
	\begin{figure*}[t]
		\centering
		\includegraphics[width=\linewidth]{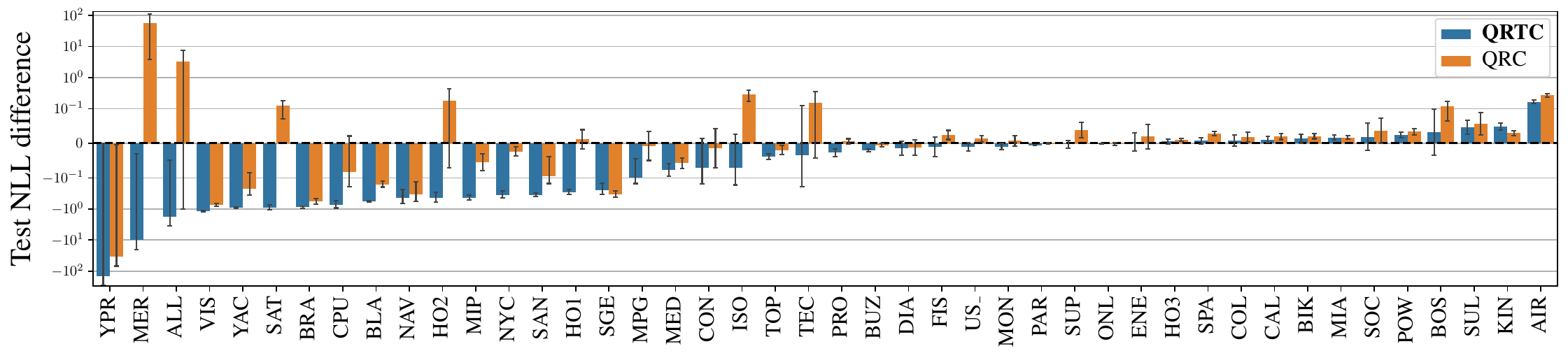}
		\caption{
			Difference in test NLL between two post-hoc methods (\tt{QRTC} and \tt{QRC}) and \tt{BASE}, where negative values indicate an improvement compared to \tt{BASE}, averaged over 5 runs with error bars corresponding to one standard error.
			We observe that \tt{QRTC} achieves a lower NLL than \tt{BASE} and \tt{QRC} on most datasets.
			Note that, for \tt{BASE}, $F_\theta$ is trained with a larger dataset that includes the calibration data of \tt{QRTC} and \tt{QRC}.
			The experimental setup is described in \cref{sec:experiments}}
		\label{fig:some/without_discrete/diff/test_nll}
		\vspace{-0.5cm}
	\end{figure*}
	
	\subsection{Benchmark datasets}
	\label{sec:benchmark_datasets}
	
	In our study, we analyze a total of 57 data sets, including 27 from the recently curated benchmark by OpenML \citep{Grinsztajn2022-nu}, 18 obtained from the AutoML Repository \citep{Gijsbers2019-xk}, and 12 from the UCI Machine Learning Repository \citep{Dua2017-ut}. We divide each dataset into four sets: training (65\%), validation (10\%), calibration (15\%), and test (10\%). To ensure robustness, we repeat this partitioning five times randomly and then average the results. During the training process, we normalize both the features, $X$, and the target, $Y$, using their respective means and standard deviations derived from the training set. After obtaining predictions, we transform them back to the original scale. For all methods, we use early stopping (with a patience of 30) to choose the epoch that gives the smallest validation NLL.

%	Although we consider regression benchmarks, we observe that certain datasets present a high level of discreteness.
%	TODO: ref Appendix and maybe give more details.
%	We propose to identify the datasets that are the most discrete using the proportions of values $Y$ in the dataset that are among the 10 most frequent values.
%	For example, if a dataset only contains 10 distinct values, this proportion would be 100\%.
%	Such dataset could be approached using classification, although information about the order of the targets $Y$ would be lost.
%	\cref{table:datasets} in the Supplementary Material shows that 13 out of 57 datasets have more than half of their target values among the 10 most frequent ones. These datasets appear in all 4 benchmark suites.

	To avoid a potential bias in our analysis, we exclude certain datasets that could not be suited for regression.
	We identify these datasets using the proportion of targets $Y$ that are among the 10 most frequent values in the dataset, and we call this proportion the level of discreteness.
	\cref{table:datasets} in the Supplementary Material shows that 13 out of 57 datasets have a level of discreteness above 0.5 and these datasets appear in all 4 benchmark suites.
	\tt{QRTC} was able to perform better on these datasets, as discussed in \cref{sec:results/with_discrete} where full results are available.

	\subsection{Experimental setup}
	\label{sec:experimental_setup}
	
	\paragraph{Base neural network model} The base model $F_\theta$ is a mixture of $K = 3$ Gaussians, where the means $\mu_k(X)$, standard deviations $\sigma_k(X)$, and weights $w_k(X)$, for each component $k = 1, ..., K$ are obtained as outputs of a hypernetwork, which is a 3-layer MLP with 128 hidden units per layer.
	We also consider other base models in \cref{sec:base_models}.
	
	\paragraph{Compared methods}
	
	We compare all methods in \cref{table:RT_framework} where QR is applied, namely \tt{QRC}, \tt{QRTC} and \tt{QREGC}.
	We also compare \tt{BASE} as a baseline method, where, to ensure a fair comparison, the calibration dataset is used as additional training data for the base model $F_\theta$ since there is no need for a calibration dataset.
	For \tt{QRTC} and \tt{QREGC}, the bandwidth $b$ of $\Phi_\theta^\text{KDE}$ is selected by minimizing the validation NLL in the set $\Set{0.01, 0.05, 0.1, 0.2}$.
	\cref{sec:impact_of_inhoc_b} shows that QRT does not require extensive tuning of the hyperparameter $b$. In fact, good results are already obtained with a default value of $b=0.1$.
	For \tt{QREGC}, we select $\lambda = -\alpha$ where $\lambda \in \Set{0, 0.01, 0.05, 0.2, 1, 5}$ and minimizes PCE with a maximum increase in continuous ranked probability score (CRPS) of 10\% in the validation set, as in \cite{Dheur2023-bo}.
	  Since none of the compared methods introduce additional parameters compared to the baseline, all methods estimate parameters in the same space $\Theta$.
	
	\paragraph{Metrics}
	We evaluate probabilistic predictions using the NLL and CRPS, which are strictly proper scoring rules.
	Probabilistic calibration is measured using PCE \eqref{eq:probabilistic_calibration}. Finally, we measure sharpness using the mean standard deviation of the predictions, denoted by SD.
	
	\begin{figure*}[t]
		\centering
		\subcaptionbox{
			Letter-value plots showing Cohen's d for different metrics with respect to \texttt{Base}.
			\label{fig:some/without_discrete/cohen_d}
		}{
			\includegraphics[width=\linewidth]{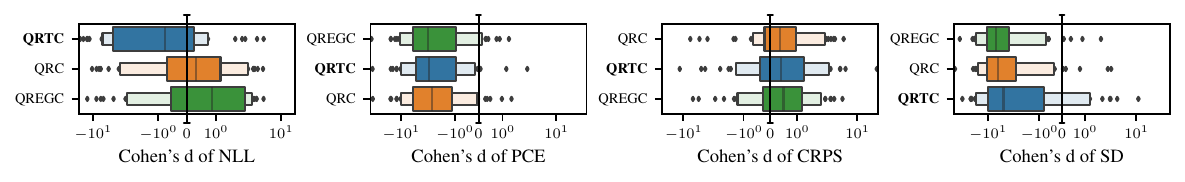}
			\vspace{-0.3cm}
		}
		\subcaptionbox{
			Critical difference diagrams for different metrics.
			\label{fig:some/without_discrete/cd_diagrams}
		}{
			\includegraphics[width=\linewidth/4]{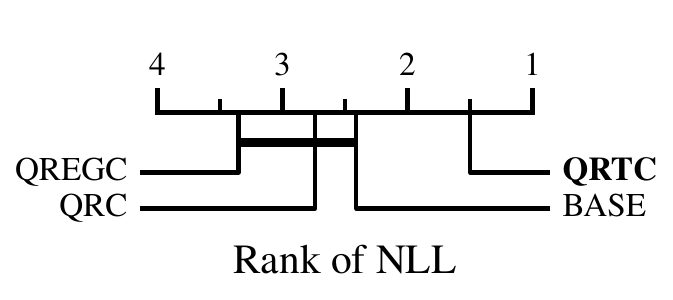}
			\includegraphics[width=\linewidth/4]{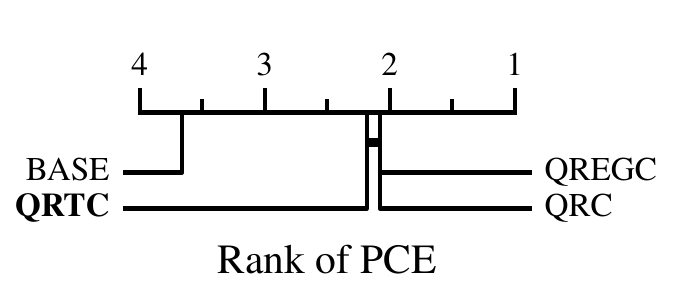}
			\includegraphics[width=\linewidth/4]{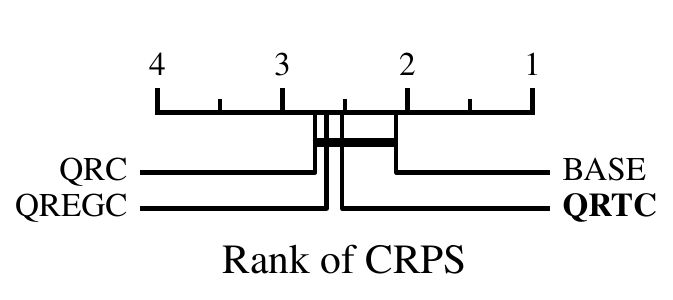}
			\includegraphics[width=\linewidth/4]{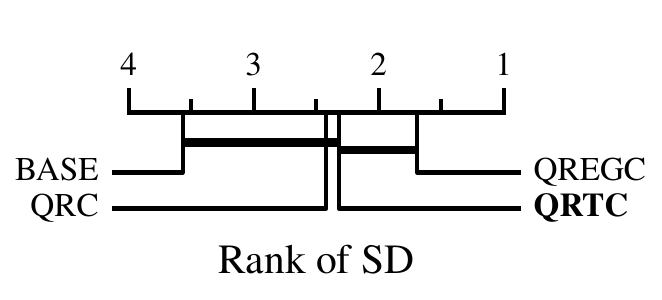}
			\vspace{-0.2cm}
		}
		\caption{Comparison of \tt{QRTC}, \tt{QRC}, \tt{QREGC} and \tt{BASE}, as detailed in \cref{sec:experiments}.}
		\vspace{-0.5cm}
		\label{fig:some/without_discrete}
	\end{figure*}
	
	\paragraph{Comparison of multiple models over many datasets} Given the different scales of NLL, CRPS, and SD across datasets, we report Cohen's d, a standardized effect size metric to compare the mean performance of a method against a baseline. Cohen's d values of $-0.8$ and $-2$ are regarded as large and huge effect sizes, respectively. Owing to the diverse nature of the datasets used in our study, the performance metrics of our models can exhibit substantial variations. To effectively illustrate the results, we employ letter-value plots to depict the distribution of Cohen's d. These plots highlight the quantiles at levels $\nicefrac{1}{8}$, $\nicefrac{1}{4}$, $\nicefrac{1}{2}$, $\nicefrac{3}{4}$ and $\nicefrac{7}{8}$, as well as any outliers. A median value below zero indicates an improvement in the metric across more than half of the datasets by the model.
    Letter-value plots are ordered based on the median value to facilitate an easy identification of the top-performing methods.
	
	In order to determine if there's a significant difference in model performance, we first apply the Friedman test \citep{Friedman1940-vi}. Subsequently, we carry out a pairwise post-hoc analysis, as advocated by \citet{Benavoli2016-mh}, using a Wilcoxon signed-rank test \citep{Wilcoxon1945-cp} complemented by Holm's alpha correction \citep{Holm1979-rk}. These findings are represented by a critical difference diagram \citep{Demsar2006-ed}. The lower the rank (further to the right), the superior the model's performance. A thick horizontal line illustrates a set of models with statistically indistinguishable performance, at a significance level of 0.05.
	
	\subsection{Results}
	
	\cref{fig:some/without_discrete/diff/test_nll} illustrates the comparison in NLL of \tt{QRTC} and \tt{QRC} across various datasets, relative to \tt{BASE}. We observe that \tt{QRTC} consistently achieves a lower NLL on the majority of the datasets. This suggests that allowing the model to adapt to the calibration map during training improves the final predictive accuracy, without the need for extra parameters.
	
	\cref{fig:some/without_discrete} shows the letter-values plots for Cohen’s d of different metrics (top panel) as well as the associated critical difference diagram (bottom panel), for all methods and datasets. The reference model is \texttt{BASE}. %Green, blue, and red colors are used for Recalibration Training, quantile calibration, and regularization, respectively.
	\cref{fig:some/without_discrete/cd_diagrams} shows that our proposed method, \texttt{QRTC}, is able to significantly outperform the baseline and other methods in terms of test NLL, as suggested by \cref{fig:some/without_discrete/diff/test_nll}.
	In terms of PCE, since all considered methods except \tt{BASE} are combined with QR, they benefit from the finite sample guarantee \cref{eq:calibration_guarantee} and achieve a similar PCE, outperforming \tt{BASE}.
	
	We also observe that there is no significant difference in terms of the CRPS of \tt{QRTC} compared to other methods.
	This suggests that QRT is able to place a high density at the observed/realized test data points while the characteristics measured by CRPS, a distance-sensitive scoring rule \citep{Du2021-zg}, are not significantly impacted.
	Furthermore, all methods produce sharper predictions than \tt{BASE}, suggesting that \tt{BASE} is underconfident, despite achieving similar NLL than \tt{QRC}.
	
	\subsection{The importance of the base model}
	\label{sec:importance_base_model}
	
	We note that previous studies on calibration have often focused on single Gaussian predictions with a small number of layers \citep{Lakshminarayanan2017-zg,Utpala2020-nw,Zhao2020-ze}.
	These models have been outperformed in terms of NLL and CRPS by mixture predictions \citep{Dheur2023-bo}.
	Following \cite{Dheur2023-bo}, we consider a 3-layer MLP that predicts a mixture of 3 Gaussians. 
	
	To further understand the role of the flexibility of the base model, we consider a 3-layer MLP with varying number of components in the mixture as well as a ResNet.
	We observe that, in all scenarios, \tt{QRTC} outperforms \tt{QRC} on most datasets in terms of NLL.
	Moreover, the enhancement is most pronounced in the case of misspecified single Gaussian mixture predictions.
	Detailed results are available in \cref{sec:base_models}.
	%TODO: link with \cite{Minderer2021-xw} is still not clear?
	
	\section{AN ABLATION STUDY AND ANALYSIS OF QUANTILE RECALIBRATION TRAINING}
	\label{sec:other_experiments}
	
	\subsection{Ablation study}
	\label{sec:ablation_study}

	\begin{figure*}[t]
		\centering
		\subcaptionbox{
			Letter-value plots showing Cohen's d for different metrics with respect to \texttt{Base}.
			\label{fig:ablation/cohen_d}
		}{
			\includegraphics[width=\linewidth]{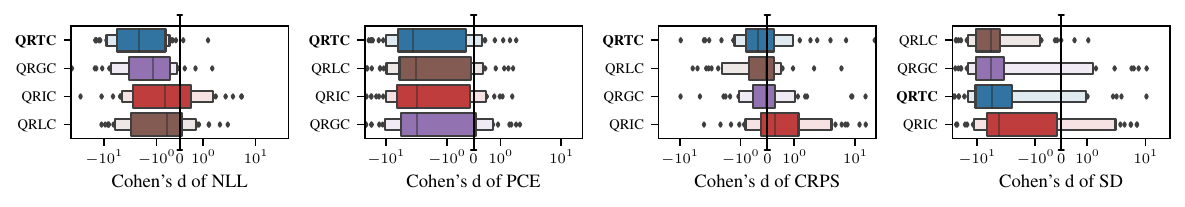}
			\vspace{-0.3cm}
		}
		\subcaptionbox{
			CD diagrams
		}{
			\includegraphics[width=\linewidth/4]{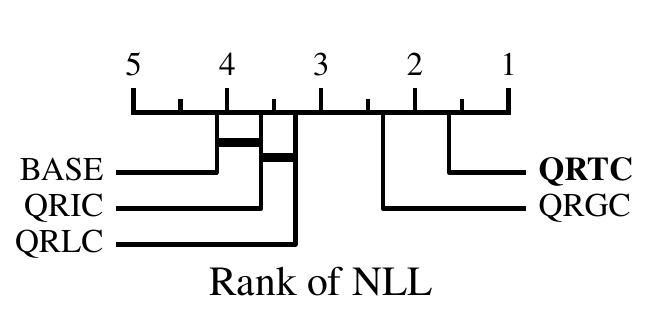}
			\includegraphics[width=\linewidth/4]{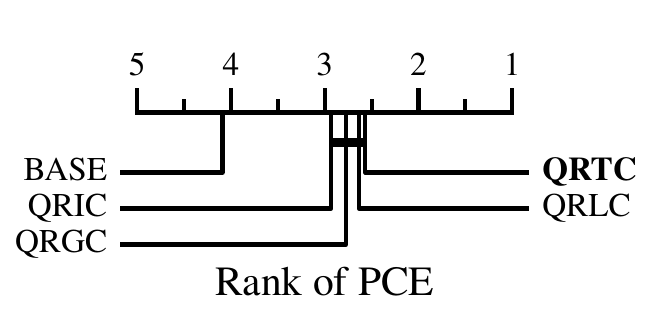}
			\includegraphics[width=\linewidth/4]{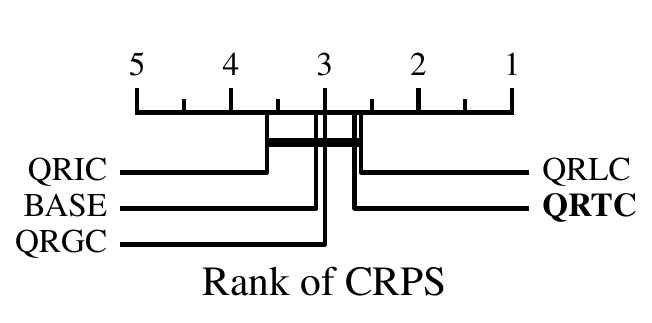}
			\includegraphics[width=\linewidth/4]{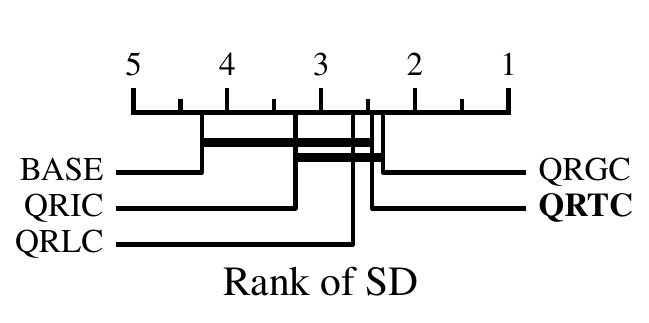}
			\vspace{-0.3cm}
		}
		\caption{Comparison of \tt{QRTC}, \tt{QRGC}, \tt{QRIC}, \tt{QRLC} and \tt{BASE} as detailed in \cref{sec:ablation_study}.}
		\vspace{-0.3cm}
		\label{fig:ablation}
	\end{figure*}
	
	In addition to the methods compared above, we provide an ablation study in order to understand the importance of the different components of QRT.
	We consider three ablated versions of QRT that differ from \texttt{QRTC} by one aspect each.
	
	\texttt{QRIC}, for \textit{QRT at initialization only}, corresponds to \texttt{QRTC} except that the calibration map is computed once before the first training step and is fixed during the rest of training (except for the last post-hoc step).
	The goal is to show that improved initialization is not the only strength of QRT.
	
	%Before predicting on the test set, the calibration map is computed once again on the calibration dataset.
	\texttt{QRGC}, for \textit{QRT without gradient backpropagation}, corresponds to \texttt{QRTC} except that backpropagation does not occur on the computation graph generated by the calibration map, i.e., when computing $Z'_i$ in \cref{algo:RT_framework}.
	While \texttt{QRTC} considers the calibration map as part of the model, \texttt{QRGC} considers the calibration map as an external actor that modifies the predictions at each step.
	The goal is to show that merely applying QR at each training step is not sufficient unless it is considered an integral part of the model.
	
	\texttt{QRLC}, for \textit{QRT with learned recalibration map}, corresponds to \texttt{QRTC} except that the PITs $Z_i$ in \cref{algo:RT_framework} are replaced by additional learned parameters initialized uniformly between 0 and 1.
	Thus, \texttt{QRLC} possesses $B$ more parameters than \texttt{QRTC}.
	The goal is to show that the benefits of QRT are not only due to the additional flexibility provided by the calibration map.
	
	\cref{fig:ablation} shows a comparison of these ablated versions of \texttt{QRTC} against \texttt{QRTC}.
	We observe that all ablated versions result in significantly decreased NLL compared to \texttt{QRTC}, highlighting the strengths of the different components of QRT.
	Additionally, the CRPS and PCE show no improvement compared to \tt{QRTC}, and all ablated versions result in slightly sharper predictions than \tt{BASE}.

	\section{CONCLUSION}
	
	We introduced Quantile Recalibration Training (QRT), a novel method that produces predictive distributions that are probabilistically calibrated by design at each training step. We demonstrated the effectiveness of this approach through a large-scale experiment and an ablation study. Our results indicate that QRT demonstrates enhanced performance in both predictive accuracy (NLL) and calibration compared to the baseline. Compared to Quantile Recalibration, QRT achieves a similar calibration improvement with an additional enhancement in NLL. This combination presents a compelling option to produce predictive distributions that are both accurate and well-calibrated. We also discussed the issue of training regression models on datasets with a discrete output variable. For future work, we suggest extending our method to encompass other calibration notions, such as distribution calibration \citep{Song2019-bk}. Additionally, integrating other calibration methods, such as Conformal Quantile Regression \citep{Romano2019-kp}, into the training procedure is an interesting direction to explore.
	
	\printbibliography
	
	%%%%%%%%%%%%%%%%%%%%%%%%%%%%%%%%%%%%%%%%%%%%%%%%%%%%%%%%%%%%
	
	\section*{Checklist}
	
	\begin{enumerate}
		
		\item For all models and algorithms presented, check if you include:
		\begin{enumerate}
			\item A clear description of the mathematical setting, assumptions, algorithm, and/or model. Yes.
			\item An analysis of the properties and complexity (time, space, sample size) of any algorithm. Yes.
			\item (Optional) Anonymized source code, with specification of all dependencies, including external libraries. Yes.
		\end{enumerate}

		\item For any theoretical claim, check if you include:
		\begin{enumerate}
			\item Statements of the full set of assumptions of all theoretical results. Yes.
			\item Complete proofs of all theoretical results. Yes.
			\item Clear explanations of any assumptions. Yes.
		\end{enumerate}

		\item For all figures and tables that present empirical results, check if you include:
		\begin{enumerate}
			\item The code, data, and instructions needed to reproduce the main experimental results (either in the supplemental material or as a URL). Yes.
			\item All the training details (e.g., data splits, hyperparameters, how they were chosen). Yes.
			\item A clear definition of the specific measure or statistics and error bars (e.g., with respect to the random seed after running experiments multiple times). Yes.
			\item A description of the computing infrastructure used. (e.g., type of GPUs, internal cluster, or cloud provider). Yes.
		\end{enumerate}
		
		\item If you are using existing assets (e.g., code, data, models) or curating/releasing new assets, check if you include:
		\begin{enumerate}
			\item Citations of the creator If your work uses existing assets. Yes.
			\item The license information of the assets, if applicable. Yes.
			\item New assets either in the supplemental material or as a URL, if applicable. Yes.
			\item Information about consent from data providers/curators. Not Applicable.
			\item Discussion of sensible content if applicable, e.g., personally identifiable information or offensive content. Not Applicable.
		\end{enumerate}
		
		\item If you used crowdsourcing or conducted research with human subjects, check if you include:
		\begin{enumerate}
			\item The full text of instructions given to participants and screenshots. Not Applicable.
			\item Descriptions of potential participant risks, with links to Institutional Review Board (IRB) approvals if applicable. Not Applicable.
			\item The estimated hourly wage paid to participants and the total amount spent on participant compensation. Not Applicable.
		\end{enumerate}
		
	\end{enumerate}
	
	\addtocounter{figure}{-1}
	\refstepcounter{figure}\label{LASTFIGURE}
	\addtocounter{table}{-1}
	\refstepcounter{table}\label{LASTTABLE}
	\addtocounter{AlgoLine}{-1}
	\refstepcounter{AlgoLine}\label{LASTALGORITHM}

\onecolumn
\appendix

\section{RESULTS ON DIFFERENT BASE MODELS}
\label{sec:base_models}

\begin{figure}[H]
	\centering
	\subcaptionbox{
		Difference of test NLL compared to \tt{BASE}.
		\label{fig:some_mixture_1/diff/test_nll}
	}{
		\includegraphics[width=\linewidth]{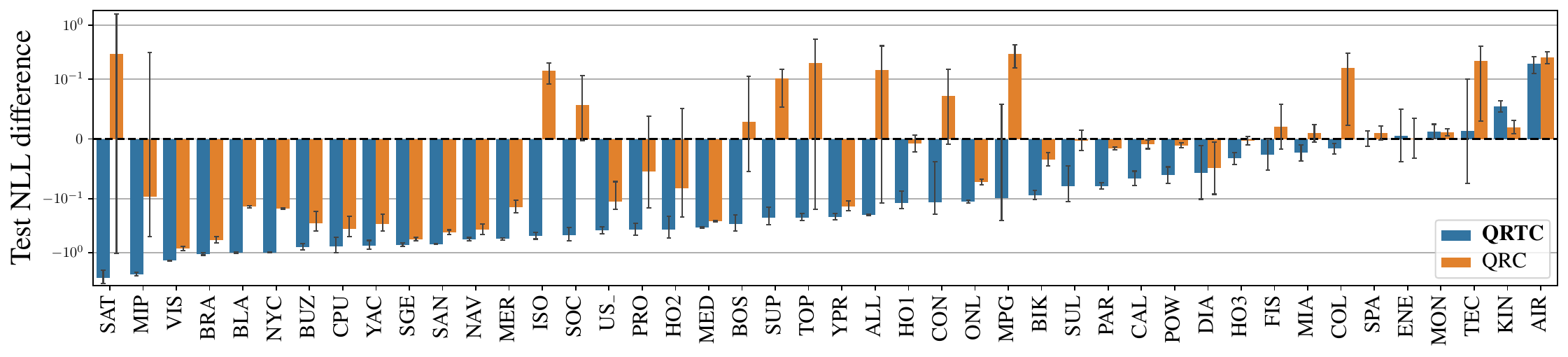}
		\vspace{-0.3cm}
	}
	\subcaptionbox{
		Letter-value plots showing Cohen's d for different metrics with respect to \texttt{BASE}.
	}{
		\includegraphics[width=\linewidth]{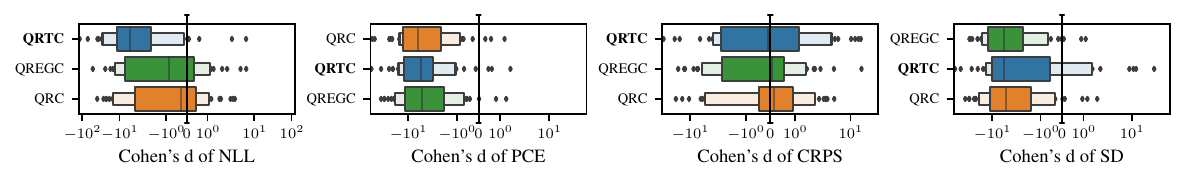}
		\vspace{-0.3cm}
	}
	\subcaptionbox{
		CD diagrams
	}{
		\includegraphics[width=\linewidth/4]{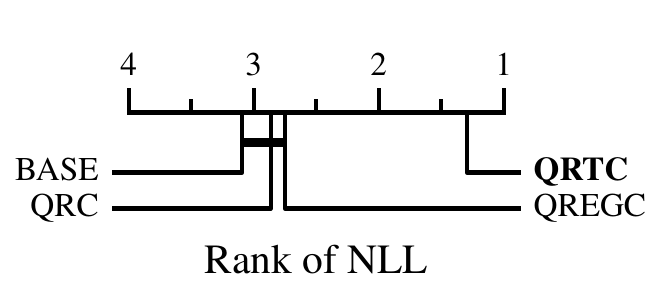}
		\includegraphics[width=\linewidth/4]{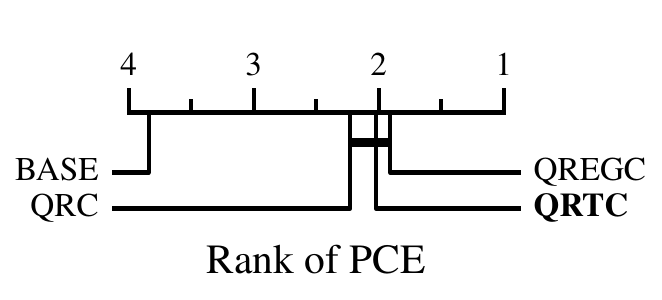}
		\includegraphics[width=\linewidth/4]{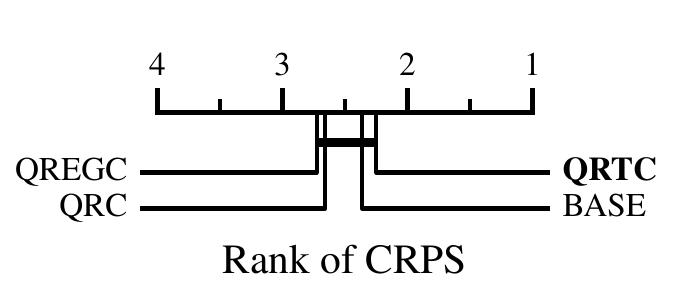}
		\includegraphics[width=\linewidth/4]{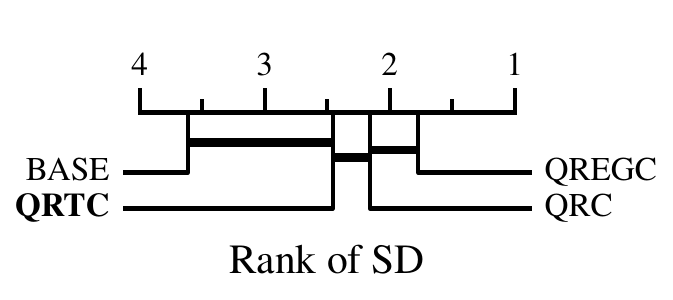}
		\vspace{-0.3cm}
	}
	\caption{Same setup than the main experiments (\cref{fig:some/without_discrete} in the main text), except that the underlying neural networks produces a single Gaussian instead of a mixture of 3 Gaussians.}
	\label{fig:some_mixture_1/without_discrete}
\end{figure}

\begin{figure}[H]
	\centering
	\subcaptionbox{
		Difference of test NLL compared to \tt{BASE}.
		\label{fig:some_mixture_10/diff/test_nll}
	}{
		\includegraphics[width=\linewidth]{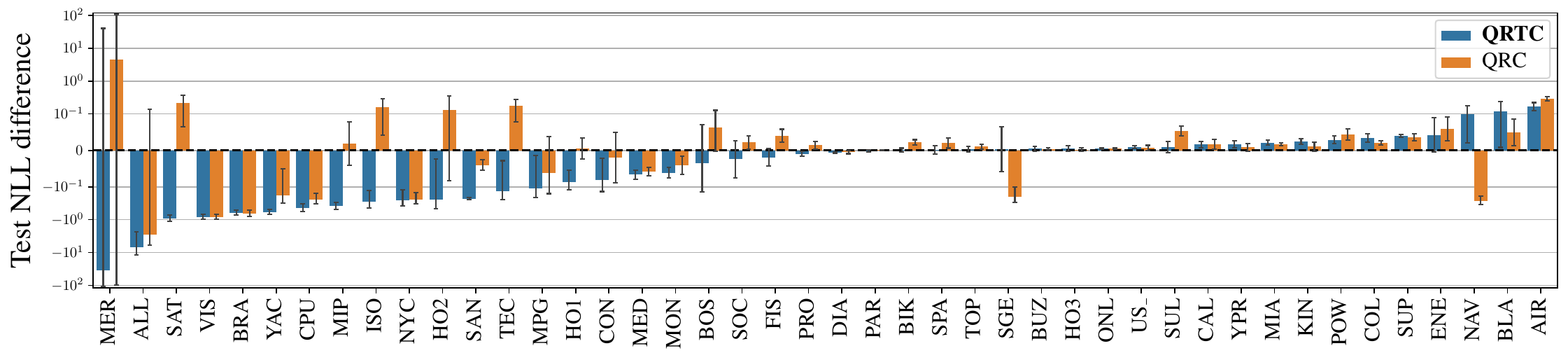}
		\vspace{-0.3cm}
	}
	\subcaptionbox{
		Letter-value plots showing Cohen's d for different metrics with respect to \texttt{BASE}.
	}{
		\includegraphics[width=\linewidth]{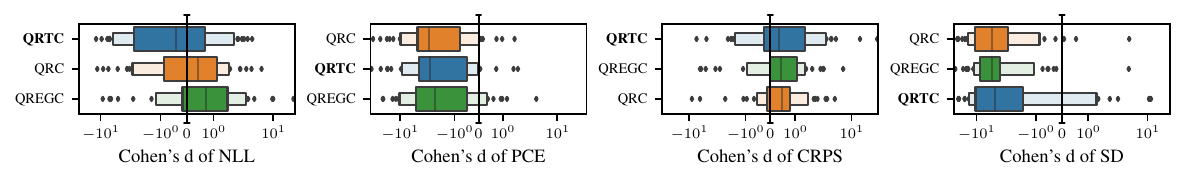}
		\vspace{-0.3cm}
	}
	\subcaptionbox{
		CD diagrams
	}{
		\includegraphics[width=\linewidth/4]{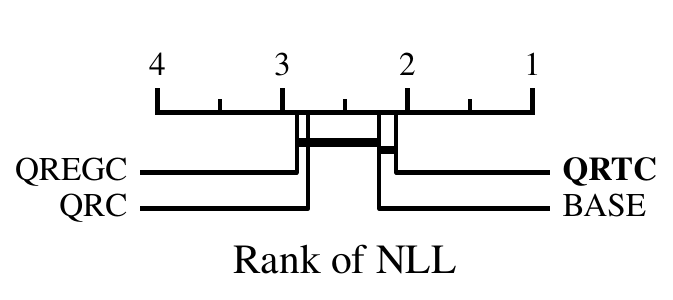}
		\includegraphics[width=\linewidth/4]{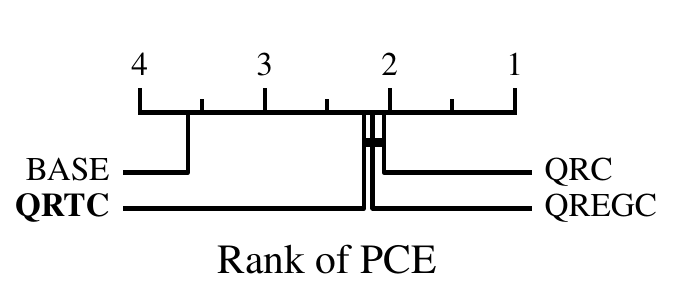}
		\includegraphics[width=\linewidth/4]{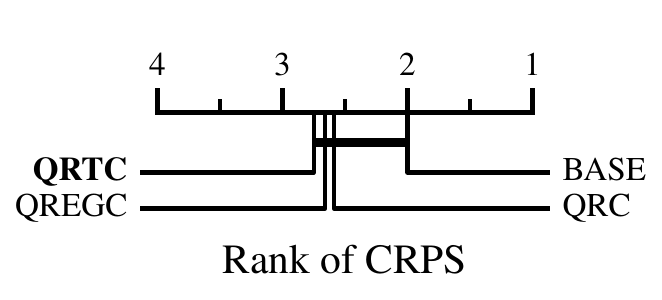}
		\includegraphics[width=\linewidth/4]{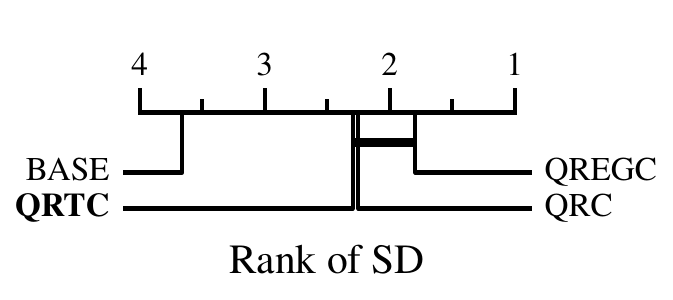}
		\vspace{-0.3cm}
	}
	\vspace{-0.3cm}
	\caption{Same setup than the main experiments (\cref{fig:some/without_discrete} in the main text), except that the underlying neural networks produces a mixture of 10 Gaussians instead of a mixture of 3 Gaussians.}
	\label{fig:some_mixture_10/without_discrete}
\end{figure}

\begin{figure}[H]
	\centering
	\subcaptionbox{
		Difference of test NLL compared to \tt{BASE}.
		\label{fig:some_mixture_3_resnet/diff/test_nll}
	}{
		\includegraphics[width=\linewidth]{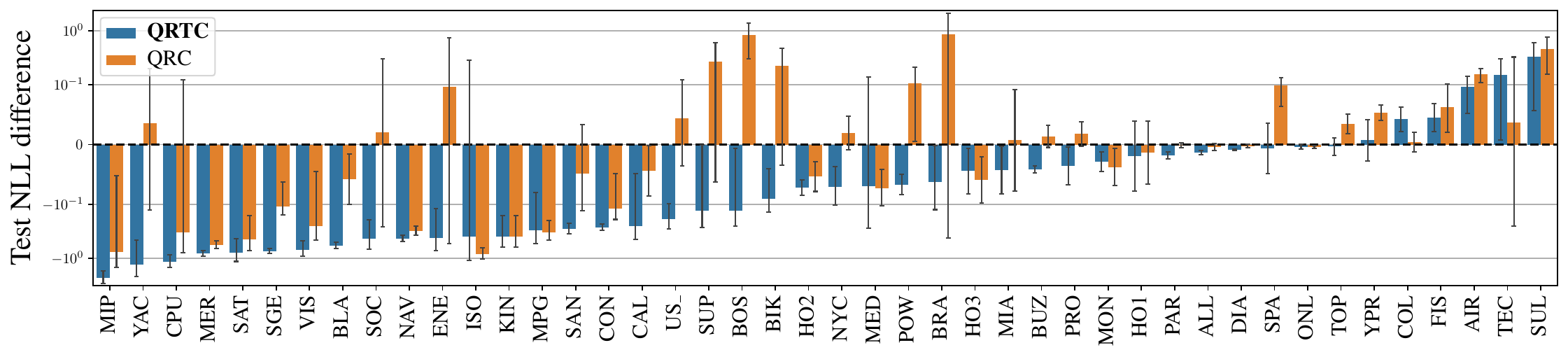}
		\vspace{-0.3cm}
	}
	\subcaptionbox{
		Letter-value plots showing Cohen's d for different metrics with respect to \texttt{BASE}.
	}{
		\includegraphics[width=\linewidth]{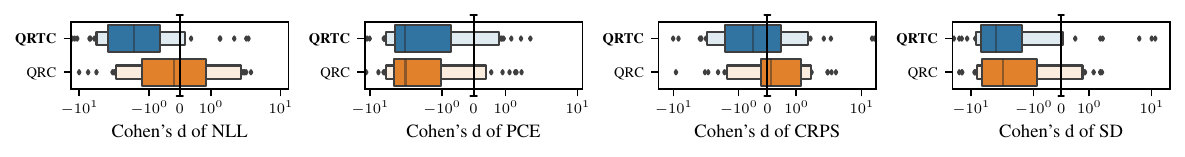}
		\vspace{-0.3cm}
	}
	\subcaptionbox{
		CD diagrams
	}{
		\includegraphics[width=\linewidth/4]{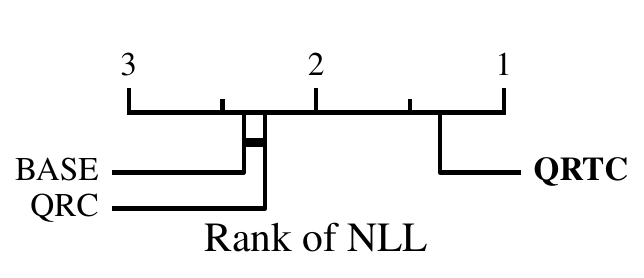}
		\includegraphics[width=\linewidth/4]{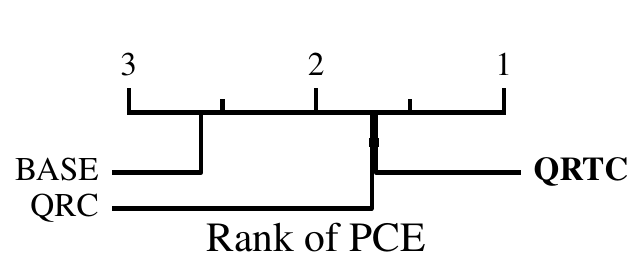}
		\includegraphics[width=\linewidth/4]{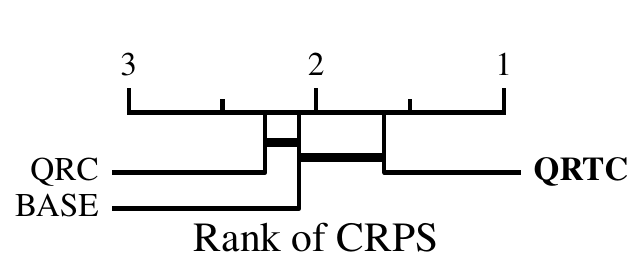}
		\includegraphics[width=\linewidth/4]{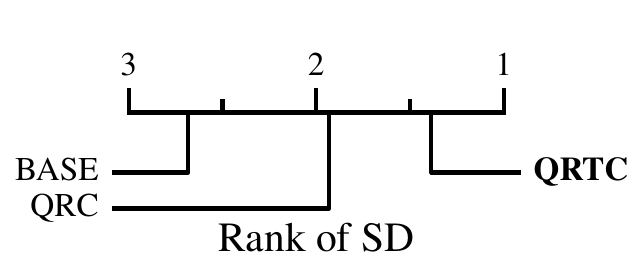}
		\vspace{-0.3cm}
	}
	\vspace{-0.3cm}
	\caption{Same setup than the main experiments (\cref{fig:some/without_discrete} in the main text), except that the underlying neural networks is a ResNet.}
	\label{fig:some_mixture_3_resnet/without_discrete}
\end{figure}

We present detailed results on the significance of the base model in influencing the performance of QRT, as discussed in \cref{sec:importance_base_model} of the main paper. While our primary experiments utilize a 3-layer MLP predicting a mixture of three Gaussians, we also explore both less flexible mixtures with a single Gaussian and more flexible mixtures comprising ten Gaussians. Additionally, we evaluate a neural network adopting a ResNet-like architecture, referred to as ResNet.
In \cref{fig:some_mixture_1/without_discrete,fig:some_mixture_10/without_discrete,fig:some_mixture_3_resnet/without_discrete}, we follow the exact same setup as in the main experiments except that the underlying neural network is modified.

\cref{fig:some_mixture_1/without_discrete} presents the results where the neural network is 3-layer MLP predicting a single Gaussian (i.e., one mean and one standard deviation).
In this misspecified case, we observe on \cref{fig:some_mixture_1/diff/test_nll} that, on many datasets, both \tt{QRTC} and \tt{QRC} provide an improvement in NLL compared to \tt{BASE}, despite \tt{BASE} having access to the calibration data.
Moreover, \tt{QRTC} provides an improvement in NLL compared to \tt{QRC} in almost all cases.
As in the main experiments, \tt{QRTC}, \tt{QRC} and \tt{QREG} are all able to provide a significant improvement in PCE compared to \tt{BASE}, with no significant difference between these three post-hoc models.
There is also no significant difference in CRPS.

\cref{fig:some_mixture_10/without_discrete} shows the same experiment except that the underlying neural network produces a mixture of 10 Gaussians (i.e., 10 means, 10 standard deviations, and 10 weights for each mixture component), offering high flexibility.
In this case, \tt{QRTC} provides an improvement in NLL in slightly more than half the datasets and the improvement is not significant, in contrast to the case of mixtures of size one and three.
However, if we compare the post-hoc models, \tt{QRTC} is still significantly better than \tt{QRC} and \tt{QREGC} in terms of NLL while achieving a similar PCE as \tt{QRC} and \tt{QREGC}.
In terms of CRPS, \tt{BASE} is slightly better than the post-hoc methods, but not significantly, which could be explained by the fact that the training dataset of \tt{BASE} also contains the calibration data.
All post-hoc methods achieve a similar CRPS.
Finally, all post-hoc methods are significantly sharper than \tt{BASE}, with \tt{QREG} being slightly sharper than \tt{QRTC} and \tt{QRC}.

\cref{fig:some_mixture_3_resnet/without_discrete} shows the same experiments except that the model is a ResNet-like architecture predicting a mixture of size 3, with 18 fully-connected hidden layers in total.
The architecture was proposed by \cite{Gorishniy2021-wt} and implemented with the default hyperparameters of \cite{Grinsztajn2022-nu}.
The architecture from \cite{Gorishniy2021-wt} is reproduced here for completeness:

\begin{align*}
	\tt{ResNet}(x) &= \tt{Prediction}(\tt{ResNetBlock}(\dots \tt{ResNetBlock}(\tt{Linear}(x)))) \\
	\tt{ResNetBlock}(x) &= x + \tt{Dropout}(\tt{Linear}(\tt{Dropout}(\tt{ReLU}(\tt{Linear}(\tt{BatchNorm}(x)))))) \\
	\tt{Prediction}(x) &= \tt{Linear}(\tt{ReLU}(\tt{BatchNorm}(x)))
\end{align*}

Since we predict a mixture of size $K = 3$, $\tt{Output}(x)$ is of dimension $K * 3 = 9$.
As for our MLP model, $\tt{Output}(x)$ is split into $\mu(x)$, $\rho(x)$ and $l(x)$.
Then, we define $\sigma(x) = \texttt{Softplus}(\rho(x))$ and $w(x) = \tt{Softmax}(l(x))$.
Finally, the mixture is defined as:
\begin{equation*}
	f_\theta(y \mid x) = \sum_{k=1}^K w_k(x) \mc{N}(y; \mu_k(x), \sigma_k^2(x)),
\end{equation*}
where $\mc{N}(y; \mu, \sigma^2)$ is the density of a normal distribution with mean $\mu$ and standard deviation $\sigma$ evaluated at $y$.

\cref{fig:some_mixture_3_resnet/diff/test_nll} shows that \tt{QRTC} remains advantageous even in deep models, with a notable improvement in NLL on most datasets compared to \tt{QRC}.
Similarly, observations from \cref{fig:some_mixture_3_resnet/without_discrete} align with previous findings in PCE.
Finally, \tt{QRTC} is both significantly better than \tt{QRC} in CRPS and significantly sharper.

%Specifically, the input of dimension $d_\text{in}$ is transformed to dimension $d = 256$. Then, 8 blocks with residual connections transform their input to dimension $d_h = 512$, then back to dimension $d = 512$ using two fully-connected hidden layers.
%Finally, the last fully-connected layer transforms the input into the 9 parameters of the Gaussian mixture of size 3.

\begin{figure}[H]
	\centering
	\subcaptionbox{
		Letter-value plots showing Cohen's d for different metrics with respect to \texttt{BASE} (using an MLP model and mixture predictions of size 3 as in the main text).
	}{
		\includegraphics[width=\linewidth]{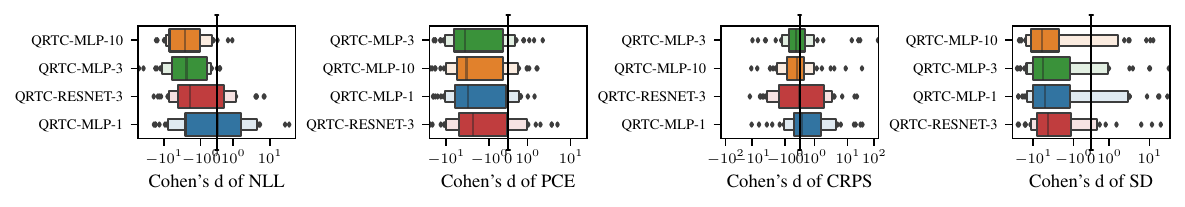}
		\vspace{-0.3cm}
	}
	\subcaptionbox{
		CD diagrams
	}{
		\includegraphics[width=\linewidth/4]{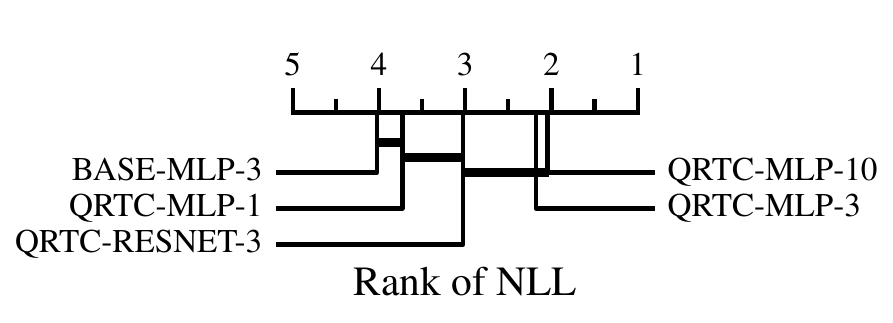}
		\includegraphics[width=\linewidth/4]{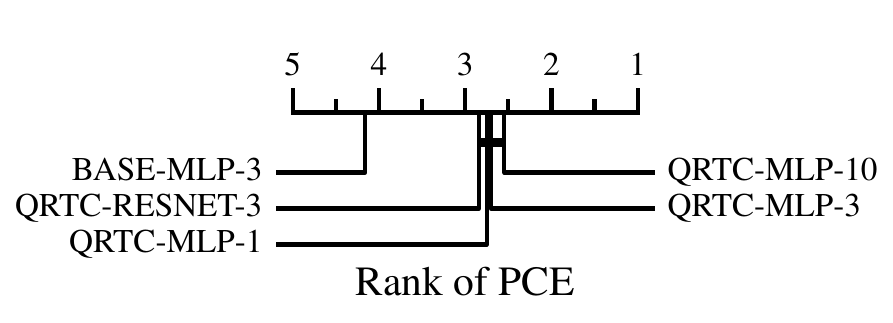}
		\includegraphics[width=\linewidth/4]{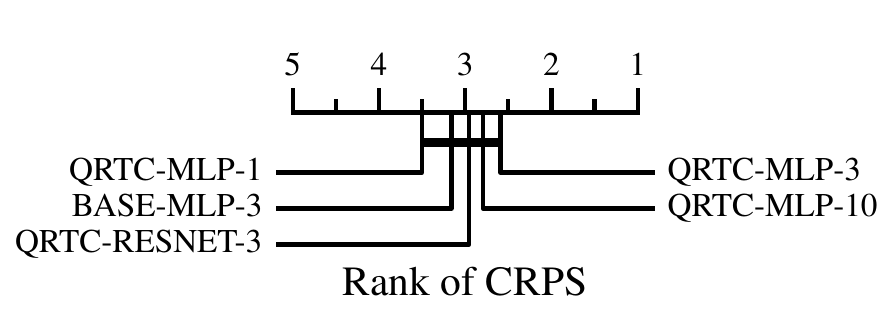}
		\includegraphics[width=\linewidth/4]{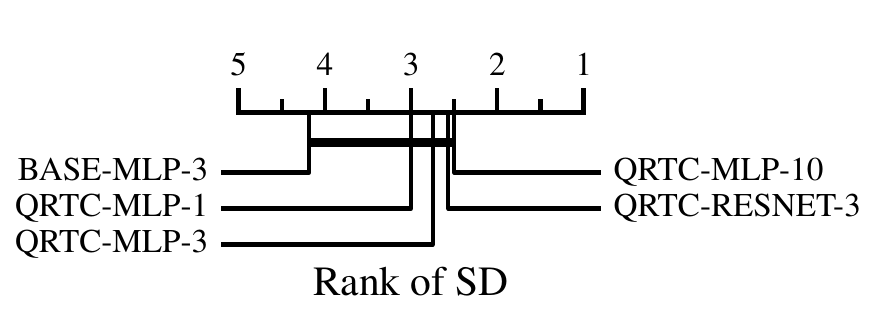}
		\vspace{-0.3cm}
	}
	\vspace{-0.3cm}
	\caption{Comparison of \tt{QRTC} with different base models.}
	\label{fig:base_models/without_discrete}
\end{figure}

Finally, we provide a comparison of the performance of \tt{QRTC} on all base models under consideration. Each model is denoted by \tt{QRTC-<BM>-$K$} where \tt{<BM>} is the base model and $K$ is the mixture size.
As illustrated in \cref{fig:base_models/without_discrete}, mixtures of size 3 and 10 achieve the best NLL and CRPS, and a simple MLP achieves a better performance than a ResNet on these datasets.

%\begin{figure}[H]
%	\centering
%	\subcaptionbox{
	%		Mixture of size 1 (single Gaussian) on a 3-layer MLP
	%		\label{fig:some_mixture_1/diff/test_nll}
	%	}{
	%		\includegraphics[width=\linewidth]{images/diff/test_nll.pdf}
	%		\vspace{-0.3cm}
	%	}
%	\subcaptionbox{
	%		Mixture of size 3 on a 3-layer MLP
	%		\label{fig:some/diff/test_nll}
	%	}{
	%		\includegraphics[width=\linewidth]{images/some/without_discrete/diff/test_nll.pdf}
	%		\vspace{-0.3cm}
	%	}
%	\subcaptionbox{
	%		Mixture of size 10 on a 3-layer MLP
	%		\label{fig:some_mixture_10/diff/test_nll}
	%	}{
	%		\includegraphics[width=\linewidth]{images/some_mixture_10/without_discrete/diff/test_nll.pdf}
	%		\vspace{-0.3cm}
	%	}
%	\subcaptionbox{
	%		Mixture of size 3 on a 18-layer ResNet
	%		\label{fig:some_mixture_3_resnet/diff/test_nll}
	%	}{
	%		\includegraphics[width=\linewidth]{images/some_mixture_3_resnet/without_discrete/diff/test_nll.pdf}
	%		\vspace{-0.3cm}
	%	}
%	\caption{Comparison of QRT and QR with respect to BASE by showing the difference between the compared methods and BASE according to the test NLL, with different size of mixtures and base models, in average over 5 runs.}
%	\vspace{-0.5cm}
%	\label{fig:some_base_model/without_discrete/diff/test_nll}
%\end{figure}

\section{IMPACT OF THE SIZE OF THE CALIBRATION MAP}
\label{sec:impact_of_cal_size}

As discussed in \cref{sec:RT_framework}, we investigate the impact of computing the calibration map from a dataset of size $M$ sampled randomly from the training dataset instead of the current batch.
%Specifically, this would correspond to changing line 2 in \cref{algo:RT_framework} from
Specifically, this would correspond to changing the training loop of \cref{algo:RT_framework} as depicted by \cref{algo:RT_cal_map}.

\SetAlgoNoLine
\SetKwInOut{Input}{Input}

\setcounter{algocf}{1}
\begin{algorithm}[H]
	\caption{QRT framework where $\phi^\text{REFL}_{\theta}$ is computed from a random sample of the training dataset.}
	\label{algo:RT_cal_map}
	\Input{Predictive CDF $F_\theta$, training dataset $\D$, size of calibration map $M$}
	$M \gets \min\Set{M, \lvert D \rvert}$ \;
	\ForEach{minibatch $\Set{(X_i, Y_i)}_{i=1}^{B} \subseteq \D$, until early stopping}{
		Sample $\Set{(X'_i, Y'_i)}_{i=1}^M$ from $\D$ without replacement \;
		Compute $Z'_i \gets F_\theta(Y'_i \mid X'_i)$, for $i = 1, \dots, M$ \;
		Define $\phi^\text{REFL}_{\theta}$ from $Z'_1, \dots, Z'_M$ using \cref{eq:refl_pdf} \;
		Compute $Z_i \gets F_\theta(Y_i \mid X_i)$, for $i = 1, \dots, B$ \;
		$\mc{L}(\theta) = -\frac{1}{B} \sum_{i=1}^B \underbrace{\log f_\theta(Y_i \mid X_i) + \log \phi^\text{REFL}_\theta(Z_i)}_{\log f'_\theta(Y_i \mid X_i)}$ \;
		Update parameters $\theta$ using $\nabla_\theta \mc{L}(\theta)$ \;
	}
\end{algorithm}

%\begin{align*}
%\text{Compute $Z_i \gets F_\theta(Y_i \mid X_i)$, for $i = 1, \dots, B$.}
%\end{align*}
%to
%\begin{align*}
%&M \gets \min\Set{M, \lvert D \rvert.} \\
%&\text{Sample $\Set{(X'_i, Y'_i)}_{i=1}^M$ from $\D$ without replacement.} \\
%&\text{Compute $Z_i \gets F_\theta(Y'_i \mid X'_i)$, for $i = 1, \dots, M$.}
%\end{align*}

While the approach proposed in the main text requires $B$ neural network evaluations per minibatch, \cref{algo:RT_cal_map} requires $M + B$ neural network evaluations per minibatch, making it relatively slower.

\begin{figure}[H]
	\centering
	\subcaptionbox{
		Boxplots of Cohen's d of different metrics on all datasets, with respect to \texttt{BASE}.
	}{
		\includegraphics[width=\linewidth]{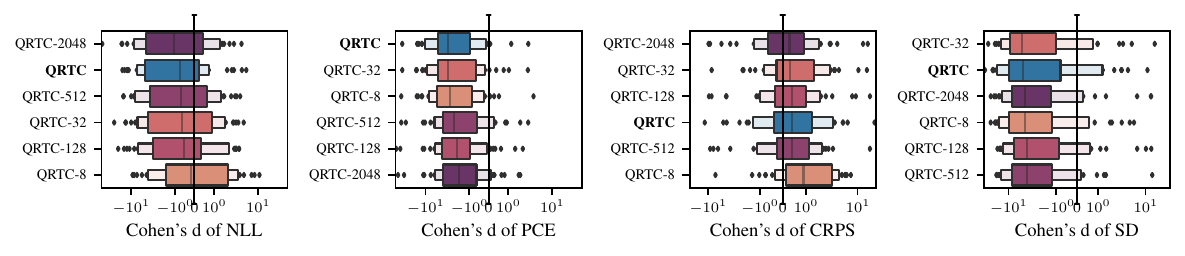}
		\vspace{-0.3cm}
	}
	\subcaptionbox{
	CD diagrams
	\label{fig:inhoc_cal_size/without_discrete/cd_diagrams}
	}{
		\includegraphics[width=\linewidth/4]{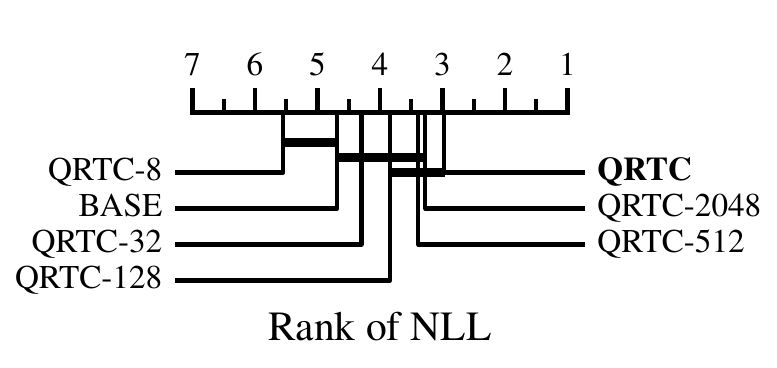}
		\includegraphics[width=\linewidth/4]{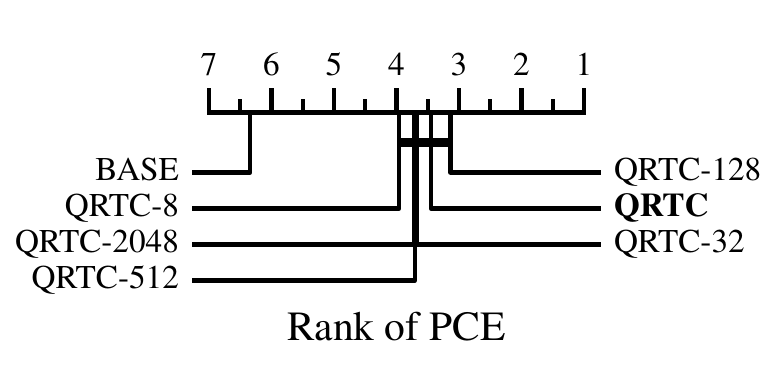}
		\includegraphics[width=\linewidth/4]{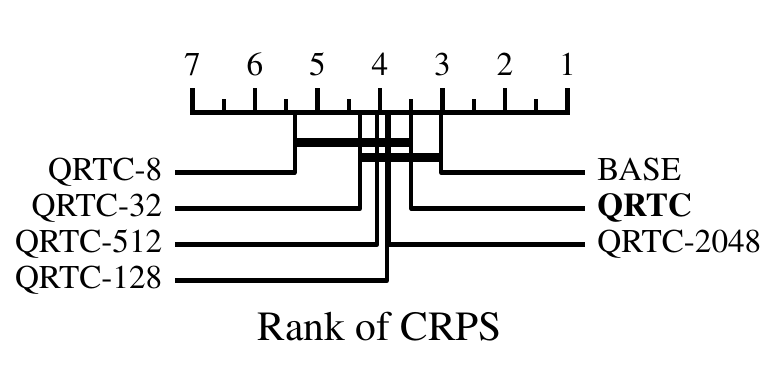}
		\includegraphics[width=\linewidth/4]{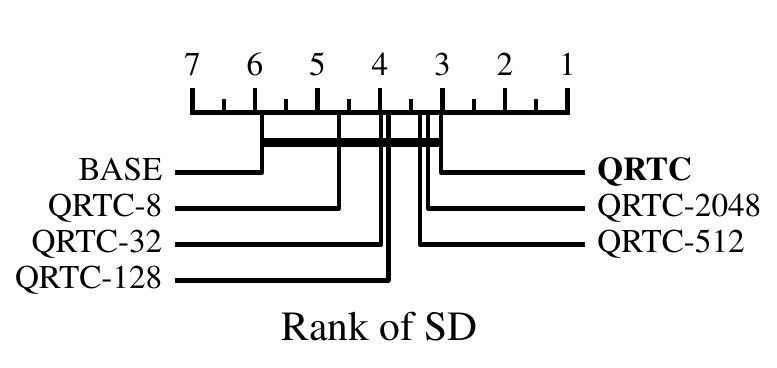}
		\vspace{-0.3cm}
	}
		\caption{Comparison of \tt{QRTC}, where the calibration map has been computed on calibration datasets of different sizes.}
	\label{fig:cal_size/without_discrete}
\end{figure}

In \cref{fig:cal_size/without_discrete}, we investigate the performance of \tt{QRTC} using \cref{algo:RT_cal_map} with calibration maps of size $M$, and denote these models \tt{QRTC-$M$}. The model \tt{QRTC} in blue corresponds to the same model as in the main paper, with a calibration map computed from the current batch of size $B = 512$.
It is worth noting that the post-hoc step is still performed on a calibration dataset of the same size for all models.
In terms of NLL, models with a larger calibration map tend to perform better.
In terms of PCE, all post-hoc models perform similarly.
While no decisive conclusions can be drawn, \cref{fig:inhoc_cal_size/without_discrete/cd_diagrams} suggests that larger calibration maps tend to result in improved CRPS and sharper predictions.
Overall, estimating the NLL of QRT using a larger calibration map tends to give more accurate predictions.

%The proposed method introduces increased computational demands since it requires evaluating the neural network on an additional set of $N'$ supplementary points at each iteration. However, limiting the size of the calibration dataset to $N' = 2048$ points is in line with the guidelines presented in \citet{Angelopoulos2021-rc}, which advocate for selecting a minimum of $N' = 1000$ points to establish a robust conformal guarantee.

\section{IMPACT OF THE BANDWIDTH HYPERPARAMETER $b$}
\label{sec:impact_of_inhoc_b}

\begin{figure}[H]
	\centering
	\subcaptionbox{
		Boxplots of Cohen's d of different metrics on all datasets, with respect to \texttt{BASE}.
	}{
		\includegraphics[width=\linewidth]{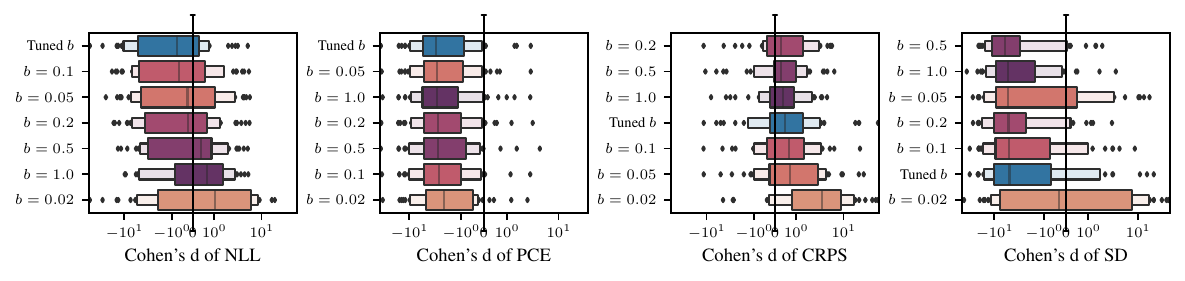}
		\vspace{-0.3cm}
	}
	\subcaptionbox{
		CD diagrams
	}{
		\includegraphics[width=\linewidth/4]{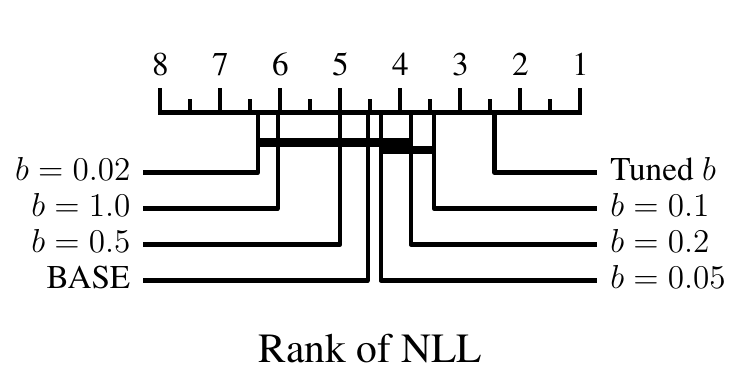}
		\includegraphics[width=\linewidth/4]{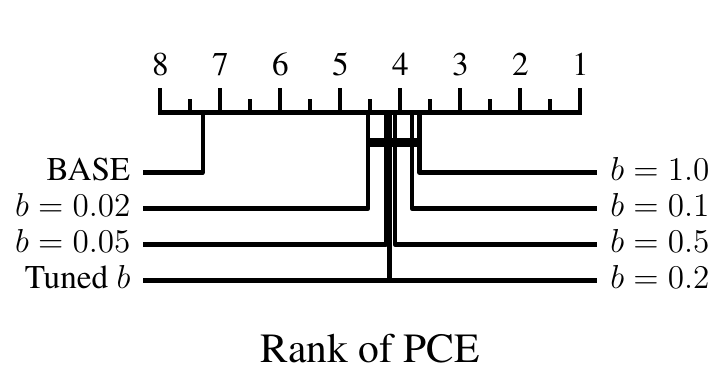}
		\includegraphics[width=\linewidth/4]{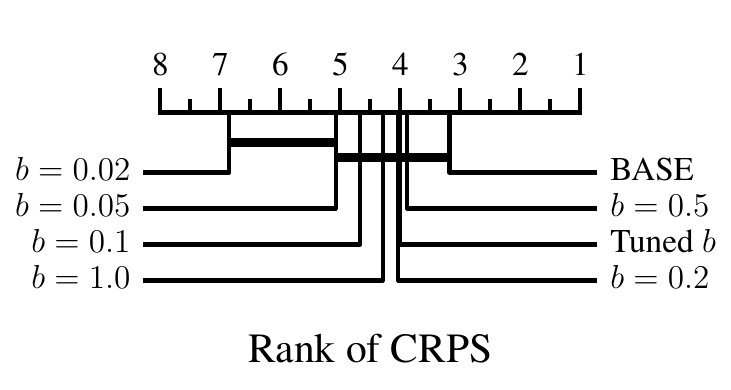}
		\includegraphics[width=\linewidth/4]{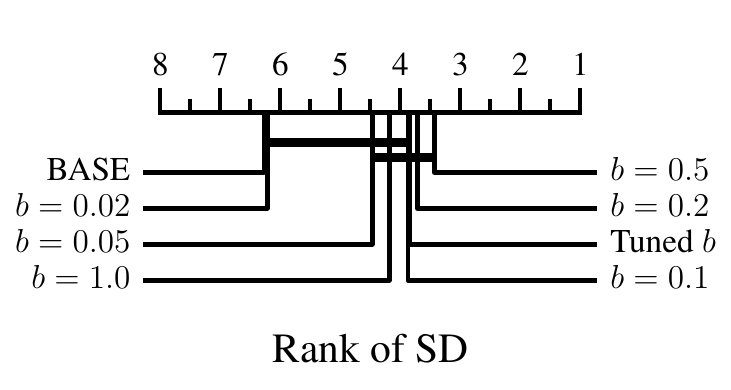}
		\vspace{-0.3cm}
	}
	\caption{Comparison of \tt{QRTC} with different values of the hyperparameter $b$.}
	\label{fig:inhoc_b/without_discrete}
\end{figure}

We evaluate the effect of tuning the bandwidth hyperparameter $b$ in QRT.
In \cref{fig:inhoc_b/without_discrete}, the bandwidth is either selected by minimizing the validation NLL from the set {0.02, 0.05, 0.1, 0.2, 0.5, 1} (denoted by Tuned $b$), or it is set to a fixed value.
The results show that tuning $b$ results in a significant improvement in NLL compared to fixed values of $b$.
Values of 0.1, 0.2 and 0.05 yield the best NLL improvement while values of 0.2 and 0.5 yield the best CRPS improvement compared to \tt{BASE}.
%This is however not the case for extreme values such as 0.02 and 1.

\section{KERNEL DENSITY ESTIMATION ON A FINITE DOMAIN}
\label{sec:kde_finite}

We provide more motivation and details regarding the calibration map $\Phi_\theta^\text{REFL}$ discussed in \cref{sec:decomposition} in the main paper.

The limitation of a standard kernel density estimation within a finite domain $[a, b]$ using a kernel like the logistic distribution is that the resulting distribution becomes ill-defined due to non-null density values extending below $a$ and beyond $b$.
In the following, to simplify notation, we denote $\Phi_\theta^\text{KDE}$ and $\phi_\theta^\text{KDE}$ by $F$ and $f$ respectively.

We would like to highlight that following our independent development of the "Reflected Kernel", we later discovered that this concept had originally been introduced by \cite{Blasiok2023-oh}.

\subsection{Truncated distribution}

A standard approach is to truncate the distribution and redistribute the density below $a$ and above $b$, namely $F(b) - F(a)$, such that the distribution is normalized.
The resulting CDF is:
\begin{equation}
	\Phi_\theta^\text{TRUNC}(x) = \begin{cases}
		\nicefrac{F(x) - F(a)}{F(b) - F(a)} &\quad \text{if $x \in [a, b]$} \\
		0 &\quad \text{if $x < a$} \\
		1 &\quad \text{if $x > b$}
	\end{cases}
\end{equation}
and the resulting PDF is:
\begin{equation}
	\phi_\theta^\text{TRUNC}(x) = \begin{cases}
		\nicefrac{f(x)}{F(b) - F(a)} &\quad \text{if $x \in [a, b]$} \\
		0 &\quad \text{if $x \not\in [a, b]$}.
	\end{cases}
\end{equation}

A drawback of truncating the distribution on a finite domain is that the resulting distribution will be biased to have lower density close to $a$ and $b$ and higher density elsewhere, as illustrated on \cref{fig:standard_vs_truncated_vs_reflected}.

\subsection{Proposed approach: Reflected distribution}

To remedy this problem, we define a new PDF $\phi_\theta^\text{REFL}$ that "reflects" the base density $f$ around $a$ and $b$.
More precisely, for a given $z > 0$, the density in $a - z$ is redistributed to $a + z$ and the density in $b + z$ is redistributed to $b - z$.
We assume that the density $f$ is not too spread out, specifically $f(x) = 0$ for $x \not\in [a - (b - a), b + (b - a)]$. 
The resulting CDF is defined by:
\begin{equation}
	\label{eq:refl_cdf}
	\Phi_\theta^\text{REFL}(x) = \begin{cases}
		F(x) - F(2a - x) + 1 - F(2b - x) &\quad \text{if $x \in [a, b]$} \\
		0 &\quad \text{if $x < a$} \\
		1 &\quad \text{if $x > b$} \\
	\end{cases} 
\end{equation}
and the corresponding PDF is defined by:	
\begin{equation}
	\label{eq:refl_pdf}
	\phi_\theta^\text{REFL}(x) = \begin{cases}
		f(x) + f(2a - x) + f(2b - x) &\quad \text{if $x \in [a, b]$} \\
		0 &\quad \text{if $x \not\in [a, b]$}.
	\end{cases} 
\end{equation}

\begin{figure}[H]
	\centering
	\includegraphics[width=\linewidth]{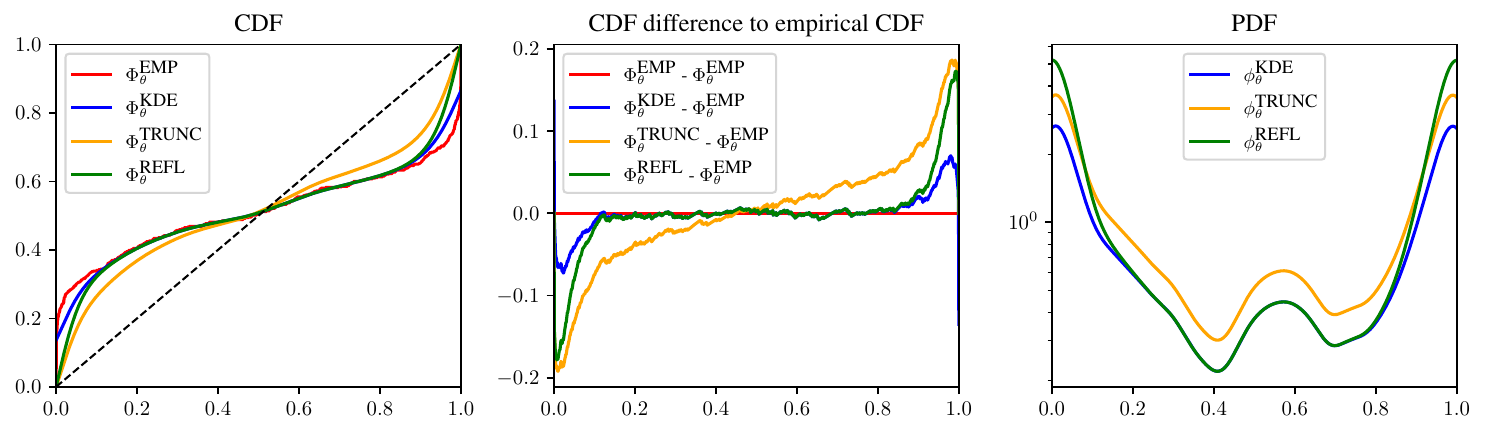}
	\caption{Comparison of different methods to estimate the calibration map. In this example, 512 PITs have been sampled from a beta distribution $Z \sim \text{Beta}(0.2, 0.2)$ and the calibration map is estimated using $\Phi_\theta^\text{KDE}$ with $b = 0.1$ (Equation \cref{eq:Phi_theta^KDE} in the main text).}
	\label{fig:standard_vs_truncated_vs_reflected}
\end{figure}

\cref{fig:standard_vs_truncated_vs_reflected} compares four methods to estimate the calibration map from PIT realizations $Z_1, \dots, Z_N$.
The method $\Phi_\theta^\text{EMP}$ was introduced in \cref{sec:background} and corresponds to the empirical CDF, which is not smooth.
In contrast, the methods $\Phi_\theta^\text{KDE}$, $\Phi_\theta^\text{TRUNC}$ and $\Phi_\theta^\text{REFL}$ offer smooth estimations.
This figure shows that $\Phi_\theta^\text{REFL}$ is closer to the empirical CDF than $\Phi_\theta^\text{TRUNC}$ and the value of the corresponding PDF $\phi_\theta^\text{REFL}$ is not overestimated, suggesting the superiority of this estimator.

\begin{figure}[H]
	\centering
	\subcaptionbox{
		Boxplots of Cohen's d of different metrics on all datasets, with respect to \texttt{BASE}.
	}{
		\includegraphics[width=2\linewidth/3]{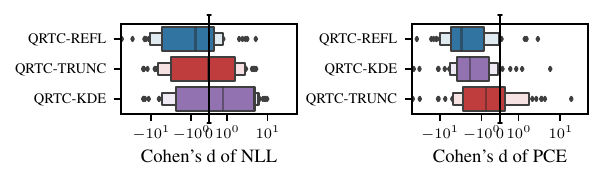}
		\vspace{-0.3cm}
	}
	\subcaptionbox{
		CD diagrams
	}{
		\includegraphics[width=\linewidth/3]{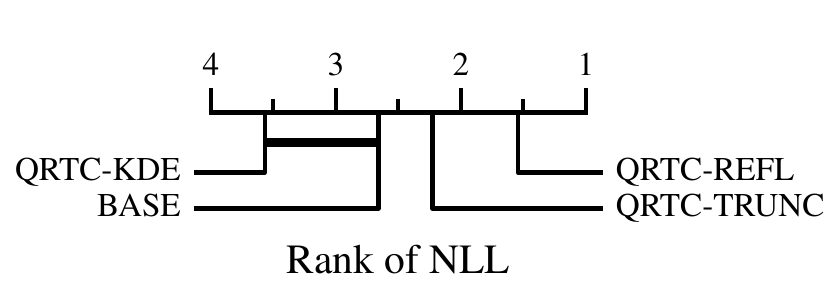}
		\includegraphics[width=\linewidth/3]{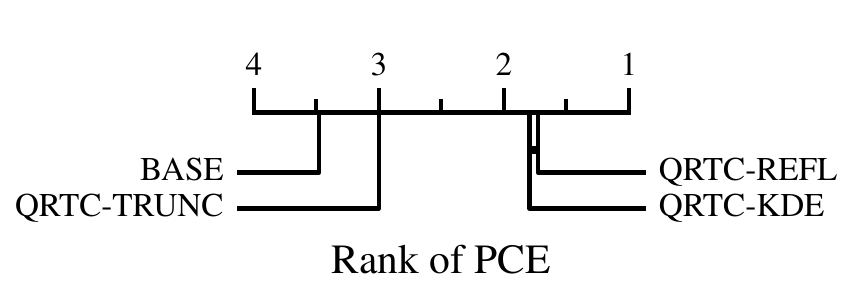}
		\vspace{-0.3cm}
	}
	\caption{Comparison between different kernel density estimation approaches. Note that the metrics CRPS and SD are not provided because they are ill-defined for \tt{QRTC-KDE}. More precisely, since the quantile function $\left(\Phi_\theta^\text{KDE}\right)^{-1}$ returns values outside the interval $[0, 1]$, we can not correctly sample from the model.}
	\label{fig:reflected/without_discrete}
\end{figure}

\cref{fig:reflected/without_discrete} compares \tt{QRTC} where the calibration map of the post-hoc model has been estimated using either $\Phi_\theta^\text{KDE}$, $\Phi_\theta^\text{TRUNC}$ and $\Phi_\theta^\text{REFL}$.
We denote these methods \tt{QRTC-KDE}, \tt{QRTC-TRUNC} and \tt{QRTC-REFL} respectively. It is worth noting that \tt{QRTC-REFL} corresponds to the method \tt{QRTC} in the main text.
In terms of NLL, \tt{QRTC-REFL} performs significantly better in terms of NLL and \tt{QRTC-KDE} is the least effective.
In terms of PCE, \tt{QRTC-REFL} and \tt{QRTC-KDE} perform similarly and \tt{QRTC-TRUNC} is the least effective.
This confirms that the method of Reflected Kernel should be preferred. % in combination with QRT.

\newpage
\section{DETAILED METRICS ON INDIVIDUAL DATASETS}
\label{sec:metrics_per_dataset}

\begin{figure}[H]
	\centering
	\subcaptionbox{
		Test NLL
		\label{fig:some/diff/test_nll_2}
	}{
		\includegraphics[width=\linewidth]{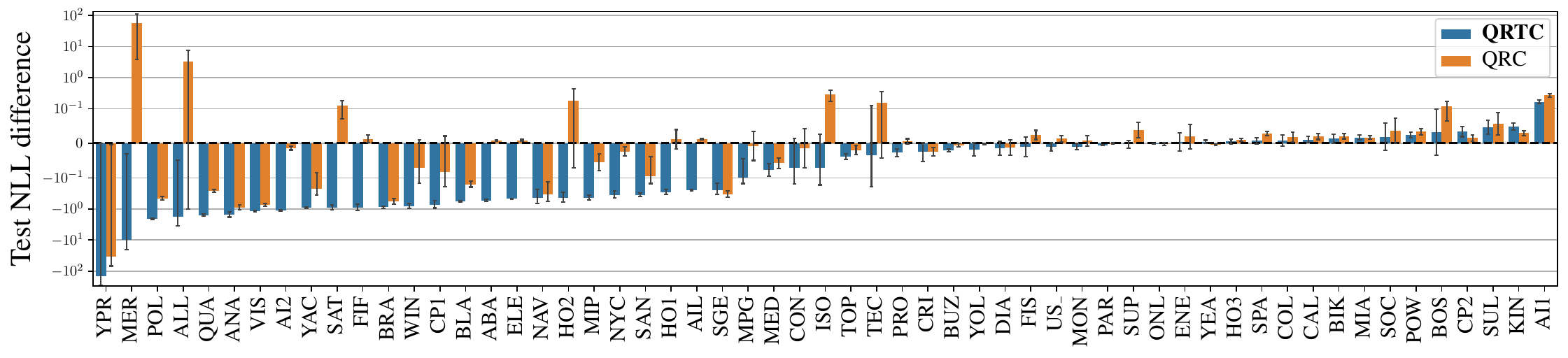}
		\vspace{-0.3cm}
	}
	\subcaptionbox{
		Test PCE
		\label{fig:some/diff/test_calib_l1}
	}{
		\includegraphics[width=\linewidth]{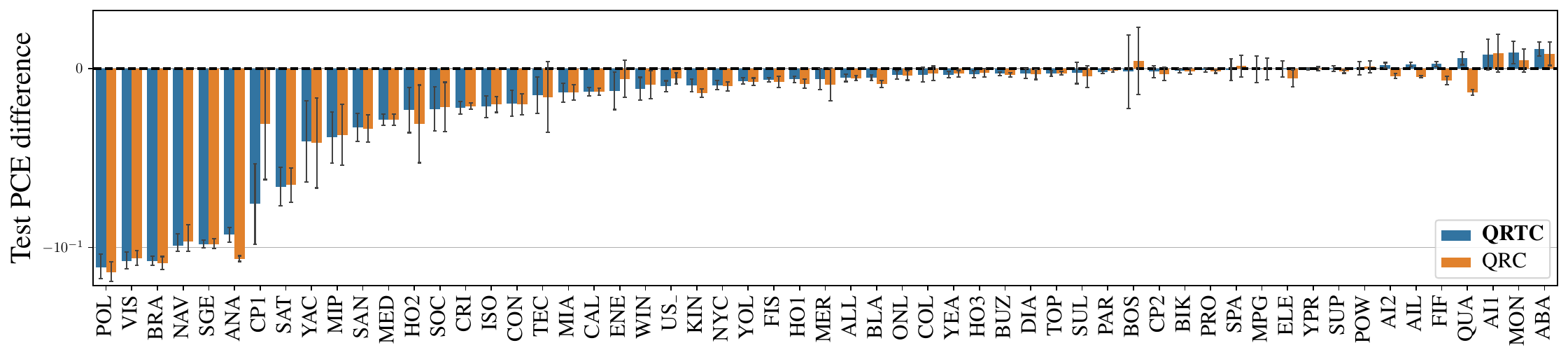}
		\vspace{-0.3cm}
	}
	\subcaptionbox{
		Test CRPS
		\label{fig:some/diff/test_wis}
	}{
		\includegraphics[width=\linewidth]{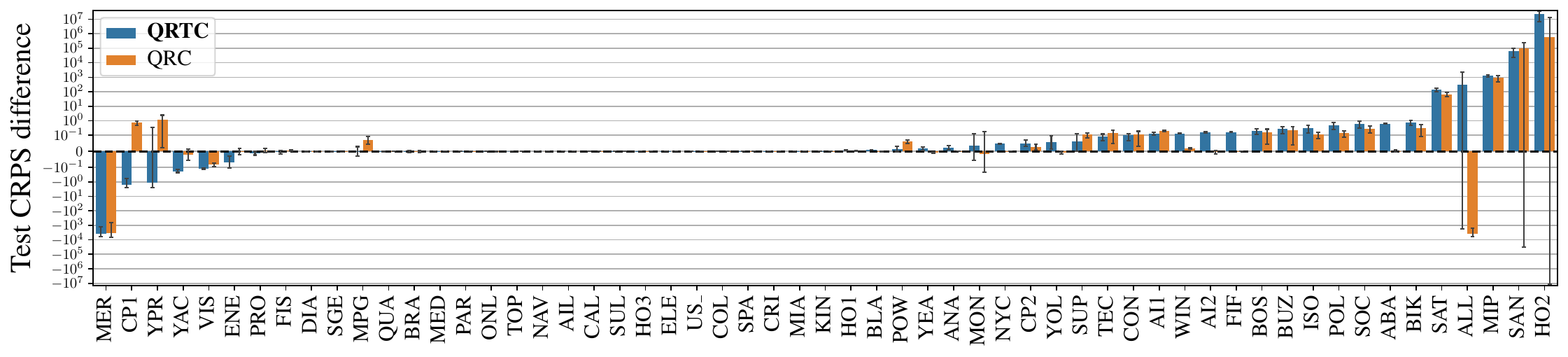}
		\vspace{-0.3cm}
	}
	\subcaptionbox{
		Test SD
		\label{fig:some/diff/test_stddev}
	}{
		\includegraphics[width=\linewidth]{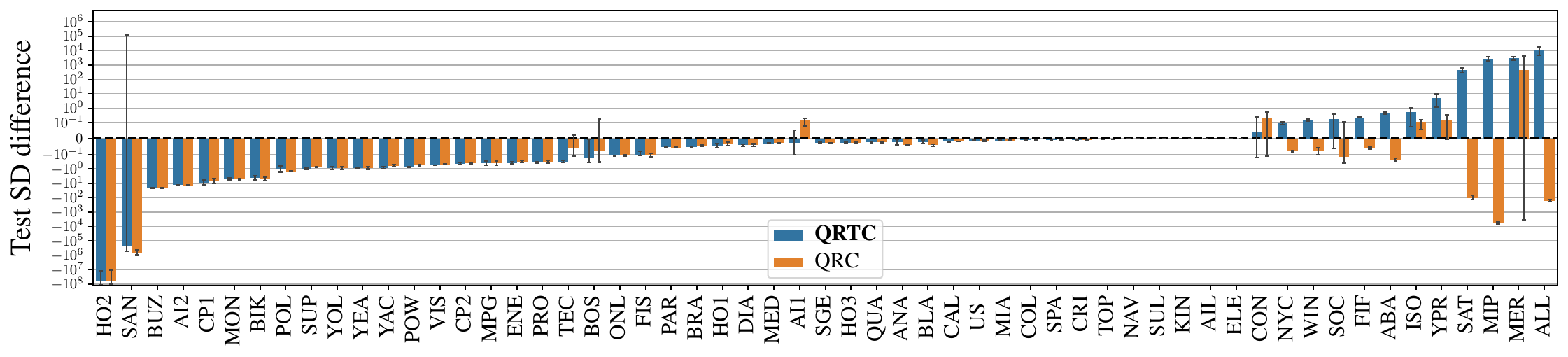}
		\vspace{-0.3cm}
	}
	\caption{Comparison of \tt{QRTC} and \tt{QRC} with respect to \tt{BASE} by showing the difference between the compared methods and \tt{BASE} according to a given metric, in average over 5 runs.}
	\vspace{-0.5cm}
	\label{fig:some/diff}
\end{figure}

For a more comprehensive view, \cref{fig:some/diff} presents a diagram analogous to \cref{fig:some/without_discrete/diff/test_nll} from the main text, but extends the comparison across NLL, PCE, CRPS, and SD metrics. Additionally, these diagrams incorporate datasets previously omitted in \cref{sec:benchmark_datasets}. With respect to NLL, \tt{QRTC} consistently surpasses \tt{QRC} in the majority of datasets. In terms of PCE, both post-hoc methods perform similarly. In terms of CRPS, both post-hoc methods display comparable performances but are sometimes outperformed by \tt{BASE}. Analyzing SD, \tt{QRC} exhibits greater sharpness than \tt{BASE} in nearly all instances, while \tt{QRTC} sometimes does not exhibit increased sharpness.

\section{RESULTS WITH DIFFERENT VALUES OF $\alpha$}
\label{sec:inhoc_alpha}

We provide detailed results regarding the hyperparameter $\alpha$ of \cref{algo:RT_framework} in the main text.
As discussed in \cref{sec:RT_framework}, it is possible to design an algorithm unifying QRTC, QRC and QREGC, where the methods only differ by the hyperparameter $\alpha$.
We assume here that Quantile Recalibration is applied ($C = \text{True}$) and only consider variations of the hyperparameter $\alpha$.
As discussed previously, a value of $\alpha = 1$ corresponds to \tt{QRTC}, $\alpha = 0$ corresponds to \tt{QRC} and tuning $\alpha$ in order to minimize $\text{PCE}(F_\theta)$ corresponds to \tt{QREGC} with regularization strength $\lambda = -\alpha$.

\begin{figure}[H]
	\centering
	\subcaptionbox{
		Boxplots of Cohen's d of different metrics on all datasets, with respect to \texttt{QRC}.
	}{
		\includegraphics[width=\linewidth]{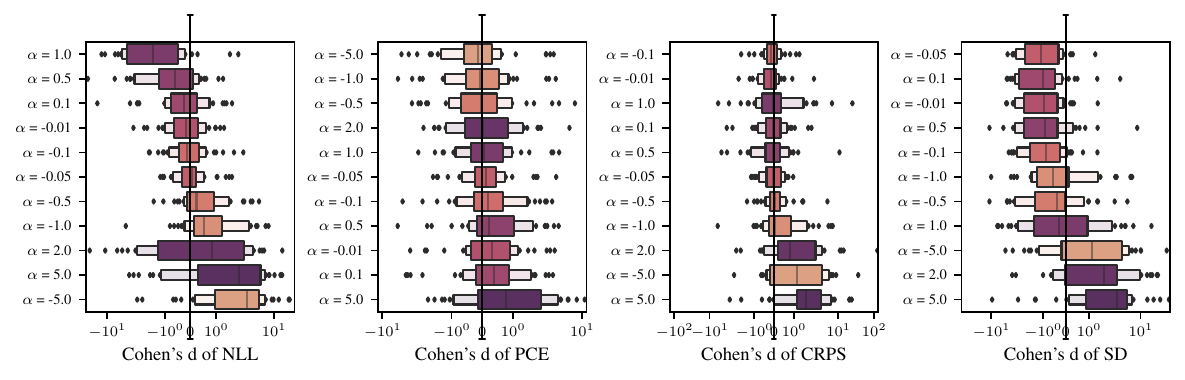}
		\vspace{-0.3cm}
		}
	\subcaptionbox{
		CD diagrams
	}{
		\includegraphics[width=\linewidth/4]{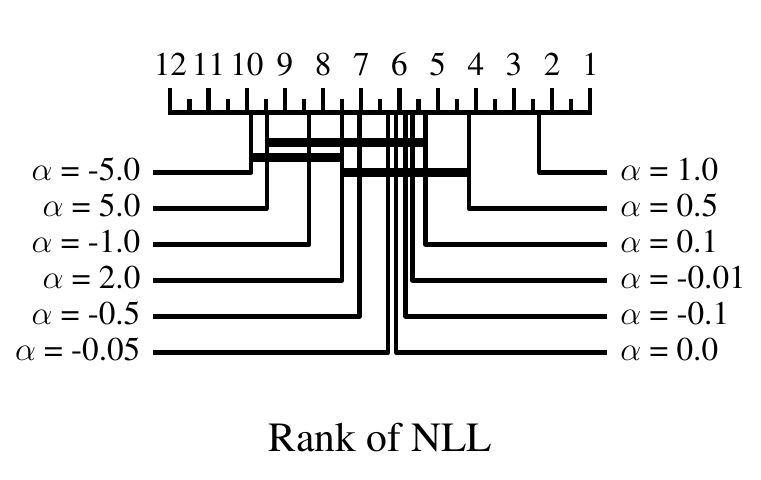}
		\includegraphics[width=\linewidth/4]{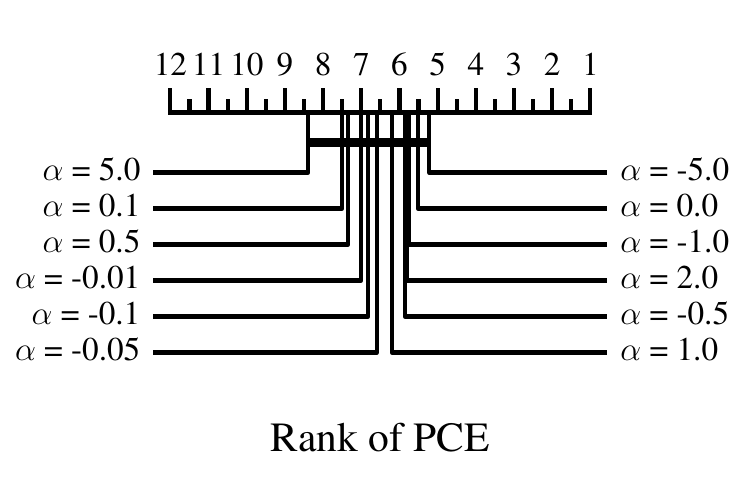}
		\includegraphics[width=\linewidth/4]{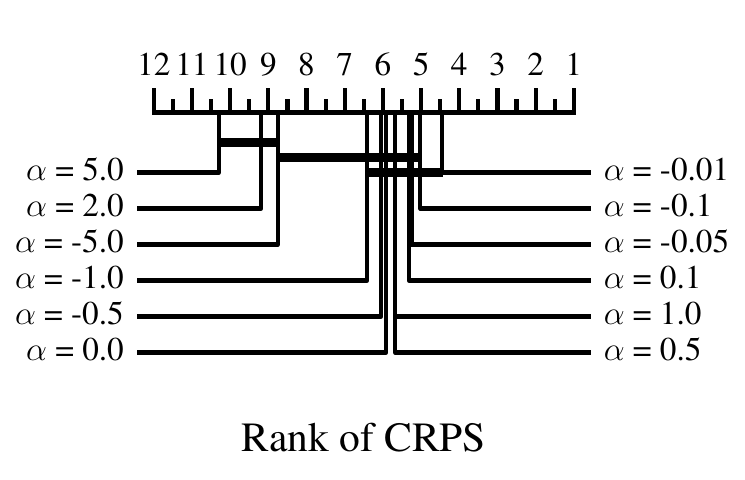}
		\includegraphics[width=\linewidth/4]{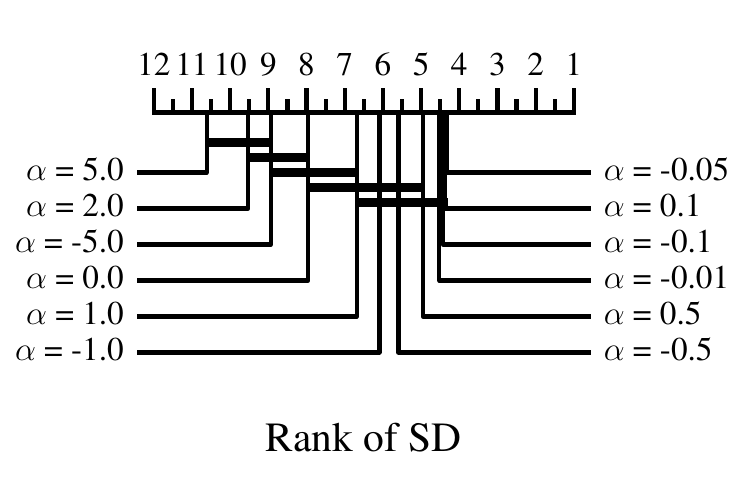}
		\vspace{-0.3cm}
	}
	\caption{Comparison of different values of the hyperparameter $\alpha$.}
	\label{fig:inhoc_alpha/without_discrete}
\end{figure}

In \cref{fig:inhoc_alpha/without_discrete}, we provide results with different values of $\alpha$.
Values of $\alpha$ between 0 and 1 can be considered as an intermediate version between \tt{QRC} and \tt{QRTC}, while negative values correspond to \tt{QREGC}.
We also explore values greater than 1 to visualize trends.

As expected, $\alpha = 1$, corresponding to the NLL decomposition of the recalibrated model, obtains the best NLL, and is significantly better than other values of $\alpha$.
In terms of CRPS, there is no significant differences for values of $\alpha$ between 0 and 1.

\section{METRICS PER EPOCH}
\label{sec:metrics_per_epoch}

In this section, we compare the behavior of training and validation NLL and PCE for both \tt{QRT} and \tt{BASE} throughout the training process. We present learning curves for all the datasets considered in this study, sorted by the number of training points. For detailed information and the complete names of these datasets, please refer to \cref{table:datasets}.

%The setup and conclusions are the same than in the illustrative example (\cref{sec:motivating_example}).
%The training curves are averaged over 5 runs and the shaded area corresponds to one standard error.
%The vertical bars represent the epoch that was selected by early stopping (the one that minimizes the validation NLL), averaged over the 5 runs.
%The horizontal bars represent the value of the metric at the selected epoch, averaged over the 5 runs.

The setup mirrors the illustrative example presented in \cref{sec:motivating_example}. The training curves are averaged over 5 runs, while the shaded area corresponds to one standard error. The vertical bars represent the epoch selected through early stopping, which minimizes the validation NLL, averaged over the 5 runs. The horizontal bars represent the average metric value at the selected epoch across the 5 runs. We draw the same conclusions as in the illustrative example (\cref{sec:motivating_example}).
The CRPS and SD are not provided due to the high computational time required to computed these metrics after Quantile Recalibration.

\begin{figure}[H]
	\centering
	\includegraphics[width=\linewidth]{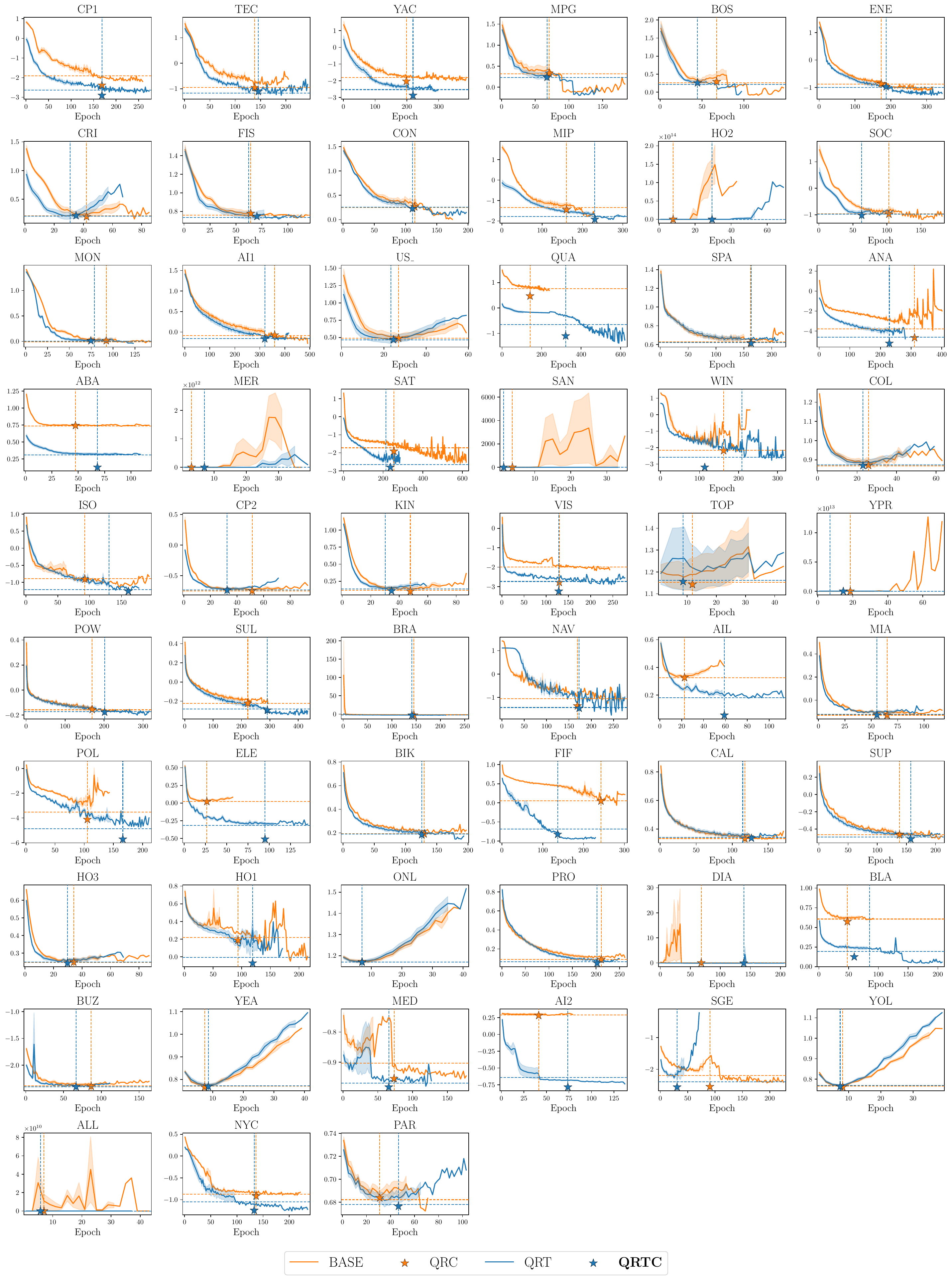}
	\caption{NLL on the validation dataset per epoch.}
	\label{fig:metric_per_epoch/val_nll}
\end{figure}

\begin{figure}[H]
	\centering
	\includegraphics[width=\linewidth]{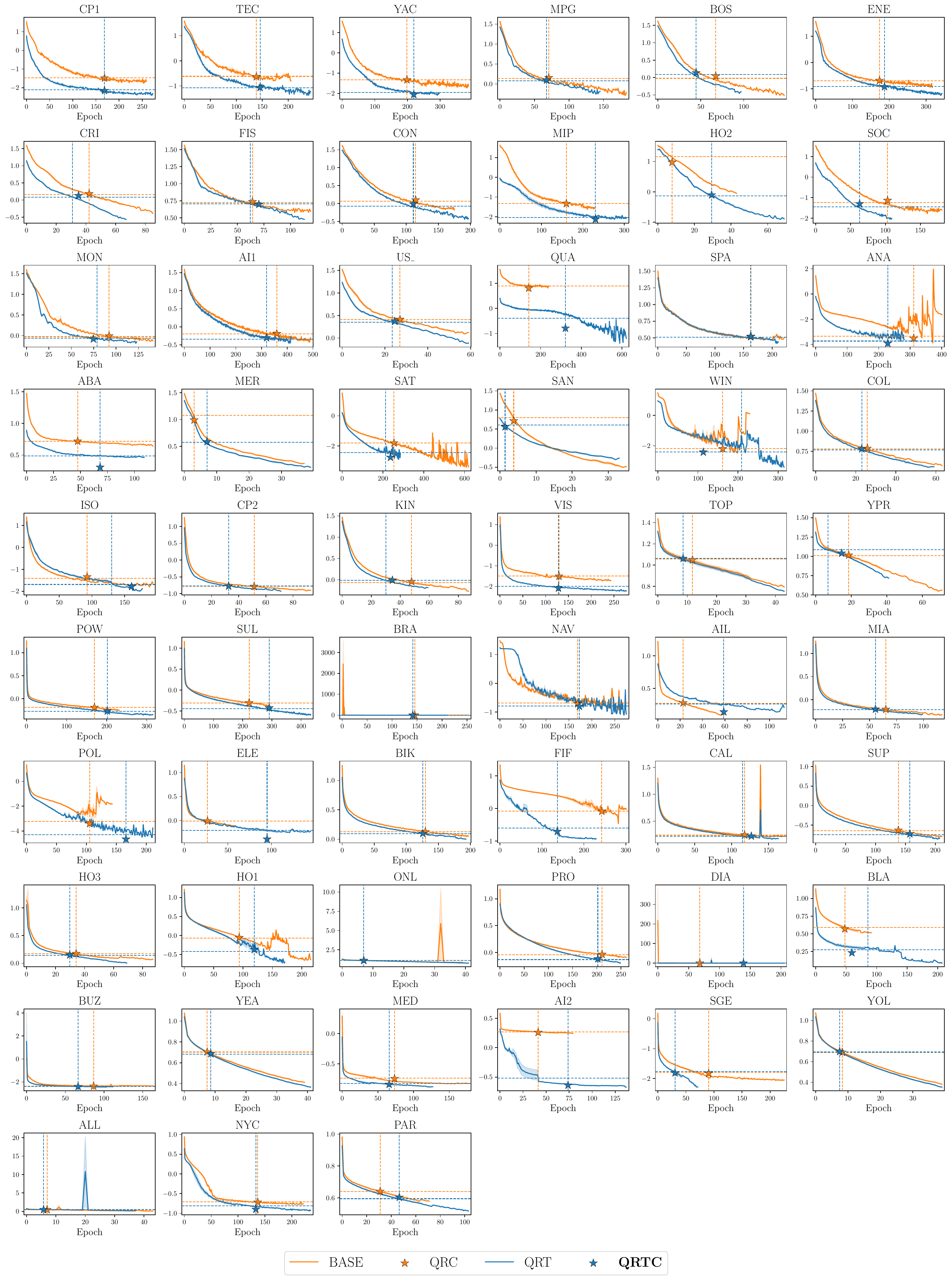}
	\caption{NLL on the training dataset per epoch.}
	\label{fig:metric_per_epoch/train_nll}
\end{figure}

In \cref{fig:metric_per_epoch/val_nll,fig:metric_per_epoch/train_nll}, we observe that the NLL of \tt{QRT} tends to be lower after the same number of epochs, indicating improved probabilistic predictions on both the training and validation datasets. Importantly, this improvement in NLL is consistent across datasets of different size. While the most significant enhancement in NLL is seen in datasets with a high level of discreteness such as \texttt{WIN}, \texttt{ANA} and \texttt{QUA}, noticeable improvements are also observed in most non-discrete datasets like \texttt{CP1}, \texttt{YAC} and \texttt{PAR}.

\begin{figure}[H]
	\centering
	\includegraphics[width=\linewidth]{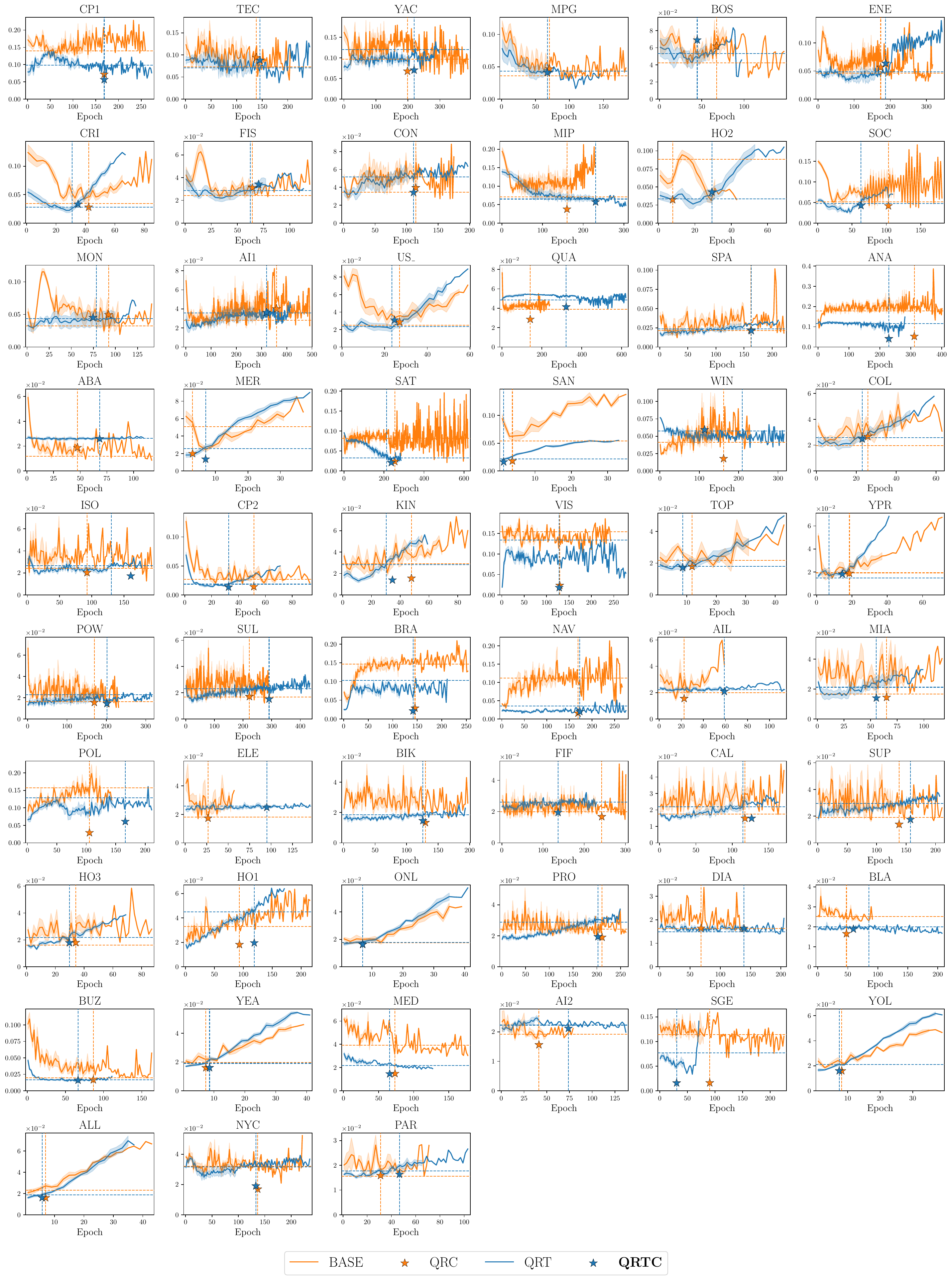}
	\caption{PCE on the validation dataset per epoch.}
	\label{fig:metric_per_epoch/val_calib_l1}
\end{figure}

\begin{figure}[H]
	\centering
	\includegraphics[width=\linewidth]{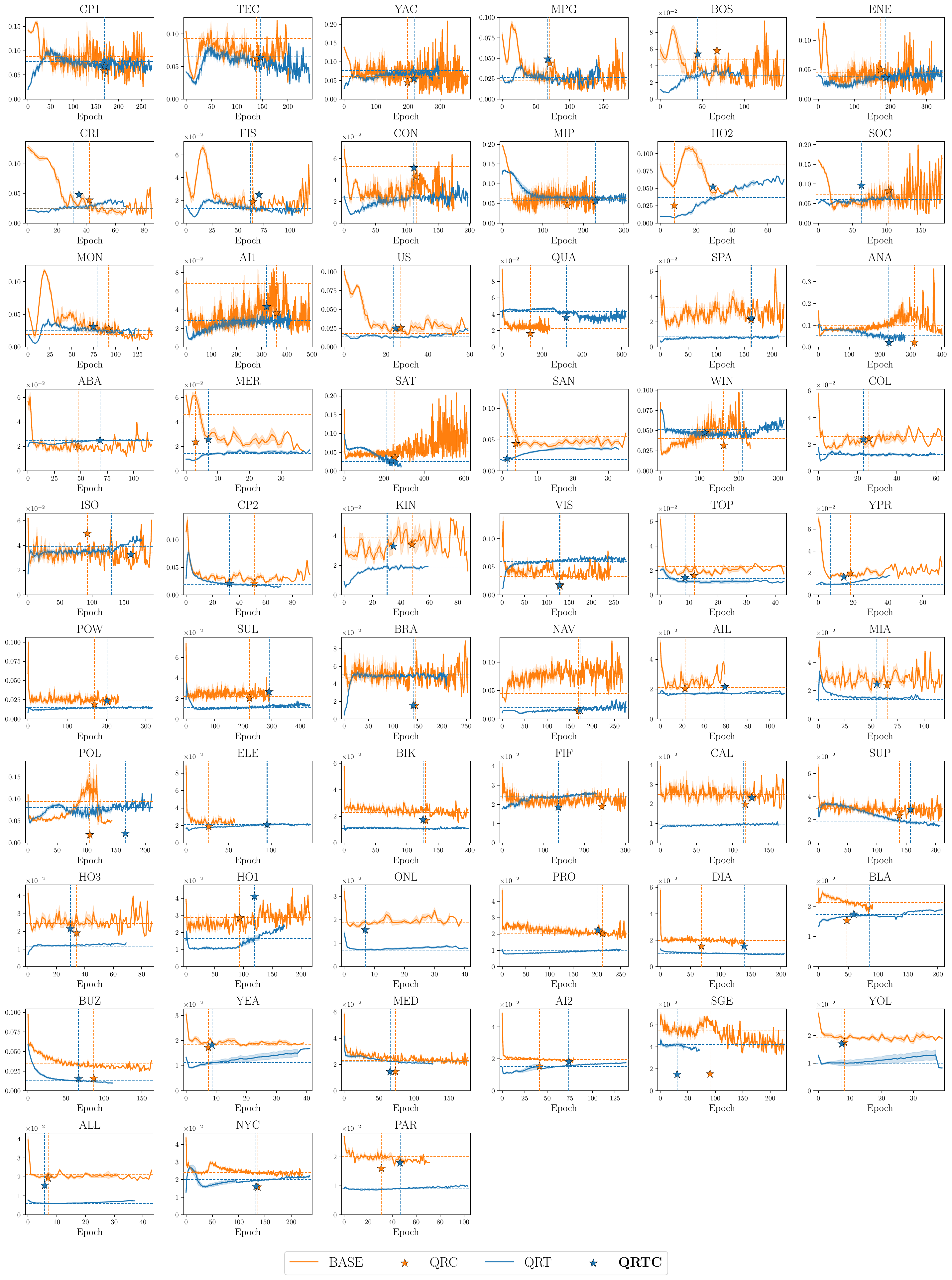}
	\caption{PCE on the training dataset per epoch.}
	\label{fig:metric_per_epoch/train_calib_l1}
\end{figure}

Referring to Figure \ref{fig:metric_per_epoch/val_calib_l1} and Figure \ref{fig:metric_per_epoch/train_calib_l1}, we can observe that the PCE of \texttt{BASE} often exhibits higher variability across epochs compared to \texttt{QRT} on both the validation and training datasets. This phenomenon indicates the regularization effect of \tt{QRT}. Additionally, we notice that the PCE is frequently lower at the same epochs for \tt{QRT}, although there are instances where this is not the case.

After reaching a certain epoch (indicated by the vertical bar), the model starts to overfit, leading to an expected increase or stabilization of NLL and PCE on the validation dataset. On the other hand, the NLL on the training dataset continues to decrease as anticipated, while the PCE exhibits high variation depending on the dataset.

%While, in this paper, we generally report the CRPS and standard deviation on the test set, we however do not report the CRPS and standard deviation per epoch.
%In our experiments, we computed the CRPS and standard deviation based on quantiles of the predictive distribution, which required to invert the CDF using binary search.
%An accurate evaluation would be prohibitively computationally demanding.

\clearpage
\section{RELATIONSHIP BETWEEN THE DISCRETENESS OF A DATASET AND THE PERFORMANCE OF DIFFERENT MODELS}
\label{sec:discreteness}

In this section, we discuss the issue that certain datasets from the UCI and OpenML benchmarks may not be suitable for regression, as introduced in \cref{sec:benchmark_datasets}
Although we consider regression benchmarks, we observe that, in many datasets, the target $Y$ presents some level of discreteness.
This is not surprising due to the finite precision of numbers and to the roundings that can appear during data collection.
For example, \cref{table:datasets} shows that, on 44 out of 57 datasets, more than half of the targets $Y$ appear at least twice.
This potential issue is more important for certain datasets where some values of the targets $Y$ appear very frequently.

We propose to identify these datasets using the proportions of values $Y$ in the dataset that are among the 10 most frequent values, and we call this proportion the level of discreteness.
For example, if a dataset only contains 10 distinct values, the level of discreteness would be 100\%.
\cref{table:datasets} in the Supplementary Material shows that 13 out of 57 datasets have a level of discreteness above 0.5, i.e., more than half of the targets are among the 10 most frequent ones. These datasets appear in all 4 benchmark suites.

\begin{figure}[H]
	\centering
	\includegraphics[width=\linewidth]{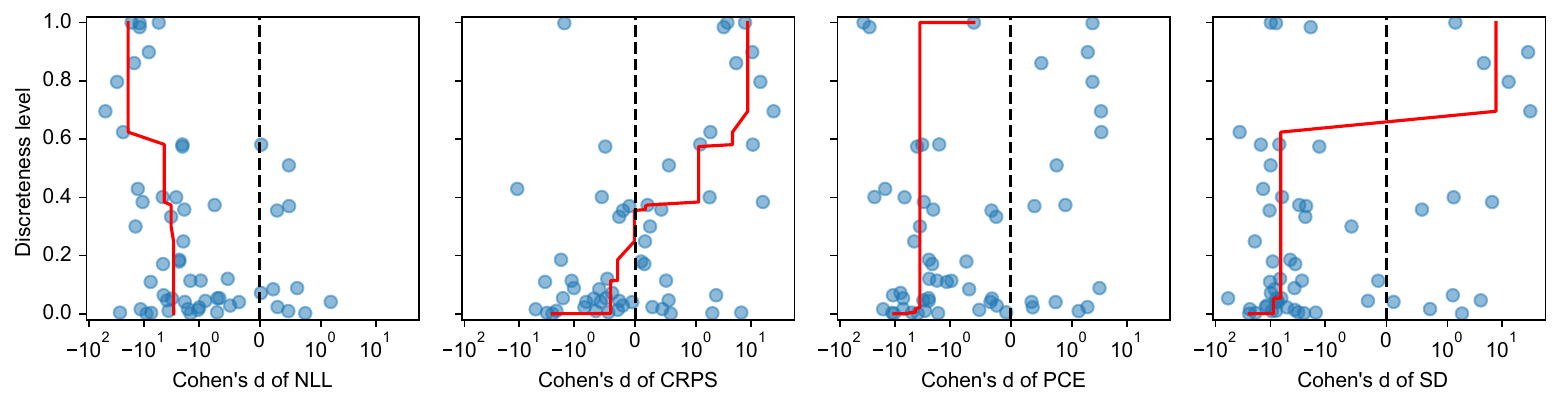}
	\caption{Cohen's d of different metrics compared to the discreteness level of a dataset for the \texttt{QRTC} model relative to the \tt{BASE} model.}
	\vspace{-0.35cm}
	\label{fig:cohens_d_vs_top10/QRTC}
\end{figure}

\begin{figure}[H]
	\centering
	\includegraphics[width=\linewidth]{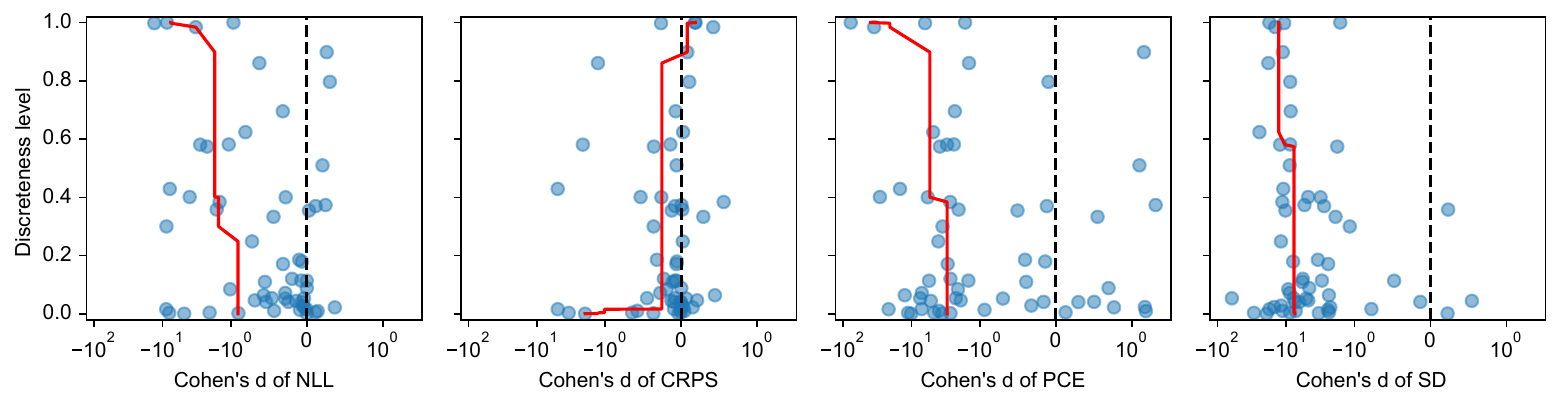}
	\caption{Cohen's d of different metrics compared to the discreteness level of a dataset for the \texttt{QRC} model relative to the \tt{BASE} model.}
	\label{fig:cohens_d_vs_top10/QRC}
\end{figure}

In \cref{fig:cohens_d_vs_top10/QRTC,fig:cohens_d_vs_top10/QRC}, we plot for each dataset the Cohen's d of different metrics, averaged over 5 runs, compared to the discreteness level of the dataset.
For the NLL, CRPS and PCE, negative values of the Cohen's d correspond to an improvement.
In order to show the average Cohen's d conditional to the discreteness level, we provide an isotonic regression estimate in red.

\cref{fig:cohens_d_vs_top10/QRTC} shows that \tt{QRTC} tends to provide a decreased NLL and increased CRPS for higher discreteness levels.
This can be explained by the ability of \tt{QRTC} to put a high likelihood on a few values by minimizing the NLL but neglect other aspects of the distributions. While previous work \citep{Kohonen2006-vs} has highlighted the unsuitability of NLL as a metric for discrete datasets, they are still commonly found in regression benchmarks. For example, \citet{Lakshminarayanan2017-zg} and \citet{Amini2019-vz} trained a model based on NLL on the \texttt{wine\_quality} dataset for which the output variable only takes 7 distinct values.

\cref{fig:cohens_d_vs_top10/QRC} shows the same metrics for \tt{QRC}, where we observe that the improvement in NLL is less marked on datasets with a higher discreteness level.
The CRPS, however, is not decreased as much as with \tt{QRTC}, which suggests that \tt{QRTC} is not suitable for datasets with a high level of discreteness.
We don't observe a notable trend in terms of PCE.

%\let\clearpage\relax
%\clearpage

\section{RESULTS ON ALL DATASETS}
\label{sec:results/with_discrete}

As discussed in \cref{sec:benchmark_datasets}, we provide the full results including the datasets with a high discreteness level.
Despite the potential issues discussed in \cref{sec:discreteness}, the conclusions drawn in Section \ref{sec:experiments} remain unchanged.
\texttt{QRTC} demonstrates a significant improvement in NLL with a negligible loss in CRPS.
Figure detailing metrics on the individual datasets are also available in \cref{sec:metrics_per_dataset}.

%Since the experiments discussed in the main text (see Section \ref{sec:experiments}) could potentially be influenced by the presence of the most discrete datasets, we have conducted additional experiments excluding these 13 datasets. Discreteness was measured based on the proportions of values that appear among the 10 most frequent values in these datasets. The results of these new experiments are presented in Figure \ref{fig:some/without_discrete}. We have observed that the conclusions drawn in Section \ref{sec:experiments} remain unchanged. Both \texttt{RecTr (tr)} and \texttt{RecTr (tr) + Rec(ca)} demonstrate a significant improvement in NLL and PCE, with a negligible loss in CRPS. In summary, the combined approach of \texttt{RecTr (tr) + Rec(ca)} achieves a favorable balance between NLL and PCE.

\begin{figure}[H]
	\centering
	\subcaptionbox{
		Letter-value plots showing Cohen's d for different metrics with respect to \texttt{Base}.
		\label{fig:some/cohen_d}
	}{
		\includegraphics[width=\linewidth]{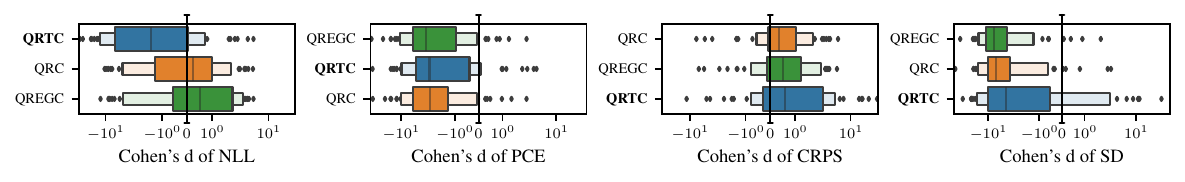}
		\vspace{-0.3cm}
	}
	\subcaptionbox{
		Critical difference diagrams for different metrics.
		\label{fig:some/cd_diagrams/test_nll}
	}{
		\includegraphics[width=\linewidth/4]{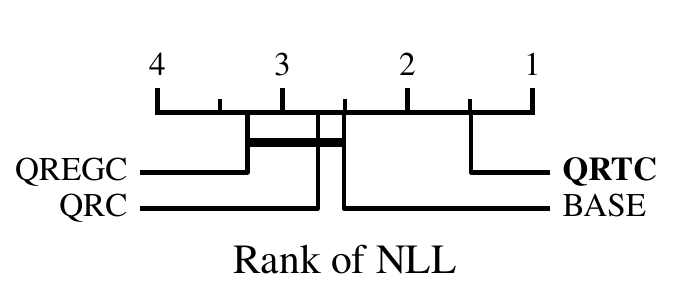}
		\includegraphics[width=\linewidth/4]{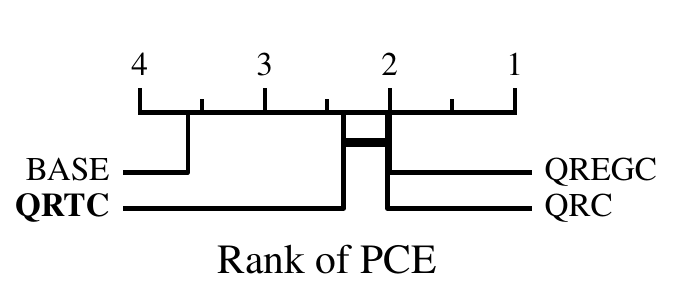}
		\includegraphics[width=\linewidth/4]{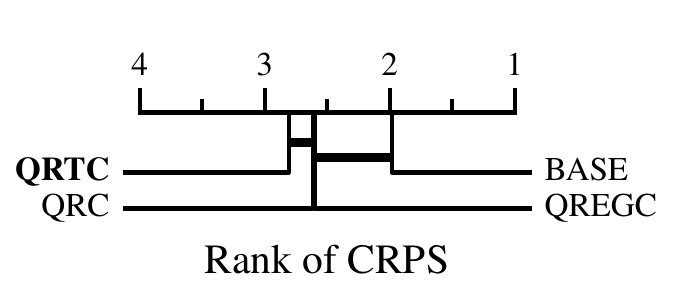}
		\includegraphics[width=\linewidth/4]{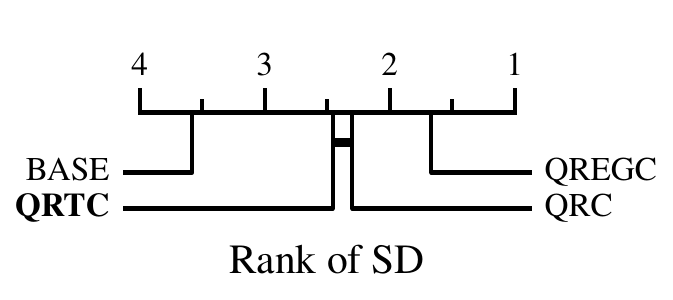}
		\vspace{-0.2cm}
	}
	\caption{Same setup than the main experiments (\cref{fig:some/without_discrete} in the main text), with all the datasets.}
	\vspace{-0.5cm}
	\label{fig:some}
\end{figure}

\clearpage
\section{RESULTS WHERE BASE DOES NOT HAVE ACCESS TO CALIBRATION DATA}
\label{sec:results/base_without_cal_data}

As discussed in \cref{sec:experimental_setup}, we aimed to provide a fair comparison between \tt{BASE} and the post-hoc methods by training it on a larger dataset than the post-hoc methods \tt{QRTC}, \tt{QRC} and \tt{QREGC} in all of our experiments.
Since the post-hoc methods benefit from the calibration data during the post-hoc step, all methods end up benefiting from the same amount of data.

In order to gain deeper insights into the effect of the post-hoc step, we repeat our main experiments with the exception that \tt{BASE} does not have access to calibration data.
Thus, \tt{QRC} has the same base model than \tt{BASE} and performs an additional post-hoc step.

\begin{figure}[H]
	\centering
	\subcaptionbox{
		Difference of test NLL compared to \tt{BASE}.
		\label{fig:base_without_calib/diff/test_nll}
	}{
		\includegraphics[width=\linewidth]{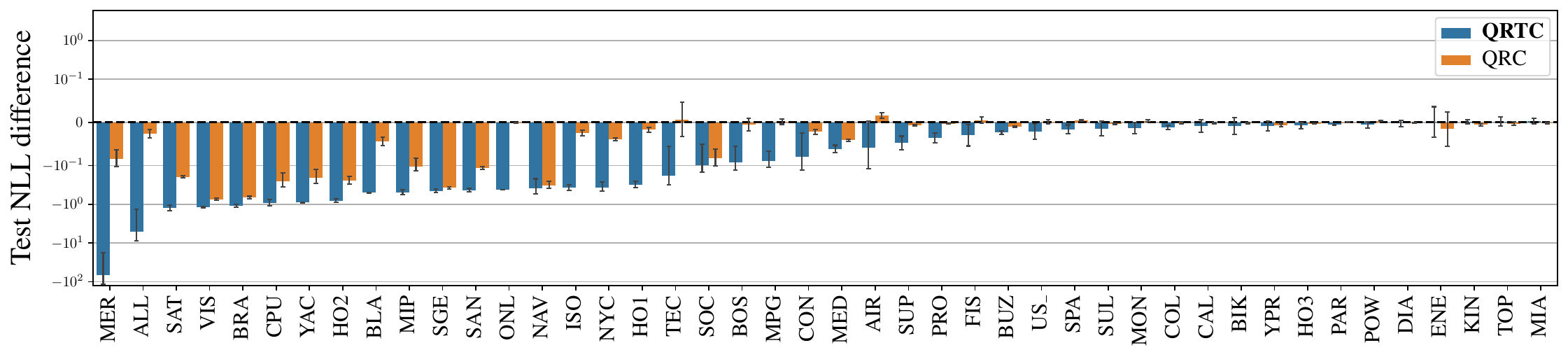}
		\vspace{-0.3cm}
	}
	\subcaptionbox{
		Letter-value plots showing Cohen's d for different metrics with respect to \texttt{BASE}.
	}{
		\includegraphics[width=\linewidth]{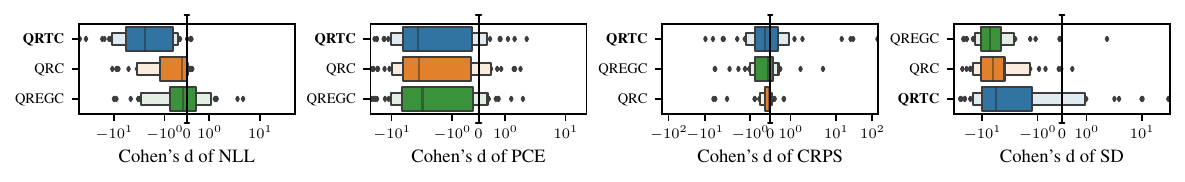}
		\vspace{-0.3cm}
	}
	\subcaptionbox{
		CD diagrams
		\label{fig:base_without_calib/without_discrete/cd_diagrams}
	}{
		\includegraphics[width=\linewidth/4]{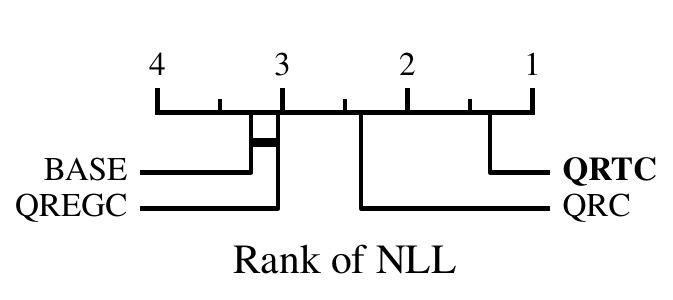}
		\includegraphics[width=\linewidth/4]{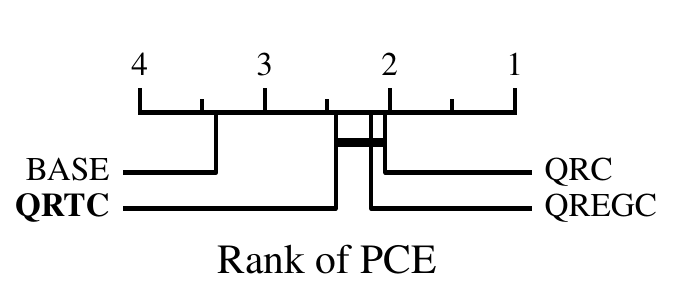}
		\includegraphics[width=\linewidth/4]{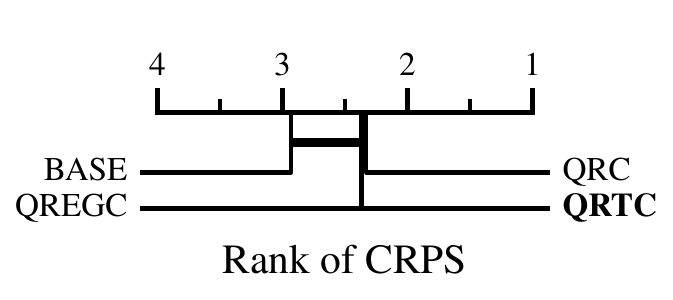}
		\includegraphics[width=\linewidth/4]{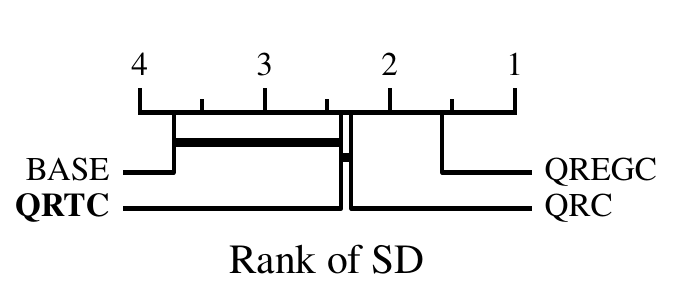}
		\vspace{-0.3cm}
	}
	\caption{Same setup than the main experiments (\cref{fig:some/without_discrete} in the main text), except that \tt{BASE} is not trained on the calibration data.}
	\label{fig:base_without_calib/without_discrete}
\end{figure}

In \cref{fig:base_without_calib/diff/test_nll}, \tt{QRC} shows that the post-hoc step never degrades NLL and sometimes results in a notable NLL improvement, which suggests that a post-hoc step on additional calibration data is always beneficial.
\cref{fig:base_without_calib/without_discrete/cd_diagrams} shows that \tt{QRC} results in a significant NLL improvement compared to \tt{BASE}, and \tt{QRTC} results in an additional significant NLL improvement compared to \tt{QRC}.
In terms of CRPS, there is no significant difference, and post-hoc methods result in sharper predictions.

\section{COMPUTATIONAL TIME}
\label{sec:time_measurements}

In \cref{table:times}, we present a comparative analysis of the training time for various methods across all datasets. Notably, \tt{QRTC} occasionally exhibits a training time that is approximately two times slower per epoch compared to \tt{BASE}. As discussed in \cref{sec:time_complexity}, this disparity can be attributed to the extra computational overhead associated with the computation of the calibration map, i.e., $-\frac{1}{B} \sum_{i=1}^B \log \phi^\text{REFL}_\theta(Z_i)$.

\begin{table}[H]
	\fontsize{7.5pt}{8.5pt}
	\selectfont
	\caption{Comparison of the training time for different methods on all datasets.}
	\label{table:times}
	\centering
	\begin{tabular}{l|rrr|rrr|rrr}
	\toprule
	& \multicolumn{3}{r}{Training time} & \multicolumn{3}{r}{Number of epochs} & \multicolumn{3}{r}{Time per epoch} \\
	& BASE & QREGC & QRTC & BASE & QREGC & QRTC & BASE & QREGC & QRTC \\
	Dataset &  &  &  &  &  &  &  &  & \\
	\midrule
	Airlines\_DepDelay\_10M & 485.59 & 521.41 & 732.92 & 31.40 & 35.40 & 44.20 & 15.46 & 14.73 & 16.58 \\
	Allstate\_Claims\_Severity & 284.53 & 1152.66 & 425.80 & 7.00 & 65.80 & 5.20 & 40.65 & 17.52 & 81.89 \\
	Buzzinsocialmedia\_Twitter & 923.56 & 997.29 & 863.84 & 84.60 & 93.00 & 53.20 & 10.92 & 10.72 & 16.24 \\
	MIP-2016-regression & 14.97 & 22.07 & 39.65 & 167.00 & 181.40 & 228.00 & 0.09 & 0.12 & 0.17 \\
	Moneyball & 7.55 & 22.62 & 14.63 & 78.00 & 141.00 & 74.60 & 0.10 & 0.16 & 0.20 \\
	SAT11-HAND-runtime-regression & 119.73 & 150.92 & 113.64 & 241.40 & 232.60 & 182.60 & 0.50 & 0.65 & 0.62 \\
	Santander\_transaction\_value & 24.01 & 26.02 & 26.60 & 3.80 & 5.00 & 3.80 & 6.32 & 5.20 & 7.00 \\
	Yolanda & 364.63 & 834.97 & 406.50 & 6.60 & 55.40 & 9.80 & 55.25 & 15.07 & 41.48 \\
	abalone & 35.27 & 31.20 & 77.67 & 52.60 & 49.60 & 107.80 & 0.67 & 0.63 & 0.72 \\
	boston & 7.38 & 5.56 & 7.32 & 72.60 & 57.00 & 48.20 & 0.10 & 0.10 & 0.15 \\
	colleges & 40.12 & 59.70 & 53.58 & 29.50 & 32.00 & 25.00 & 1.36 & 1.87 & 2.14 \\
	house\_prices\_nominal & 6.74 & 7.07 & 10.22 & 51.80 & 7.50 & 28.20 & 0.13 & 0.94 & 0.36 \\
	quake & 26.22 & 44.89 & 166.82 & 100.20 & 132.20 & 577.40 & 0.26 & 0.34 & 0.29 \\
	socmob & 9.04 & 20.84 & 15.47 & 112.60 & 118.60 & 67.80 & 0.08 & 0.18 & 0.23 \\
	space\_ga & 42.02 & 59.70 & 67.64 & 137.80 & 167.80 & 142.40 & 0.30 & 0.36 & 0.48 \\
	tecator & 11.87 & 7.37 & 7.83 & 195.00 & 185.80 & 111.80 & 0.06 & 0.04 & 0.07 \\
	topo\_2\_1 & 41.00 & 38.62 & 33.81 & 15.80 & 14.20 & 11.40 & 2.60 & 2.72 & 2.97 \\
	us\_crime & 10.19 & 28.18 & 16.38 & 21.00 & 67.40 & 22.60 & 0.49 & 0.42 & 0.72 \\
	Ailerons & 66.42 & 87.46 & 136.04 & 21.00 & 37.80 & 41.00 & 3.16 & 2.31 & 3.32 \\
	Bike\_Sharing\_Demand & 291.76 & 355.97 & 328.56 & 161.00 & 162.20 & 136.80 & 1.81 & 2.19 & 2.40 \\
	Brazilian\_houses & 159.17 & 337.55 & 169.54 & 137.40 & 230.20 & 128.20 & 1.16 & 1.47 & 1.32 \\
	MiamiHousing2016 & 118.08 & 181.59 & 125.74 & 62.20 & 101.80 & 64.20 & 1.90 & 1.78 & 1.96 \\
	california & 254.69 & 414.27 & 278.82 & 110.60 & 151.00 & 125.00 & 2.30 & 2.74 & 2.23 \\
	cpu\_act & 62.56 & 105.46 & 65.00 & 63.00 & 84.20 & 46.60 & 0.99 & 1.25 & 1.39 \\
	diamonds & 646.27 & 755.47 & 782.51 & 107.00 & 88.60 & 85.40 & 6.04 & 8.53 & 9.16 \\
	elevators & 92.17 & 97.59 & 308.20 & 21.00 & 30.20 & 103.00 & 4.39 & 3.23 & 2.99 \\
	fifa & 392.70 & 634.12 & 297.08 & 207.80 & 288.20 & 107.60 & 1.89 & 2.20 & 2.76 \\
	house\_16H & 200.43 & 539.11 & 388.17 & 67.80 & 185.00 & 115.40 & 2.96 & 2.91 & 3.36 \\
	house\_sales & 138.38 & 325.24 & 121.14 & 34.60 & 113.80 & 29.40 & 4.00 & 2.86 & 4.12 \\
	isolet & 128.97 & 222.28 & 187.13 & 136.20 & 222.60 & 167.00 & 0.95 & 1.00 & 1.12 \\
	medical\_charges & 687.66 & 984.32 & 840.41 & 62.20 & 75.00 & 49.80 & 11.06 & 13.12 & 16.88 \\
	nyc-taxi-green-dec-2016 & 1058.06 & 2083.84 & 1289.36 & 109.40 & 172.20 & 94.20 & 9.67 & 12.10 & 13.69 \\
	pol & 164.51 & 364.97 & 409.18 & 91.80 & 165.00 & 214.60 & 1.79 & 2.21 & 1.91 \\
	sulfur & 324.37 & 323.69 & 272.26 & 319.00 & 264.60 & 218.40 & 1.02 & 1.22 & 1.25 \\
	superconduct & 342.96 & 553.18 & 514.55 & 147.80 & 196.20 & 164.40 & 2.32 & 2.82 & 3.13 \\
	wine\_quality & 83.41 & 163.38 & 157.51 & 110.20 & 175.00 & 148.60 & 0.76 & 0.93 & 1.06 \\
	year & 327.90 & 392.19 & 465.72 & 7.40 & 9.00 & 9.40 & 44.31 & 43.58 & 49.54 \\
	Mercedes\_Benz\_Greener\_Manufacturing & 16.23 & 24.78 & 26.14 & 8.60 & 9.80 & 7.80 & 1.89 & 2.53 & 3.35 \\
	OnlineNewsPopularity & 137.57 & 182.44 & 232.98 & 7.80 & 10.20 & 14.60 & 17.64 & 17.89 & 15.96 \\
	SGEMM\_GPU\_kernel\_performance & 1229.68 & 1046.17 & 901.34 & 117.80 & 80.20 & 83.80 & 10.44 & 13.04 & 10.76 \\
	analcatdata\_supreme & 103.73 & 213.92 & 141.89 & 255.80 & 323.40 & 203.60 & 0.41 & 0.66 & 0.70 \\
	black\_friday & 558.56 & 742.27 & 1164.06 & 42.20 & 65.00 & 77.40 & 13.24 & 11.42 & 15.04 \\
	particulate-matter-ukair-2017 & 690.45 & 756.80 & 760.01 & 35.80 & 59.40 & 34.60 & 19.29 & 12.74 & 21.97 \\
	visualizing\_soil & 124.87 & 242.37 & 135.87 & 123.80 & 166.60 & 99.60 & 1.01 & 1.45 & 1.36 \\
	yprop\_4\_1 & 36.91 & 53.51 & 34.24 & 11.80 & 13.80 & 17.50 & 3.13 & 3.88 & 1.96 \\
	Airfoil & 32.33 & 52.81 & 44.84 & 270.20 & 370.60 & 329.40 & 0.12 & 0.14 & 0.14 \\
	CPU & 4.59 & 6.80 & 5.64 & 160.60 & 170.60 & 140.60 & 0.03 & 0.04 & 0.04 \\
	Concrete & 12.40 & 20.60 & 16.92 & 179.80 & 149.40 & 119.40 & 0.07 & 0.14 & 0.14 \\
	Crime & 4.76 & 8.83 & 7.27 & 43.00 & 53.80 & 42.20 & 0.11 & 0.16 & 0.17 \\
	Energy & 14.45 & 25.44 & 22.17 & 219.00 & 219.00 & 201.80 & 0.07 & 0.12 & 0.11 \\
	Fish & 6.13 & 10.07 & 11.06 & 76.20 & 66.60 & 72.20 & 0.08 & 0.15 & 0.15 \\
	Kin8nm & 49.79 & 72.99 & 58.75 & 45.00 & 51.40 & 40.60 & 1.11 & 1.42 & 1.45 \\
	MPG & 3.54 & 6.02 & 5.27 & 64.20 & 87.00 & 73.00 & 0.06 & 0.07 & 0.07 \\
	Naval & 180.63 & 318.65 & 261.46 & 145.40 & 200.60 & 190.60 & 1.24 & 1.59 & 1.37 \\
	Power & 174.46 & 207.21 & 238.07 & 203.80 & 186.20 & 193.00 & 0.86 & 1.11 & 1.23 \\
	Protein & 1030.51 & 1329.29 & 1275.15 & 244.60 & 235.00 & 216.60 & 4.21 & 5.66 & 5.89 \\
	Yacht & 6.73 & 8.85 & 11.13 & 189.40 & 169.40 & 228.40 & 0.04 & 0.05 & 0.05 \\
	\bottomrule
\end{tabular}
\end{table}

\section{TABULAR REGRESSION DATASETS}
\label{sec:appendix_datasets}

The datasets considered in our study are detailed in \cref{table:datasets}. The table provides information about the benchmark suite, full dataset name, abbreviations, number of training instances (truncated to 53,184 instances, similarly to \citet{Dheur2023-bo}), and the number of features. Additionally, the last two columns represent measures of the dataset's discreteness levels, as discussed in \cref{sec:discreteness}.
Proportions that are superior to 0.5 are highlighted in bold.

\begin{table}[H]
	\fontsize{7pt}{8pt}
	\selectfont
	\caption{Detailed properties of all datasets}
	\label{table:datasets}
	\centering
	\begin{tabular}{lll|rrrr}
	\toprule
	&  &  & \makecell{Nb of \\ training instances} & \makecell{Nb of \\ features} & \makecell{Proportion of \\ top 10 most \\ frequent values} & \makecell{Proportion of \\ duplicate values} \\
	Group & Dataset & Abbrev. &  & \\
	\midrule
	\multirow[t]{12}{*}{uci} & CPU & CP1 & 135 & 7 & 0.33 & \bfseries 0.64 \\
	& Yacht & YAC & 200 & 6 & 0.11 & 0.21 \\
	& MPG & MPG & 254 & 7 & 0.37 & \bfseries 0.78 \\
	& Energy & ENE & 499 & 9 & 0.05 & 0.22 \\
	& Crime & CRI & 531 & 104 & \bfseries 0.57 & \bfseries 0.95 \\
	& Fish & FIS & 590 & 6 & 0.04 & 0.12 \\
	& Concrete & CON & 669 & 8 & 0.04 & 0.10 \\
	& Airfoil & AI1 & 976 & 5 & 0.02 & 0.04 \\
	& Kin8nm & KIN & 5324 & 8 & 0.00 & 0.00 \\
	& Power & POW & 6219 & 4 & 0.01 & \bfseries 0.65 \\
	& Naval & NAV & 7757 & 17 & 0.40 & \bfseries 1.00 \\
	& Protein & PRO & 31328 & 9 & 0.01 & \bfseries 0.80 \\
	\multirow[t]{19}{*}{oml\_297} & wine\_quality & WIN & 4223 & 11 & \bfseries 1.00 & \bfseries 1.00 \\
	& isolet & ISO & 5068 & 613 & 0.40 & \bfseries 1.00 \\
	& cpu\_act & CP2 & 5324 & 21 & \bfseries 0.51 & \bfseries 1.00 \\
	& sulfur & SUL & 6552 & 6 & 0.01 & 0.09 \\
	& Brazilian\_houses & BRA & 6949 & 8 & 0.02 & \bfseries 0.58 \\
	& Ailerons & AIL & 8942 & 33 & \bfseries 0.86 & \bfseries 1.00 \\
	& MiamiHousing2016 & MIA & 9069 & 13 & 0.12 & \bfseries 0.91 \\
	& pol & POL & 9817 & 26 & \bfseries 0.98 & \bfseries 1.00 \\
	& elevators & ELE & 10936 & 16 & \bfseries 0.80 & \bfseries 1.00 \\
	& Bike\_Sharing\_Demand & BIK & 11482 & 6 & 0.11 & \bfseries 0.99 \\
	& fifa & FIF & 11961 & 5 & \bfseries 0.70 & \bfseries 1.00 \\
	& california & CAL & 13765 & 8 & 0.08 & \bfseries 0.94 \\
	& superconduct & SUP & 14201 & 79 & 0.05 & \bfseries 0.93 \\
	& house\_sales & HO3 & 14446 & 15 & 0.07 & \bfseries 0.87 \\
	& house\_16H & HO1 & 15266 & 16 & 0.17 & \bfseries 0.96 \\
	& diamonds & DIA & 37075 & 6 & 0.02 & \bfseries 0.89 \\
	& medical\_charges & MED & 53164 & 3 & 0.00 & 0.05 \\
	& year & YEA & 53164 & 90 & \bfseries 0.58 & \bfseries 1.00 \\
	& nyc-taxi-green-dec-2016 & NYC & 53164 & 9 & 0.38 & \bfseries 1.00 \\
	\multirow[t]{8}{*}{oml\_299} & analcatdata\_supreme & ANA & 2633 & 12 & \bfseries 1.00 & \bfseries 1.00 \\
	& \makecell{Mercedes\_Benz\\\_Greener\_Manufacturing} & MER & 2735 & 735 & 0.02 & \bfseries 0.53 \\
	& visualizing\_soil & VIS & 5616 & 5 & 0.43 & \bfseries 1.00 \\
	& yprop\_4\_1 & YPR & 5775 & 82 & 0.04 & \bfseries 0.94 \\
	& OnlineNewsPopularity & ONL & 27068 & 73 & 0.35 & \bfseries 0.99 \\
	& black\_friday & BLA & 53164 & 23 & 0.00 & \bfseries 0.95 \\
	& \makecell{SGEMM\_GPU\\\_kernel\_performance} & SGE & 53164 & 15 & 0.00 & \bfseries 0.70 \\
	& \makecell{particulate-matter\\-ukair-2017} & PAR & 53164 & 26 & 0.05 & \bfseries 0.92 \\
	\multirow[t]{18}{*}{oml\_269} & tecator & TEC & 156 & 124 & 0.19 & 0.48 \\
	& boston & BOS & 328 & 22 & 0.18 & \bfseries 0.73 \\
	& MIP-2016-regression & MIP & 708 & 111 & 0.05 & 0.17 \\
	& socmob & SOC & 751 & 39 & 0.36 & \bfseries 0.76 \\
	& Moneyball & MON & 800 & 18 & 0.09 & \bfseries 0.86 \\
	& house\_prices\_nominal & HO2 & 711 & 234 & 0.11 & \bfseries 0.59 \\
	& us\_crime & US\_ & 1295 & 101 & 0.37 & \bfseries 0.99 \\
	& quake & QUA & 1415 & 3 & \bfseries 1.00 & \bfseries 1.00 \\
	& space\_ga & SPA & 2019 & 6 & 0.01 & 0.00 \\
	& abalone & ABA & 2715 & 10 & \bfseries 0.90 & \bfseries 1.00 \\
	& \makecell{SAT11-HAND-\\runtime-regression} & SAT & 2886 & 118 & 0.06 & \bfseries 0.61 \\
	& \makecell{Santander\_transaction\\\_value} & SAN & 2898 & 3611 & 0.30 & \bfseries 0.73 \\
	& colleges & COL & 4351 & 34 & 0.03 & 0.44 \\
	& topo\_2\_1 & TOP & 5775 & 252 & 0.04 & \bfseries 0.94 \\
	& Allstate\_Claims\_Severity & ALL & 53164 & 477 & 0.00 & 0.10 \\
	& Yolanda & YOL & 53164 & 100 & \bfseries 0.58 & \bfseries 1.00 \\
	& Buzzinsocialmedia\_Twitter & BUZ & 53164 & 70 & 0.25 & \bfseries 0.98 \\
	& Airlines\_DepDelay\_10M & AI2 & 53164 & 5 & \bfseries 0.62 & \bfseries 1.00 \\
	\bottomrule
\end{tabular}
\end{table}

\section{EXAMPLES OF PREDICTIONS}

\begin{figure}[H]
	\centering
	\includegraphics[width=\linewidth]{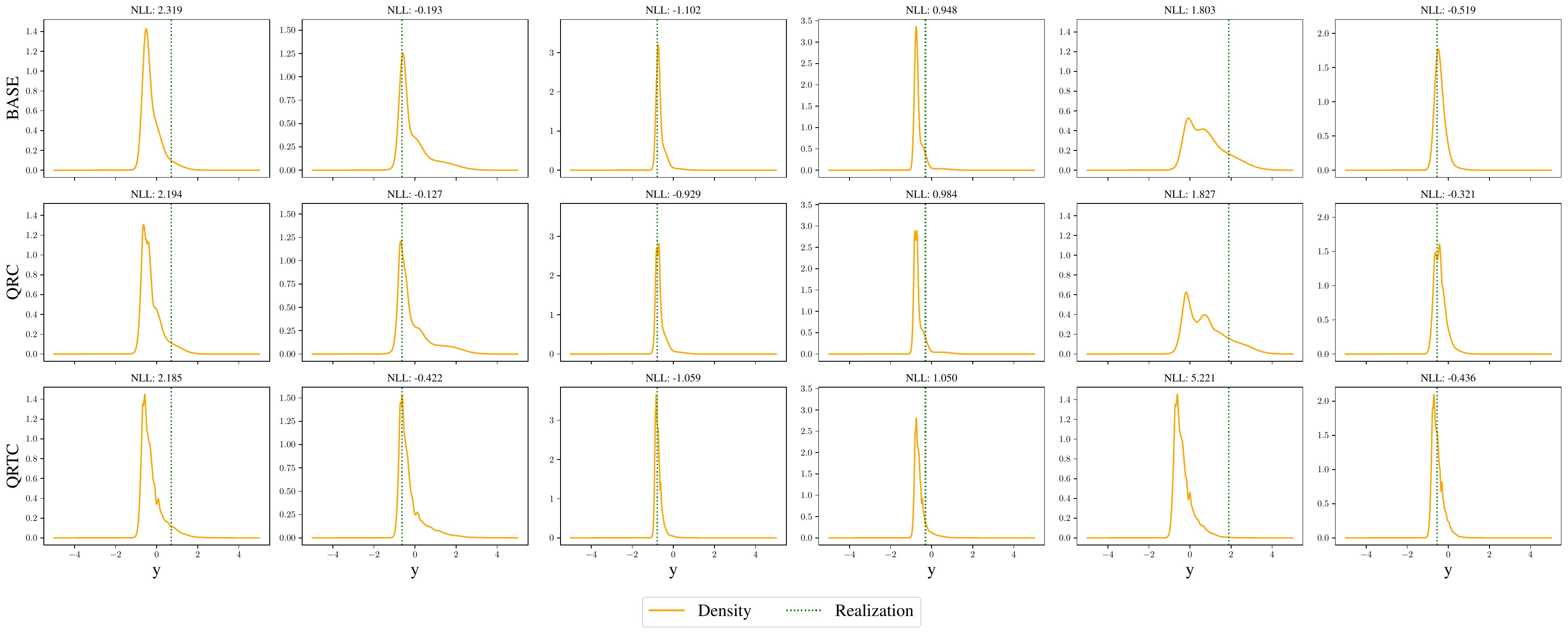}
	\vspace{-0.6cm}
	\caption{Examples of predictions of \tt{BASE}, \tt{QRC} and \tt{QRTC} on dataset \tt{Allstate\_Claims\_Severity} (ALL).}
	\label{fig:predictions/ALL}
\end{figure}

\begin{figure}[H]
	\centering
	\includegraphics[width=\linewidth]{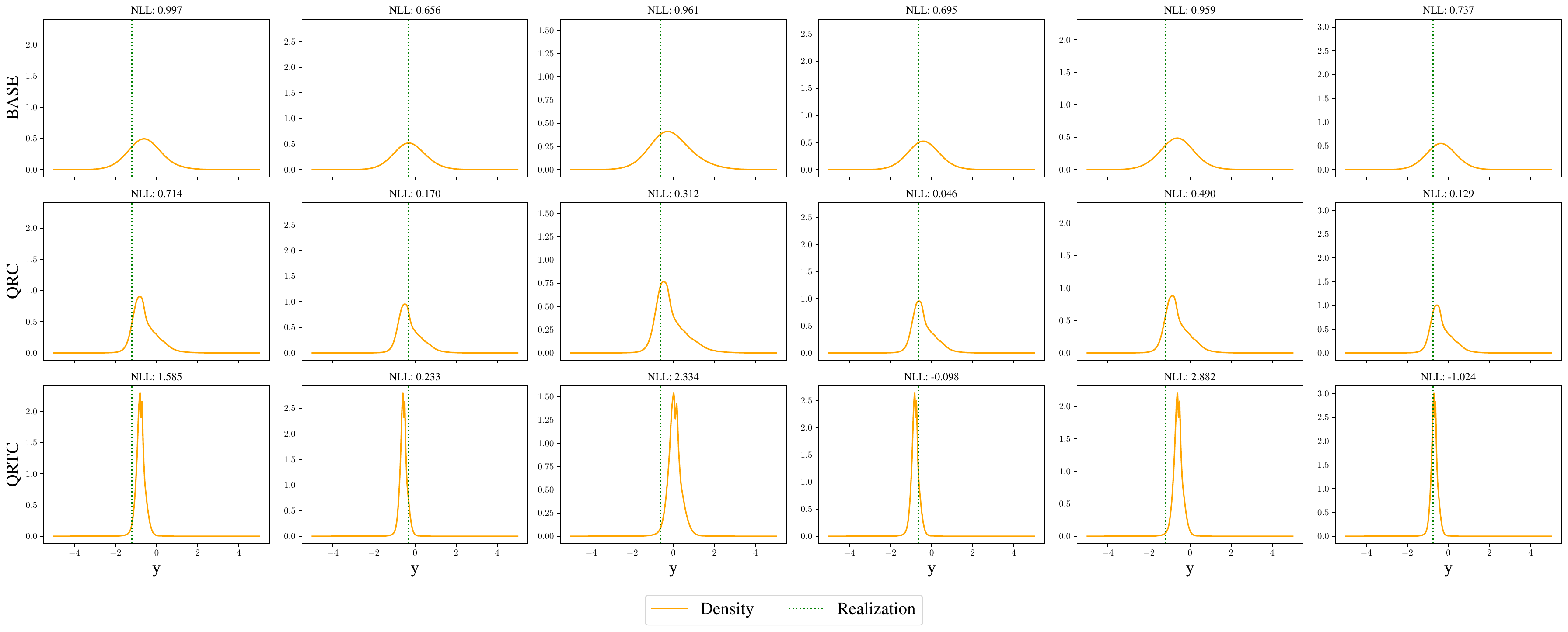}
	\vspace{-0.6cm}
	\caption{Predictions of \tt{BASE}, \tt{QRC} and \tt{QRTC} on dataset \tt{house\_prices\_nominal} (HO2).}
	\label{fig:predictions/HO2}
\end{figure}

\begin{figure}[H]
	\centering
	\includegraphics[width=\linewidth]{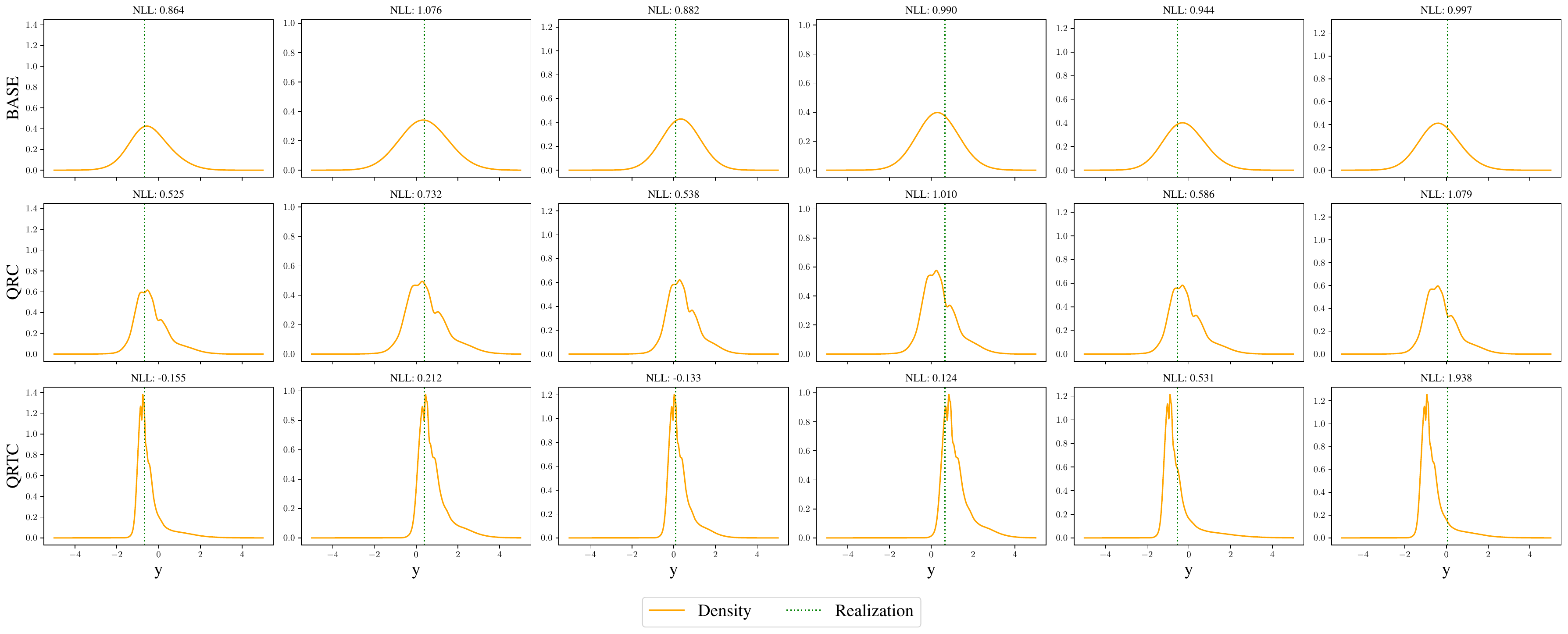}
	\vspace{-0.6cm}
	\caption{Predictions of \tt{BASE}, \tt{QRC} and \tt{QRTC} on dataset \tt{Mercedes\_Benz\_Greener\_Manufacturing} (MER).}
	\label{fig:predictions/MER}
\end{figure}

\begin{figure}[H]
	\centering
	\includegraphics[width=\linewidth]{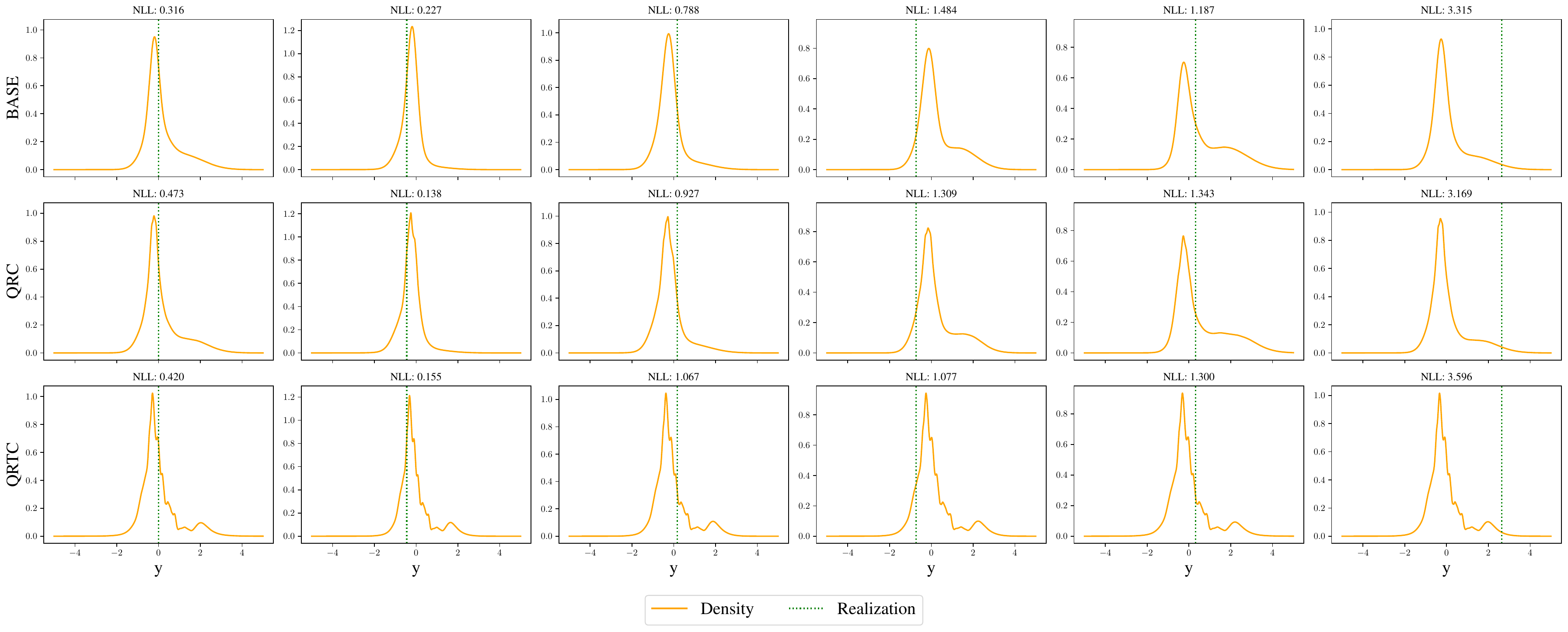}
	\vspace{-0.6cm}
	\caption{Predictions of \tt{BASE}, \tt{QRC} and \tt{QRTC} on dataset \tt{yprop\_4\_1} (YPR).}
	\label{fig:predictions/YPR}
\end{figure}

\begin{figure}[H]
	\centering
	\includegraphics[width=\linewidth]{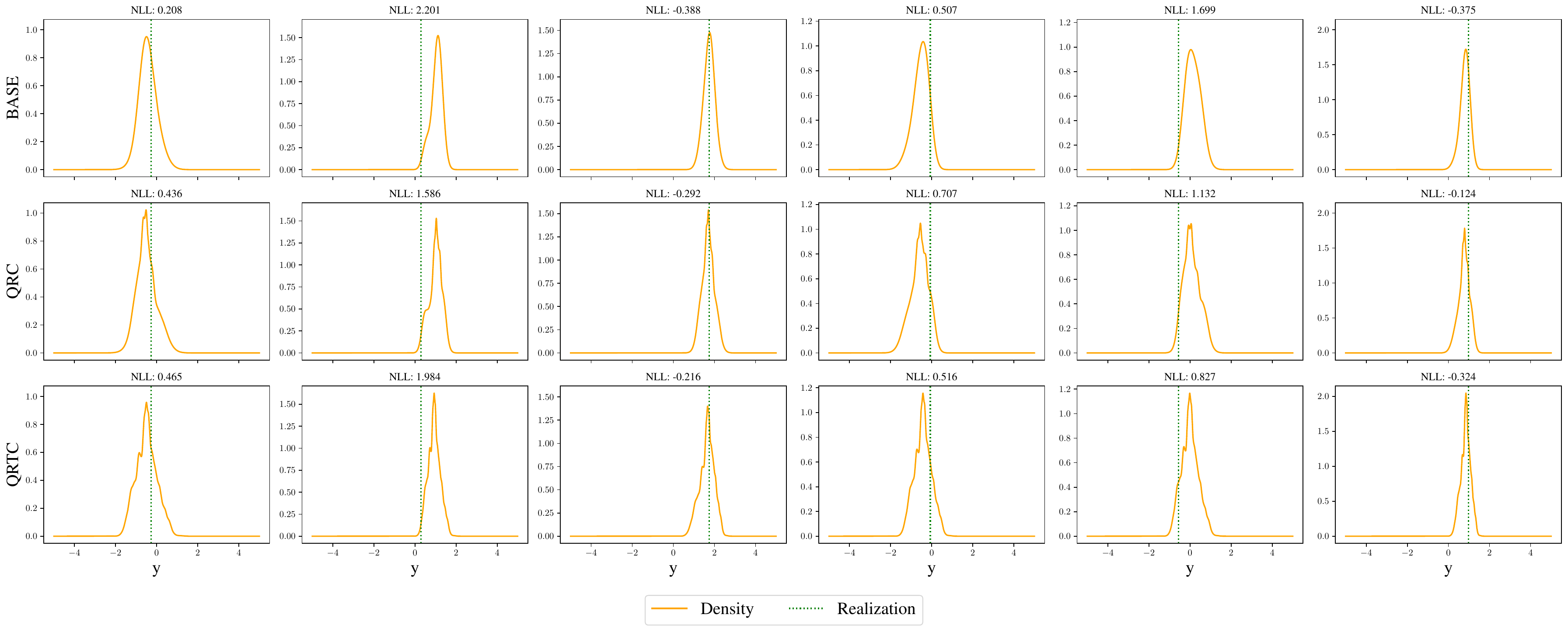}
	\vspace{-0.6cm}
	\caption{Predictions of \tt{BASE}, \tt{QRC} and \tt{QRTC} on dataset \tt{space\_ga} (SPA).}
	\label{fig:predictions/SPA}
\end{figure}

\begin{figure}[H]
	\centering
	\includegraphics[width=\linewidth]{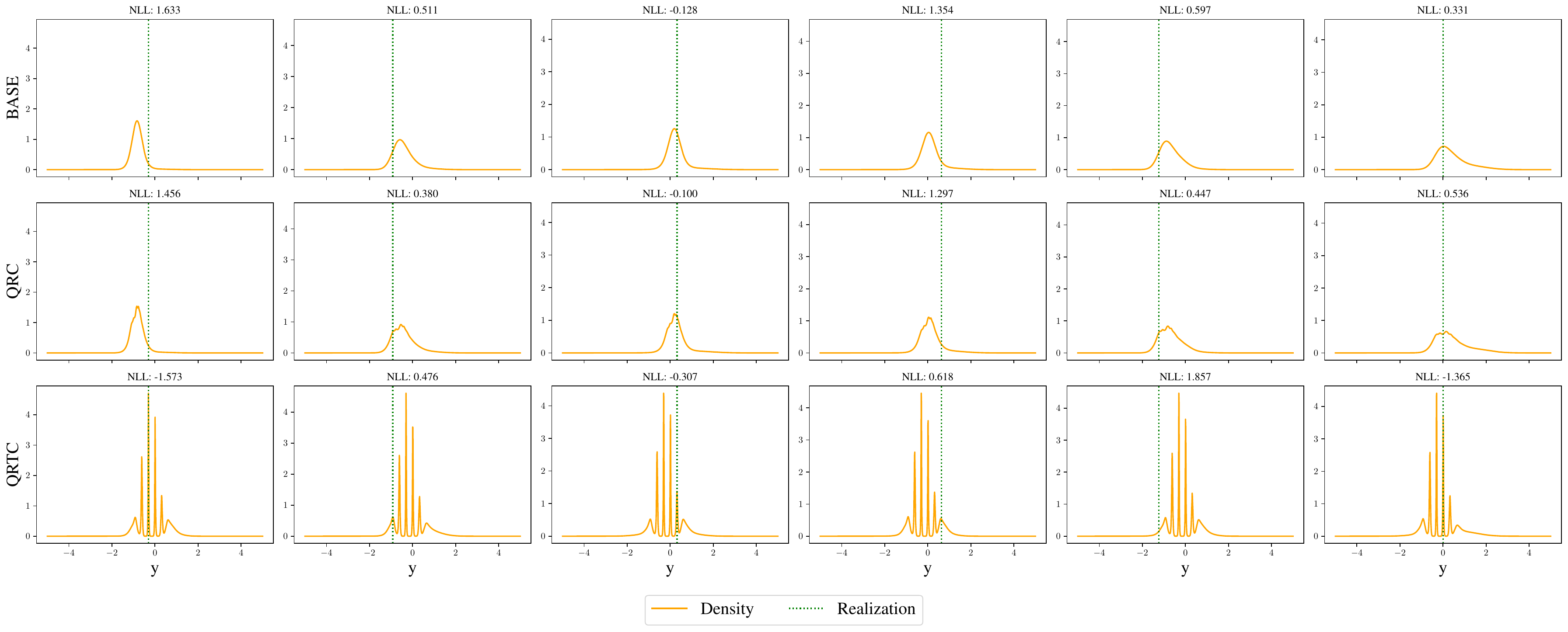}
	\vspace{-0.6cm}
	\caption{Predictions of \tt{BASE}, \tt{QRC} and \tt{QRTC} on dataset \tt{abalone} (ABA).}
	\label{fig:predictions/ABA}
\end{figure}

To offer a deeper understanding of the shape of the predictions, \cref{fig:predictions/ALL,fig:predictions/HO2,fig:predictions/MER,fig:predictions/YPR,fig:predictions/SPA,fig:predictions/ABA} display prediction examples across various datasets. In these figures, each row illustrates density predictions from the same model, while every column denotes the same instance, with the realization $y$ marked by a green vertical bar. The associated NLL for each prediction is also presented.

\cref{fig:predictions/ALL,fig:predictions/HO2,fig:predictions/MER,fig:predictions/YPR} are from datasets where \tt{QRTC} outperformed \tt{QRC} in terms of NLL. Notably, within these, \tt{QRTC} exhibits heightened confidence in its predictions for \cref{fig:predictions/HO2,fig:predictions/MER}. However, in other datasets, the NLL improvements are more subtle.
The \cref{fig:predictions/ABA} represents predictions on a dataset with a high level of discreteness which has not be considered in the main experiments. In this case, \tt{QRTC} assigns a high density to individual values $y$, highlighting a limitation of NLL minimization, as discussed in \cref{sec:discreteness}. 
Overall, the shape of the predictions can vary greatly in function of the dataset.

\vspace{-0.9cm}

\vspace{-0.9cm}

\vspace{-0.9cm}

\vspace{-0.9cm}

\end{document}